\documentclass[lettersize,journal]{IEEEtran}
\usepackage{amsmath,amsfonts,amssymb}
\usepackage{algorithmic}
\usepackage{algorithm}
\usepackage{array}
\usepackage[caption=false,font=normal,labelfont=sf,textfont=sf]{subfig}
\usepackage{textcomp}
\usepackage{stfloats}
\usepackage{url}
\usepackage{verbatim}
\usepackage{graphicx}
\usepackage{textcomp}
\usepackage{cite}[sort,compress]
\usepackage{calc}
\usepackage{booktabs}
\usepackage{adjustbox}
\usepackage{multirow}
\usepackage{hyperref}
\usepackage{color}
\usepackage{url}
\usepackage[table,usenames,dvipsnames]{xcolor}
\usepackage{calc}

\newlength\MAX
\newcommand*\Chart[5]{\rlap{\textcolor{#3!#5}{\rule[-0.5ex]{\MAX}{3ex}}}\rlap{\textcolor{#3!#4}{\rule[-0.5ex]{#2\MAX}{3ex}}}#1}

\begin{document}

\title{Image Matching by Bare Homography}

\author{Fabio Bellavia
\thanks{The author is with the Department of Mathematics and Computer Science, Universit\`{a} degli Studi di Palermo, Via Archirafi 34, 90123 Palermo, Italy (e-mail: fabio.bellavia@unipa.it).}
\thanks{Manuscript received \today; revised XX XX, 202X.}}

\markboth{arXiv,~2024}%
{Shell \MakeLowercase{\textit{et al.}}: A Sample Article Using IEEEtran.cls for IEEE Journals}


\maketitle

\begin{abstract}
This paper presents Slime, a novel non-deep image matching framework which models the scene as rough local overlapping planes. This intermediate representation sits in-between the local affine approximation of the keypoint patches and the global matching based on both spatial and similarity constraints, providing a progressive pruning of the correspondences, as planes are easier to handle with respect to general scenes.

Slime decomposes the images into overlapping regions at different scales and computes loose planar homographies. Planes are mutually extended by compatible matches and the images are split into fixed tiles, with only the best homographies retained for each pair of tiles. Stable matches are identified according to the consensus of the admissible stereo configurations provided by pairwise homographies. Within tiles, the rough planes are then merged according to their overlap in terms of matches and further consistent correspondences are extracted.

The whole process only involves homography constraints. As a result, both the coverage and the stability of correct matches over the scene are amplified, together with the ability to spot matches in challenging scenes, allowing traditional hybrid matching pipelines to make up lost ground against recent end-to-end deep matching methods.

In addition, the paper gives a thorough comparative analysis of recent state-of-the-art in image matching represented by end-to-end deep networks and hybrid pipelines. The evaluation considers both planar and non-planar scenes, taking into account critical and challenging scenarios including abrupt temporal image changes and strong variations in relative image rotations. According to this analysis, although the impressive progress done in this field, there is still a wide room for improvements to be investigated in future research.
\end{abstract}

\begin{IEEEkeywords}
Keypoint matching, local planar homography, affine constraints, SIFT, LoFTR, SuperGlue, AdaLAM, RANSAC.
\end{IEEEkeywords}

\vspace{-1em}
\section{Introduction}\label{intro}
\IEEEPARstart{I}{mage} matching is the core of many computer vision tasks requiring to localize the scene elements in the space and among images, including Structure-from-Motion (SfM), Simultaneous Localization and Mapping (SLAM) and image stitching, image registration and retrieval~\cite{szeliski}. The advancements in this research field have brought to practical applications such as autonomous driving or augmented reality for medicine and entertaining, which are changing human life.

\begin{figure}[t!]
	\center\vspace{-0.5em}
	\subfloat[\small Hz$^+$]{\label{hz_cart}
		\includegraphics[width=0.11\textwidth]{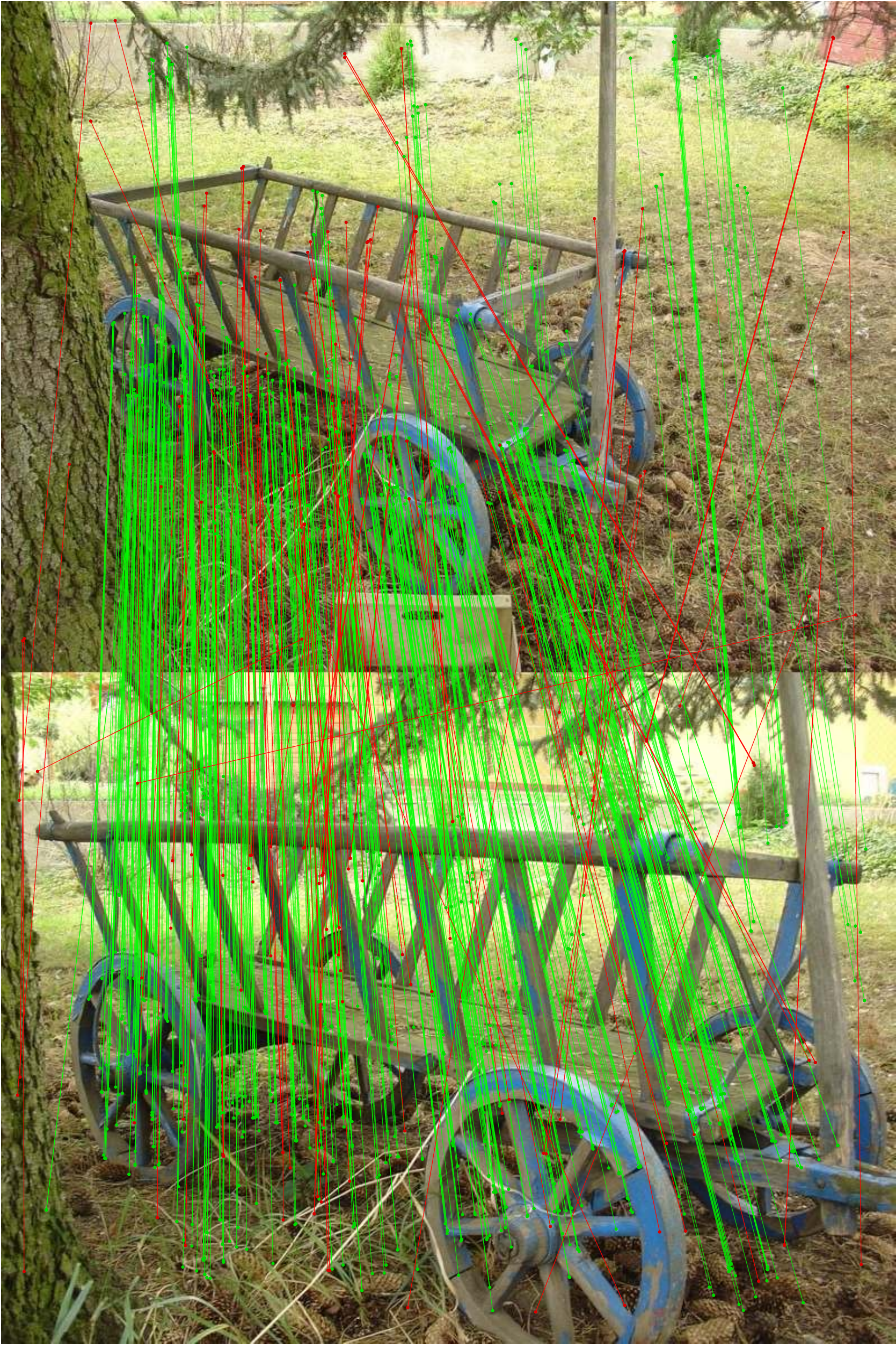}
	}
	\subfloat[\small Slime]{\label{slime_cart}
		\includegraphics[width=0.11\textwidth]{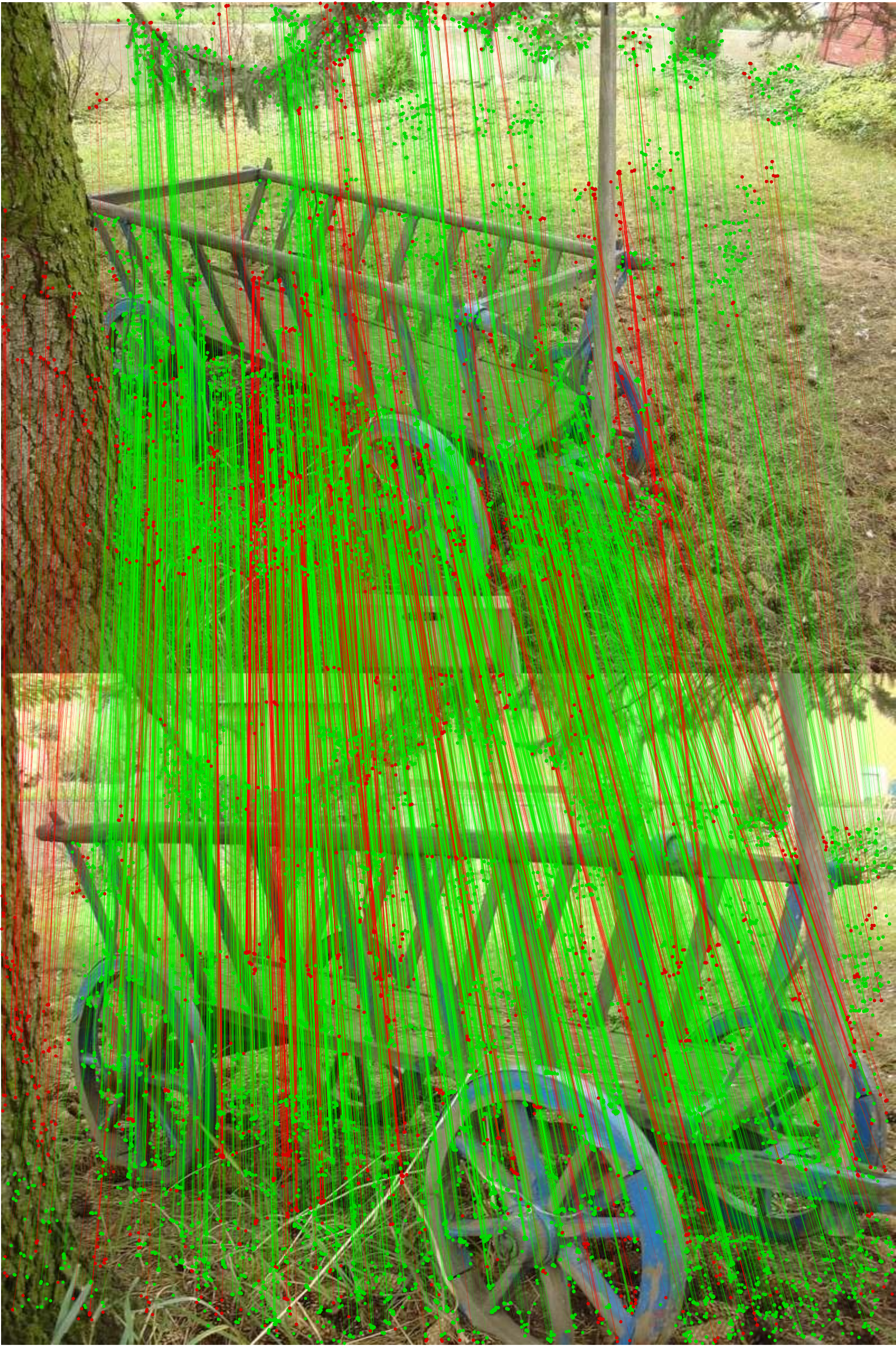}
	}
	\subfloat[\small Superglue]{\label{Superglue}
		\includegraphics[width=0.11\textwidth]{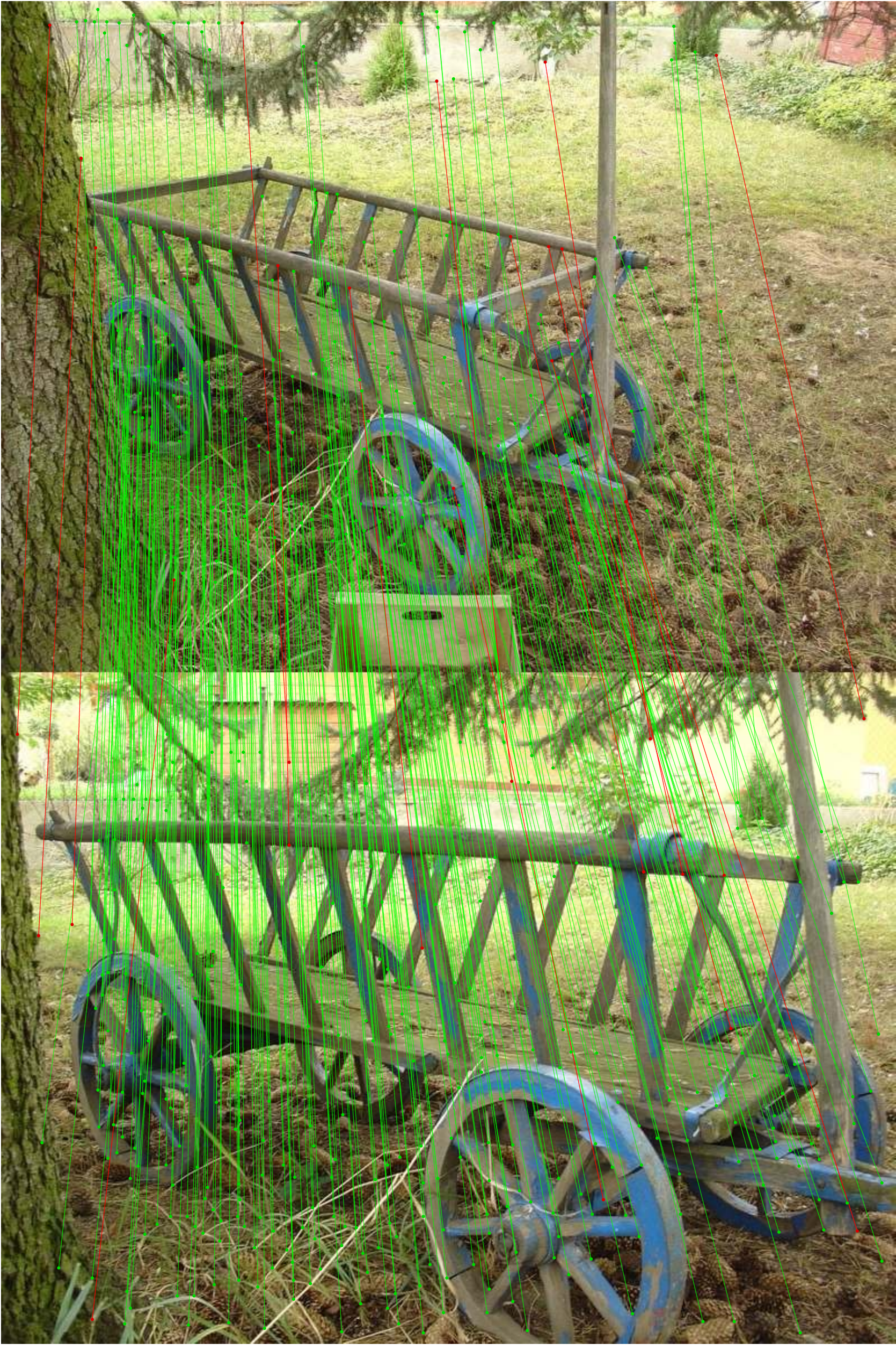}
	}
	\subfloat[\small LoFTR]{\label{LoFTR}
		\includegraphics[width=0.11\textwidth]{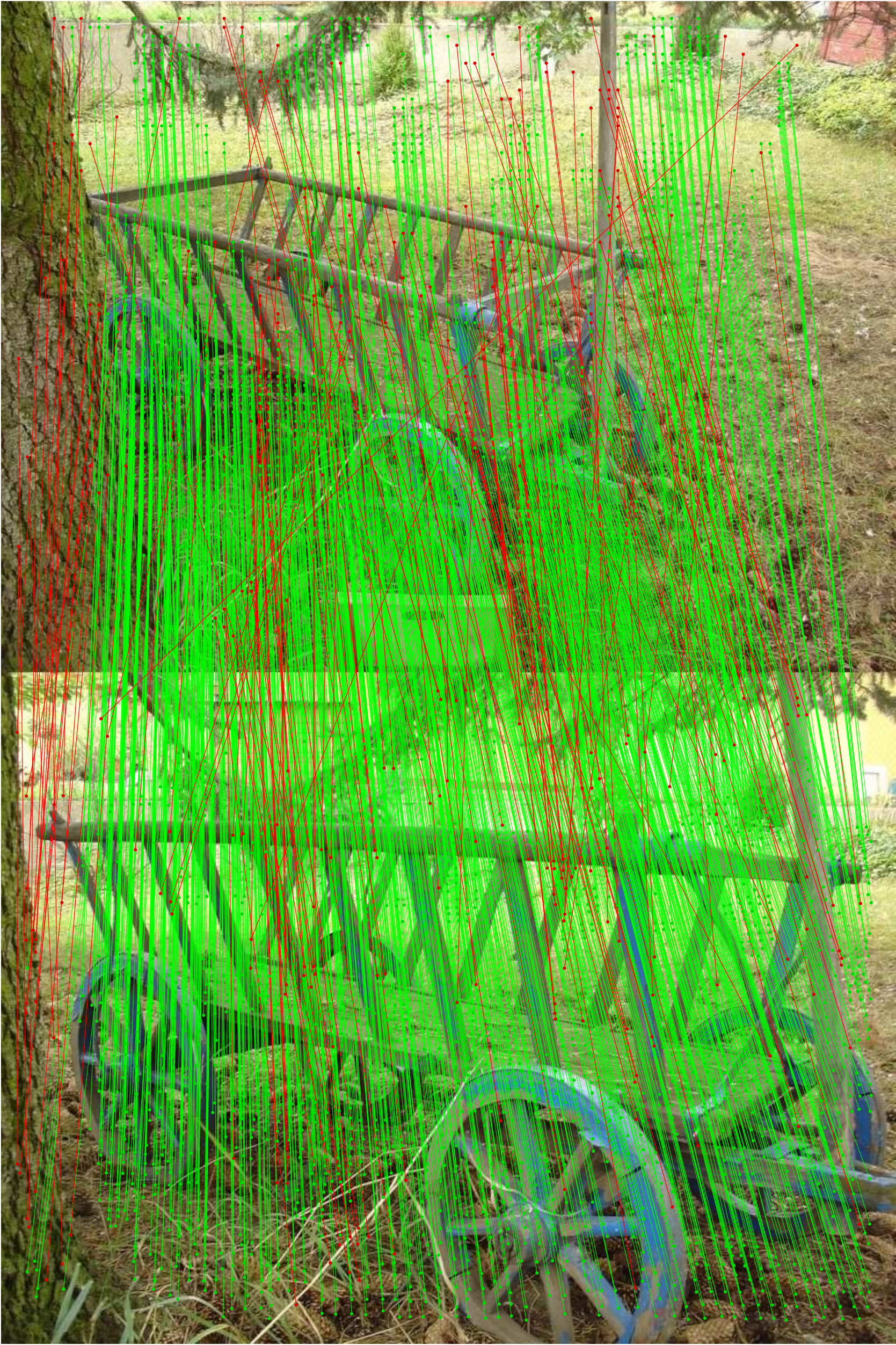}
	}	
	\caption{\label{cart}
	Extracted correct (green) and wrong (red) matches for a stereo pair without RANSAC for different image matching solutions including the proposed Slime (see Sec.~\protect\ref{intro}). Best viewed in color and zoomed in.}\vspace{-1.25em}
\end{figure}

Traditionally, image matching is designed as a pipeline that starts from a local affine approximation of the keypoint patches to define correspondences which are then globally filtered by similarity and geometric constraints~\cite{sift}. The engineered handcrafted modules of the matching pipeline have been progressively and successfully replaced, moving first to hybrid pipelines and evolving next into deep end-to-end networks. The latter ones represent nowadays the State-Of-The-Art (SOTA) in many circumstances, such as Superglue~\cite{superglue} and the semi-dense Local Feature Transformer (LoFTR)~\cite{loftr}. An interesting aspect in this evolution is that recent network structures tends to increasingly look like their handcrafted counterparts.

Against the trend, as contribution, this paper present Slime, a handcrafted image matching framework, able to ``stick together'' correspondences like Superglue, yet providing more ``gross'' results. The main idea of Slime is to add an intermediate local representation based on rough planes to better cluster together and filter spatially-close matches. Only local and global planar constraints, often in a loosen form, are used in Slime, integrating them so as to provide feedback mechanisms within the matching process.  

Slime starts by detecting candidate rough planes on different image regions and scales, by extracting matches using the hybrid Hz$^+$ pipeline~\cite{hz_pipeline} followed by the RANdom SAmple Consensus (RANSAC)~\cite{ransac} for a robust homography estimation. These planes are then expanded by incorporating further matches which adhere to the model, and unstable matches are discovered on the basis of a voting strategy relying on the observation that any two legit planes are able to define the stereo configuration of the scene. The relative global orientation between the image pair is estimated, again by voting, considering the keypoint patches of the surviving matches. Likewise the coarse-to-fine strategy of LoFTR, the images are divided into tiles, and the two mutual best homographies associated to each tile are retained, so as to model the local foreground and background as planes. Matches on tiles are then selected according to the foreground and background planar homographies, and finally filtered overall by the Delaunay Triangulation Matching (DTM)~\cite{dtm} considering spatial neighborhood constraints.

Fig.~\ref{cart} compares Slime matches against other SOTA methods, notice that paper high-resolution vector images are included in the Supplemental Material (SM) as SM\ref{high_res_fig}. Slime correct matches better cover the whole scene with respect to Superglue and the standalone Hz$^+$, being also quite close to LoFTR. Moreover, with respect to both Hz$^+$ and LoFTR, wrong matches are less spread and closer to the true optical flow due to the planar model constraints. As discussed later in the Sec.~\ref{evaluation}, Slime makes the base Hz$^+$ able to robustly detect matches in challenging conditions until now resolved only by end-to-end deep image matching architectures, where Hz$^+$ and others traditional handcrafted or hybrid SOTA pipelines fail.

As a further contribution, a thorough evaluation of the recent SOTA in image matching is provided. While latest benchmarks mainly focus on the pose error in 3D environments~\cite{imw2020}, the proposed evaluation also considers planar scenes. These can still offer challenges and lead to failure situations, as in the case of image rotations, not handled correctly by the most of end-to-end deep networks, but also in the case of repeated structures and strong viewpoint changes. Moreover, concerning non-planar scenes, day-night variations and temporal changes over long time periods in historical images are also analyzed, not commonly considered.

Another aspect in which the proposed evaluation differs from the others is for the metrics employed. In particular, the recall is replaced by a coverage measure to take into account semi-dense deep image matching networks, such as LoFTR, which can distort the recall measurement, and in the case of non-planar scenes the error on the camera pose is replaced by a error on the fundamental matrix~\cite{noransac} more directly related to the images and closer to the homography reprojection error used in planar scenes, so as to unify the two cases. 

For completeness, the evaluation on current standard indoor and outdoor pose-related benchmarks~\cite{loftr,matchformer,quadtree,superglue,lightglue,dkm} is also reported and discussed, with conclusions in accordance with the analysis drawn out in the new evaluation.

The rest of the paper is organized as follows: the related work is presented in Sec.~\ref{related_work}, Slime design is discussed in Sec.~\ref{slime_design} and the evaluation in Sec.~\ref{evaluation}. Finally, conclusions and future work are outlined in Sec.~\ref{conclusions}.

\section{Related Work}\label{related_work}
\subsection{Image matching pipelines}
The Scale Image Feature Transform (SIFT)~\cite{sift} can be reasonably assumed as the first effective and successful image matching pipeline, and it is still popular nowadays after around two decades. SIFT main pipeline consists of (a) detecting blob-like keypoints, (b) extracting and normalizing the keypoint local patches to be scale, rotation and illumination invariant, (c) computing the feature descriptors and (d) assigning matches according to the Nearest Neighbor Ratio (NNR) strategy. RANSAC is usually executed as post-process to filter the candidate matches according to the geometric constraints. Among the extensions of SIFT which have been proposed across the years RootSIFT~\cite{rootsift}, which replaces the Euclidean distance with the Hellinger's distance for comparing descriptors is generally preferred, as it usually provides slightly better results than the original SIFT. Affine patch normalization~\cite{affine} has been also included into the pipeline to improve robustness to viewpoint changes. Patch normalization have been profitably replaced by a deep counterpart in the Affine Network (AffNet)~\cite{affnet}. Deep dominant orientation estimation of the keypoint patches also obtained good results with OriNet~\cite{affnet} or previous deep architectures~\cite{learning_ori}. Local descriptor extraction was however the first element of the pipeline where machine learning was extensively experimented, as it copes with the feature extraction. The deep Hard Network (HardNet)~\cite{hardnet} is a SOTA standalone feature descriptor employed in many hybrid pipelines~\cite{hz_pipeline,keynet}.

The keypoint detection resisted more to deep approaches, but at last it also had to fall to them. A crucial roles in this process was to consider the keypoint extraction not as a standalone process, but as a step intrinsically connected with the descriptor computation as in SuperPoint\cite{superpoint}. This leads to the introduction of the concept of self-supervised training, were both the networks for keypoint extraction and description get jointly optimized in a single pass so as to better work together. Notice that self-supervised training also exploits homography adaptation to achieve its aim, where corresponding training image pairs are generated by apply general planar homographies to a single input images. Another key point towards the inclusion of deep keypoint detectors in the image matching pipeline was to imitate the handcrafted keypoint detector design in the network structure~\cite{keynet,d2net}.

The final part of the image matching pipeline associates keypoints from the image pairs according to their spatial information and their similarities. On one hand, two keypoints are deemed similar on the basis of the distance between their descriptors, and the best matches are selected by a Nearest Neighbor (NN) based strategy~\cite{dtm}. On the other hand, spatial information are used to filter the candidate matches imposing geometric constraints. This is achieved traditionally by RANSAC, distinguishing between planar and non-planar scenes so as to constraint the model respectively by a planar homography or by the epipolar geometry through the definition of the fundamental matrix~\cite{multiview}. Another constraint is provided by the relative global orientation between the images~\cite{sgloh2}, which can be estimate during the matching process or given a priori, for instance in case of autonomous driving. Clearly, better solutions are generally granted in case of planar scenes since it is a more constrained problem.

Besides RANSAC, others handcrafted spatial filters have been designed to pre-filter candidate NN matches before applying RANSAC. These spatial filters are based on the general notion of spatial neighbor. The First Geometrically Inconsistent NN (FGINN)~\cite{fginn} matching strategy follows this idea by extending NNR so as to consider also the keypoint location for the selection of the second best match, while more elaborated notions of spatial neighbor are employed in the Locality Preserving Matching (LPM)\mbox{\cite{lpm}} and the Grid-based Motion Statistics (GMS)~\cite{gms}. Local neighborhoods can be also exploited by running and fusing local affine RANSACs such in the case of the Adaptive Locally-Affine Matching (AdaLAM)~\cite{adalam}. One of the main issues of these approaches is to require an robust sets of matches for initialization. DTM implements a different strategy by alternating between contractions and expansions of the candidate match set on the basis of the neighborhood relations given by Delaunay triangulation and according to the match similarity rank, limiting the initial candidate selection issue and providing an amalgamation of similarity and spatial information.

The problem of fitting multiple instances of a geometric model, in the specific case of homographies, is closely related to matching with AdaLAM as well as the proposed Slime. J-Linkage\mbox{\cite{j_linkage}} and the more recent Connected Components Sampler (CC)~\cite{ccs} are exemplar approaches in this sense. Notice however that these strategies only implement a filtering stage, unlike Slime which also robustly builds and connects the correspondences from the images.

More recently, deep solutions has started to replace the NN selection plus RANSAC. The breakthrough was the introduction of the context normalization\cite{cne} to exploit contextual information while preserving permutation equivariance. Later, the Order-Aware Network (OANet)~\cite{oanet} has defined network layers to learn how to cluster unordered sets of correspondences so as to incorporate the data context and the spatial correlation. The advent of the attention in deep learning marked the overtake of deep methods on handcrafted ones. In particular, SuperGlue expands SuperPoint by adding an attentional graph neural network to capture spatial information followed by the Sinkhorn algorithm to get the final matches, giving the first effective end-to-end image matching network. LightGlue~\cite{lightglue} revisits SuperGlue by introducing design solutions such as an early stopping condition, making its architecture more efficient, accurate and easier to train. Other alternative approaches achieving interesting results have been also investigated, such as the DIScrete Keypoints (DISK)~\cite{disk} which leverages on reinforcement learning to train a full end-to-end image matching pipeline. For an extensive survey on image matching up to SuperGlue please refer to~\cite{im_survey}.

SuperGlue matches are quite sparse and can be problematic~\cite{kfc} in case of applications such as SfM~\cite{colmap}. In order to compensate for this lack, more recent research focuses on semi-dense deep image matching. LoFTR can be considered the forerunner of this research direction. Specifically, LoFTR uses attention to connect spatially distant image areas in a coarse-to-fine manner, so as to first roughly localize matched regions of the images and then densely refine the flow estimation on them. Recent works extend and improve the LoFTR basic network structure. SE2-LoFTR~\cite{se2loftr} replaces the backbone convolutional layer for the feature extraction with a steerable one, so as to make features invariant to the $SE(2)$ group, in practice adding rotational invariance to LoFTR. MatchFormer~\cite{matchformer} employs instead more levels of attention at different scales and region overlaps to implicitly provide positional embedding so as to better refine progressively the correspondences. Likewise, the QuadTree Attention network~\cite{quadtree} exploits a quadree-based multi-scale attention to gradually filter matched regions.

Another semi-dense attention-based approach is provided by the COrrespondence TRansformer (COTR)~\cite{cotr} where transformers recursively provide for each query point in one image the corresponding match in the other image. COTR requires a high computational cost, which has been reduced by adding multi-scale attention and adaptively clustering close matches with similar flow through the Efficient COrrespondence Transformer Network (ECO-TR)~\cite{ecotr}. 

Lastly, deep dense methods originally relegated to small-baseline image matching have evolved with impressive results. These approaches define the flow estimation between correspondences as a probabilistic regression model so as to output the estimated match together with its uncertainly, as in the case of the Enhanced Probabilistic Dense Correspondence Network (PDC-Net+)~\cite{pdcnetplus}. Among these, the Dense Kernelized feature Matching (DKM)~\cite{dkm} achieves SOTA results by modeling the coordinate regression as a Gaussian process. Notice that this kind of warping scheme has been already investigated in handcrafted matching refinement strategy such as the Vector Field Consensus (VFC)~\cite{vfc}.

\subsection{Image matching benchmarks}
Image matching benchmarks has evolved together with the matching methods. The evaluation on planar scenes is generally less complex than on non-planar ones, as planar homographies give one-to-one point mapping, so that Ground-Truth (GT) data can be straightforwardly obtained by hand. The Oxford~\cite{affine_eval} and Homography Patches (HPatches)~\cite{hpatches} datasets are designed in this way.

On the contrary, finding a reliable GT in case of non-planar scenes is more complex. The most general and common solutions to get a GT in this case rely on multiview constraints. The Turntable dataset~\cite{turntable} is an early example is this sense, exploiting epipolar constraint triplets to discard wrong matches, while recent solutions employ the 3D reconstruction of the scene obtained by SfM~\cite{colmap_eval} as GT. The Brown dataset~\cite{brown}, containing patches extracted through SfM for training and testing keypoint descriptors, is one of the most popular and among the first datasets with a SfM-based GT. The Image Matching Challenge PhotoTourism (IMC-PT) dataset~\cite{imw2020} built by SfM on photo collections from internet has been used in several recent studies. Sensor-based scans have been deployed with the aim of obtaining non-planar GT, such in case of the Strecha~\cite{strecha}, the Denmark Technical University (DTU)~\cite{dtu} and the Tanks and Temples~\cite{tanks_and_temples} datasets, but in general these solutions require complex expensive setups and are not feasible for any scenario. The development and availability of RGB-D cameras for augmented reality applications drive nowadays the benchmark progresses in this direction, with the design of indoor datasets like the Scene UNderstaning 3D (SUN3D)~\cite{sun3d} and the ScanNet~\cite{scannet} datasets, derived from the integration of multi-sensor and SfM, and of more recent datasets also including outdoor scenes, as in the case of the Localization And Mapping for Augmented Reality (LAMAR)~\cite{lamar} dataset. Finally, synthetic datasets of virtual rendered scenes may be employed~\cite{enrich}, which poses no problem with the GT estimation, yet cannot reach the fidelity level of real images.  

The choice of the error metric is a further relevant aspect in benchmark design. The localization accuracy and the coverage of correct matches are the ideal benchmark quantities for the evaluation, corresponding to the precision and recall measures. In this sense, the optimal GT must allow to compute the reprojection error between matches. This is relatively straightforward for planar scenes, but can be problematic for non-planar scenes since only a partial and approximate GT can be available. In case the evaluation is more stressed towards the matching coverage, combining epipolar geometry with local information collected by sparse hand-taken correspondences to define the GT can be a plausible approach~\cite{dtm}. An alternative metric in case the evaluation is more focused on the precision of the matches can be to use as a proxy to the effective matching error the error on the camera poses~\cite{imw2020}, also in the case of planar scenes~\cite{heb}, or the error on the fundamental matrix relating two images~\cite{noransac,fund_mat_eval}.

Another aspect to take into account in benchmark design is the kind of scene and the image transformations present in the dataset. Planar scenes commonly include several degrees of viewpoint, illumination, scale, rotation and blur changes~\cite{affine_eval,dtm} or focus on more specific image transformations~\cite{hpatches} such as viewpoint changes and illumination changes, reasonably considered in order the most critical ones~\cite{mods}. In the case of non-planar scenes, with some exceptions~\cite{wxbs}, the interest in investigating wider range of scene changes has grown only more recently with the advance of the matching methods. For instance, the Aachen Day-Night dataset~\cite{aanchen} includes quite different season and day-nigh changes on the scene, while~\cite{maiwald,low3d,historical_aerial} provides challenging temporal changes across present and historical images. Besides the kind of image transformations, having a wide assortment of different scenes is crucial for providing robust comparative analyses, as well as for training in the case of deep methods. Datasets designed for specific tasks are also available, such as KITTI~\cite{kitti} for autonomous driving, SUN3D and ScanNet for indoor environments, those proposed in~\cite{enrich,cultural_heritage} for cultural heritage specific applications, or BlendedMVS~\cite{blendedmvs} for training deep networks for Multi-View Stereo (MVS). The number of  different scenes present may also vary with the datasets. For instance, HPatches contains around on hundred different planar scenes, while the non-planar large-scale indoor and outdoor datasets ScanNet and MegaDepth~\cite{megadepth} contain more than half thousand and about two hundreds scenes, respectively. Nevertheless, in the current protocols the majority of these scenes are actually reserved to the training phase.

\section{Slime}\label{slime_design}

\subsection{Multiscale image block representation}\label{block_conf}
\begin{figure}[t!]
	\center
	\includegraphics[width=0.25\textwidth]{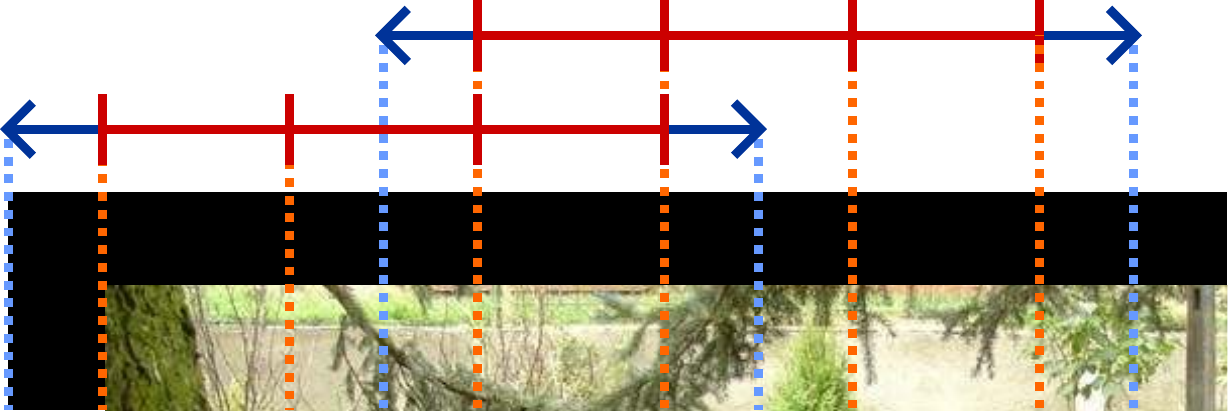}
	\caption{\label{ob}
	Image block construction (see Sec.~\ref{block_conf}). Blue and red segments are 32 and 64 px, respectively, for a block side length of $32\times2+64\times3=256$ px. Default stride is 128 px, but after adjustment ranges in $128\pm24$ px. The linear overlap is $\frac{1}{3}$ before the stride adjustment since no keypoints are extracted under the blue segments by design (see Sec.~\ref{block_conf}). Best viewed in color and zoomed in.}
\end{figure}

\begin{figure}[t!]
	\center
	\includegraphics[width=0.45\textwidth]{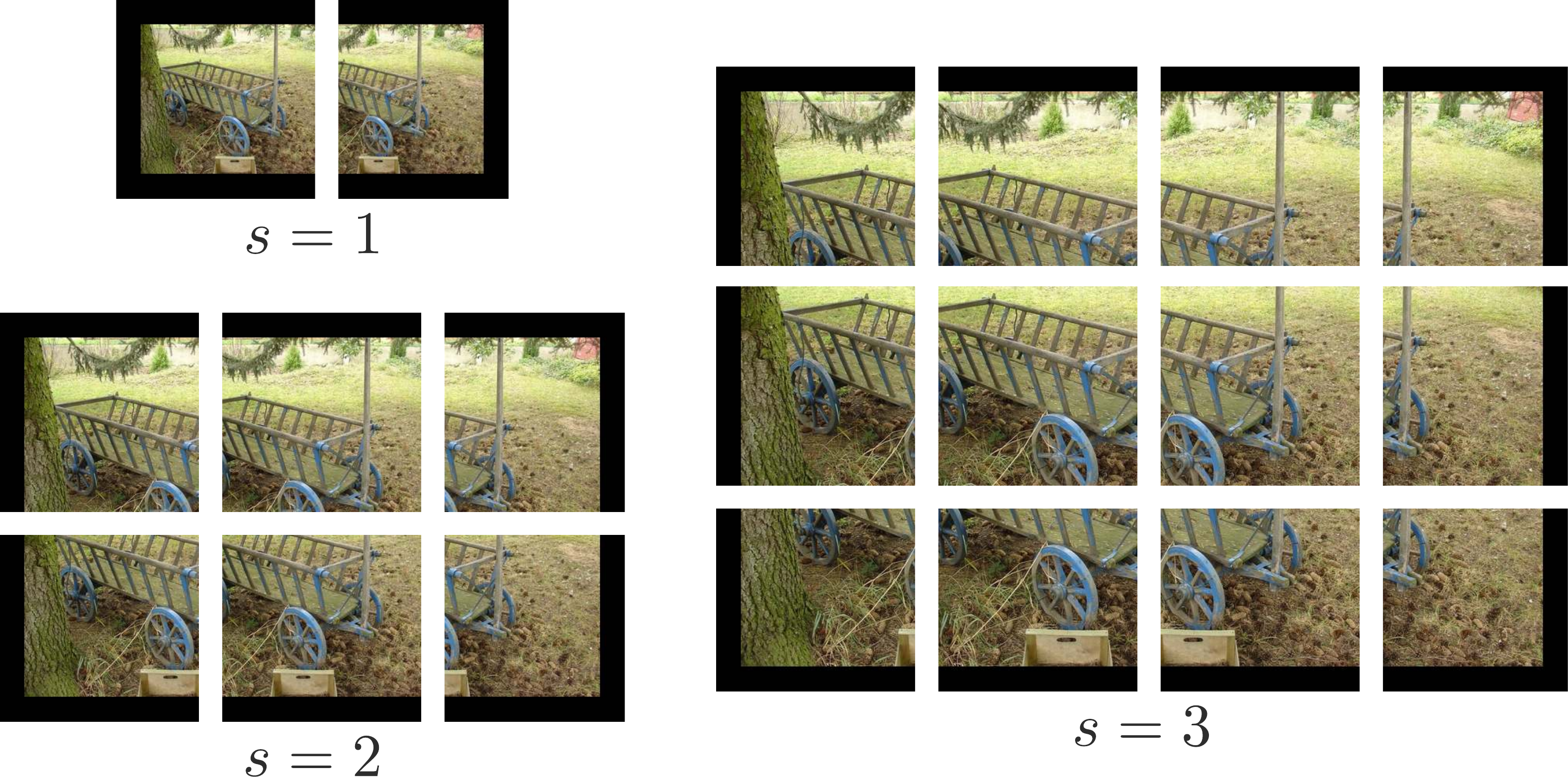}
	\caption{\label{blocks}
	Multiscale block decomposition for the top image of the pair of Fig.~\ref{cart} at the different block scale $s$, the padding area is colored in black (see Sec.~\ref{block_conf}). Best viewed in color and zoomed in.}\vspace{-0.25em}
\end{figure}

Given the input pair of images $(I_1,I_2)$, each image $I_k$, $k=1,2$, is considered according to a multi-scale representation and divided into overlapping blocks of $256\times256$ px. Three scales are used, so that along the dimension of minimum length there are respectively $s=1,2,3$ blocks. The block structure is shown in Fig.~\ref{ob}, images are padded by 32 px to correctly extract keypoints near the borders. Actually, the overlap is softly adjusted to reduce the total number of blocks and to avoid that blocks in the bottom row and in the rightmost column cover an exiguous image area. These latter blocks can be less than 256 px on each side. Lanczos3 interpolation is used for scaling image blocks since there are evidences that it provides better keypoints~\cite{hz_pipeline}. As notation, a block is denoted as $b^k_{sij}=b^k_w$, where in the first case both the scale $s$ and the row, column  grid indexes $i,j$ are indicated, while in the other case only the linear index $w$ among the set of all the blocks of the image $\mathcal{B}_k=\{b^k_{sij}\}=\{b^k_w\}$ is given. An example of the multi-scale block decomposition is shown in Fig.~\ref{blocks}.

\subsection{Block-wise planar matching}\label{multiscale_block}
The block pairs $(b^1_w,b^2_{w'})\in \mathcal{B}_1\times\mathcal{B}_2$ are independently processed to extract matches constrained by planar homography. For this aim, the Hz$^+$+AffNet+OriNet+HardNet+DTM matching pipeline with default setting is employed~\cite{hz_pipeline}, followed by RANSAC. Notice that the Hz$^{+}$ detector is adaptive, i.e. it can extract different keypoints for the overlapped region in distinct blocks, and DTM outputs many-to-many matches which are all retained for the successive steps. For RANSAC, a match is considered as inlier if the maximum reprojection error given the homography $H$ is below the given threshold $t_\perp=15$ px
\begin{equation}\label{max_reproj}
	\epsilon^H=\max(\parallel\mathbf{x}'-H\mathbf{x}\parallel,\parallel\mathbf{x}-H^{-1}\mathbf{x}'\parallel)<t_\perp	
\end{equation}
where points are intended in normalized homogeneous coordinates and the match pair $(p,p')$ identifies the keypoint patch $p=(\mathbf{x},\theta,\sigma)\in I_1$ as  triplet in which $\mathbf{x}\in\mathbb{R}^2$, $\theta$ and $\sigma$ are respectively the keypoint position, orientation and scale, and likewise for $p'\in I_2$. The notation $\epsilon^H$ should have included the pair $(p,p')$ as argument, but it is dropped for simplicity. The base assumption for the block-wise matching is that the scene mapping within the two views can be locally approximated by planes. The adopted RANSAC error guarantees to have in each image an exact bound on the achieved error, while the relatively high threshold allows to relax the strict planar model constraint. The overlap between neighbor blocks allows smooth changes over planes within the images. For each block pair $(b^1_w,b^2_{w'})$ a set of matches
\begin{equation}
	\tilde{\mathcal{M}}_l=\tilde{\mathcal{M}}_{ww'}=\{(p,p'): p\in b^1_w, p'\in b^2_{w'},\epsilon^{H_l}<t_\perp\}	
\end{equation} is obtained, where $l=(w,w')$ is the linear index for the block pair associated to indexes $w$ and $w'$. An example match set $\tilde{\mathcal{M}}_l$ is shown in Fig.~\ref{base_match}.

\subsection{Refinement of the local planes by their relative orientation}\label{block_match}
\begin{figure}[t]
	\center
	\includegraphics[height=0.15\textwidth]{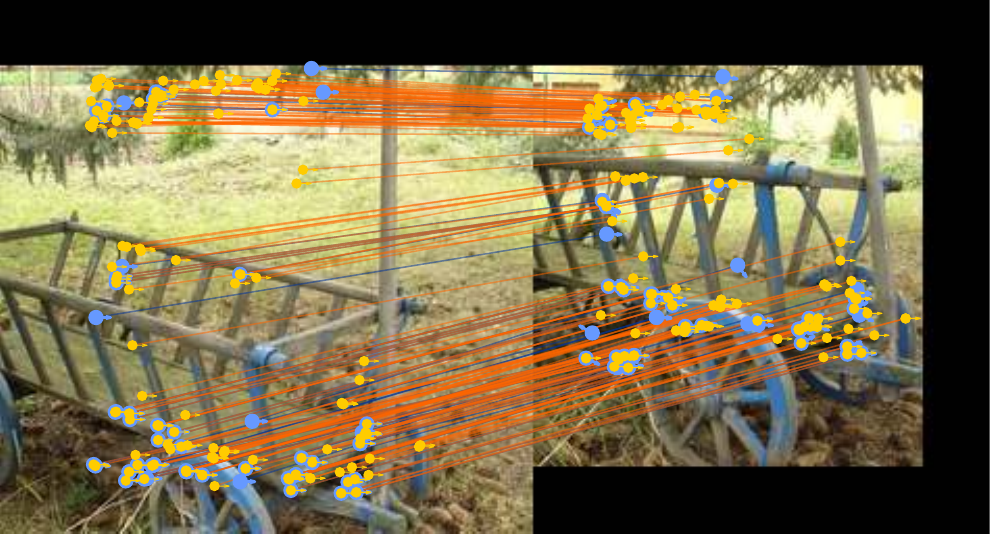}
	\caption{\label{base_match}
	A block pair $(b^1_w,b^2_{w'})$. Keypoints of the match sets $\tilde{\mathcal{M}}_l$ and $\mathcal{M}_l$ are shown in light blue and yellow, respectively, with corresponding matches highlighted in blue and orange, respectively. The light blue arrows indicate the keypoint patch orientations $\rho$ for $\tilde{\mathcal{M}}_l$ which is forced to the plane global orientation $\theta^\star_l$ in $\mathcal{M}_l$, indicated as the superimposed yellow arrows (see Secs.~\ref{multiscale_block} and~\ref{block_match}). Best viewed in color and zoomed in.}\vspace{-1em}
\end{figure}

Following~\cite{sgloh2}, in order to improve the robustness of the planar estimation, the global orientation $\theta^\star_l$ of each plane is computed by building a histogram with $m_\theta=16$ bins representing the relative orientations $\rho=\theta'-\theta$ of the corresponding patches in $\tilde{\mathcal{M}}_l$, and assigning to $\theta^\star_l$ the angle corresponding to the maximum bin. The Hz$^+$ pipeline plus RANSAC is executed again, but removing OriNet so that $\theta=0$ for all the patches $p\in I_1$ and $\theta'=\theta^\star_l$ for all the patches $p'\in I_2$, providing for each block pair $(b^1_w,b^2_{w'})$ the definitive set of planar matches $\mathcal{M}_l=\mathcal{M}_{ww'}$, together with the associated planar homography $H_l=H_{ww'}\in\mathbb{R}^{3\times 3}$. This step is illustrated in Fig.~\ref{base_match}.

For the next steps of Slime the reference system are the whole images $I_k$ and no more the specific block pairs. The same keypoint can be detected on different overlapping blocks but with a slight variation in position and it can appear almost duplicated in the image. Nevertheless, duplicates are considered independently as they act as weights for the specific match region.\vspace{-0.5em}

\subsection{Plane expansion and match pruning}\label{exp_prune}
The set $\mathcal{P}_{l}=\mathcal{P}_{ww'}$ of the matches of any block pair compatible with a given homography $H_l$ is then computed. The same match can be included in more planes. Given a match $(p,p')\in\bigcup\mathcal{M}_{z}$, where the index $z$ spans over all the allowed block pair indexes $l=(w,w')$, $(p,p')$ is included in ${P}_{l}$ if all the followed constraints are satisfied:
\subsubsection{Reprojection error} The maximum reprojection error for the match is bounded, i.e. $\epsilon^{H_l}<t_\perp$ as in Sec.~\ref{block_match}.
\subsubsection{Plane side} $H_l$ preserve the convex hull for $(p,p')\bigcup\mathcal{M}_l$ so that the point to be included is in the right side of the plane. This can be verified by checking the sign of the last coordinate of the reprojected homogeneous point $H_l\mathbf{x}$~\cite{multiview}.
\subsubsection{Relative orientation} The patch orientation after reprojection is compatible with the plane. Providing the relative orientation $\rho=\theta'-\theta$ between the patches of $(p,p')$, for the points $\mathbf{n}$ in the local neighborhood of $\mathbf{x}$, a planar homography $H$ is consistent with the given patches if the relative orientation after reprojection
\begin{equation}
	\rho^H_\mathbf{n}=\arctan(H\mathbf{n}-H\mathbf{x})-\arctan(\mathbf{n}-\mathbf{x})	
\end{equation}
is almost equal to $\rho$. Denoting by $\mathcal{N}$ the 4-connected neighborhood of the keypoint $\mathbf{x}$
\begin{equation}
	\mathcal{N}=\{\mathbf{x}\pm\left[1\;0\right]^T\cup\mathbf{x}\pm\left[0\;1\right]^T\}
\end{equation}
a match is considered feasible if $\exists\mathbf{n}\in\mathcal{N}$ such that
\begin{equation}\label{roteq}
	|\rho-\rho^{H_l}_\mathbf{n}|<t_\theta	
\end{equation}
where $t_\theta=\frac{3}{8}\pi=67.5^\circ$. Note that the above equation must take into account the cyclic nature of rotation angles. The inverse mapping considering $\mathbf{x}'$ and $H^{-1}_l$ is also evaluated, and it is sufficient that among the eight possibilities (four for $H_l$ and four for $H^{-1}_l)$ only one is satisfied.
\subsubsection{Relative scale} 
\begin{figure}[t]
	\center\vspace{-1em}
	\subfloat[\small $(b^1_w,b^2_{w'})$]{\label{plane_exp}
		\includegraphics[width=0.14\textwidth]{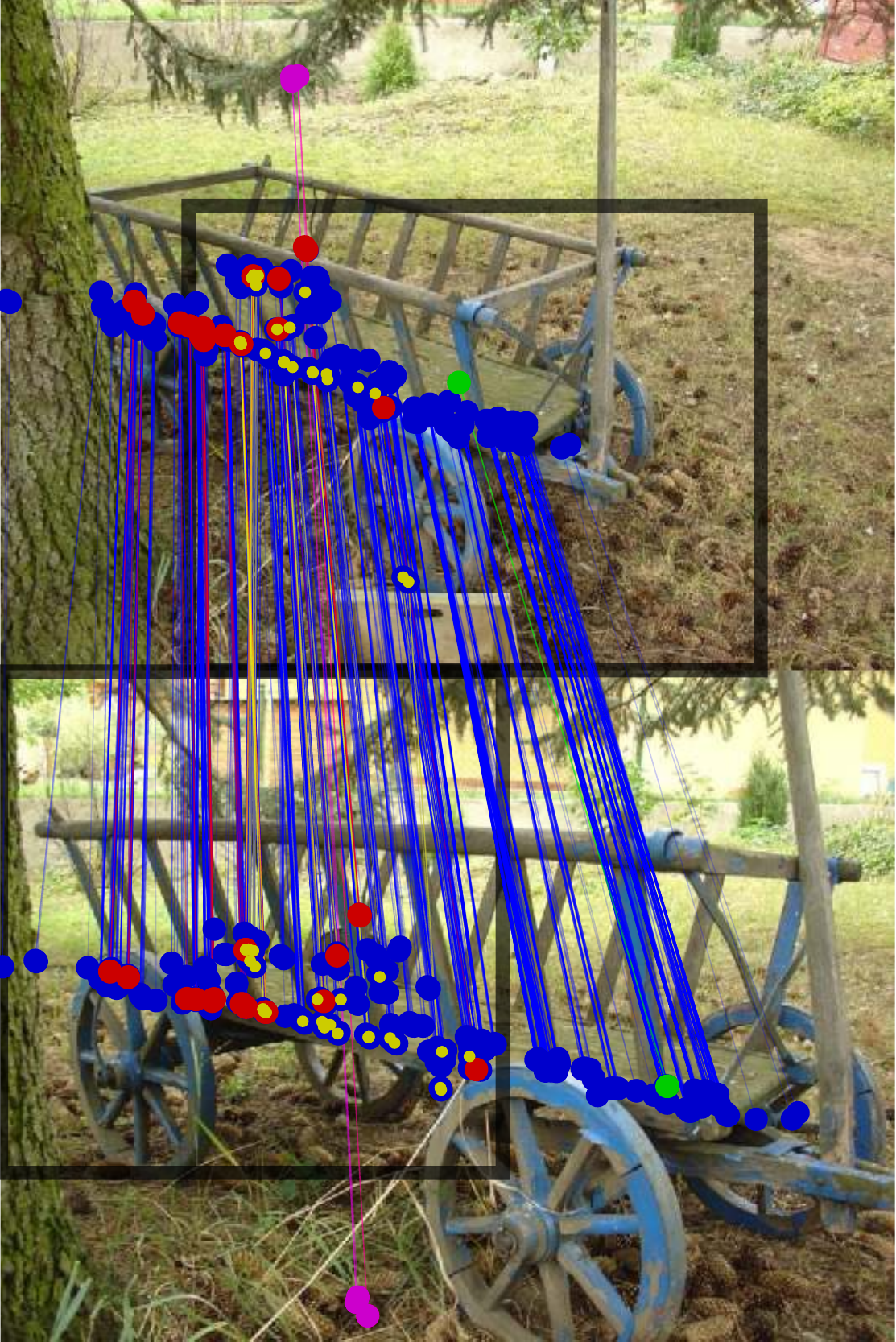}
		\includegraphics[width=0.14\textwidth]{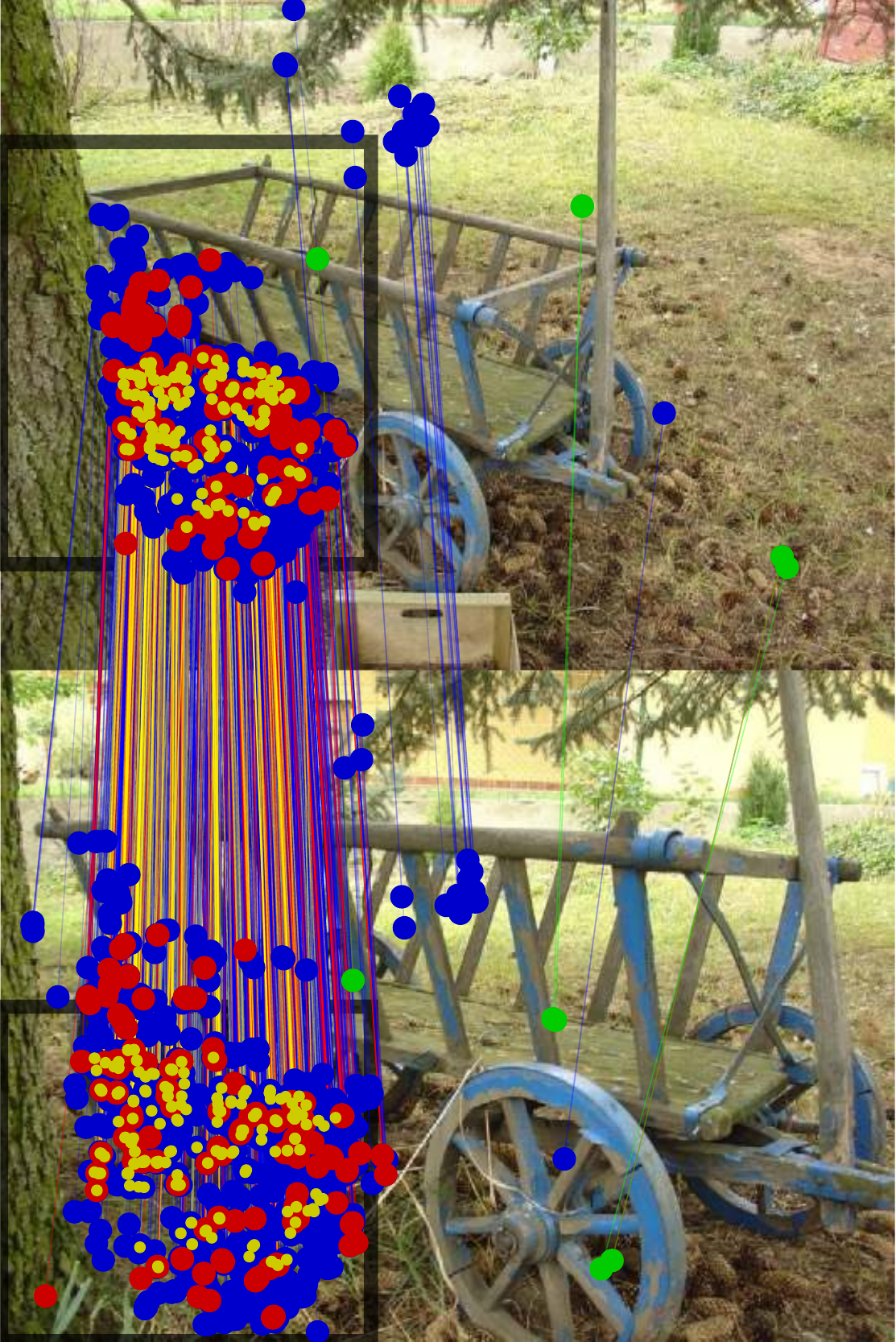}		
	}
	\hfil
	\subfloat[\small $(t^1_c,t^2_{c'})$]{\label{tile_best}
		\includegraphics[width=0.14\textwidth]{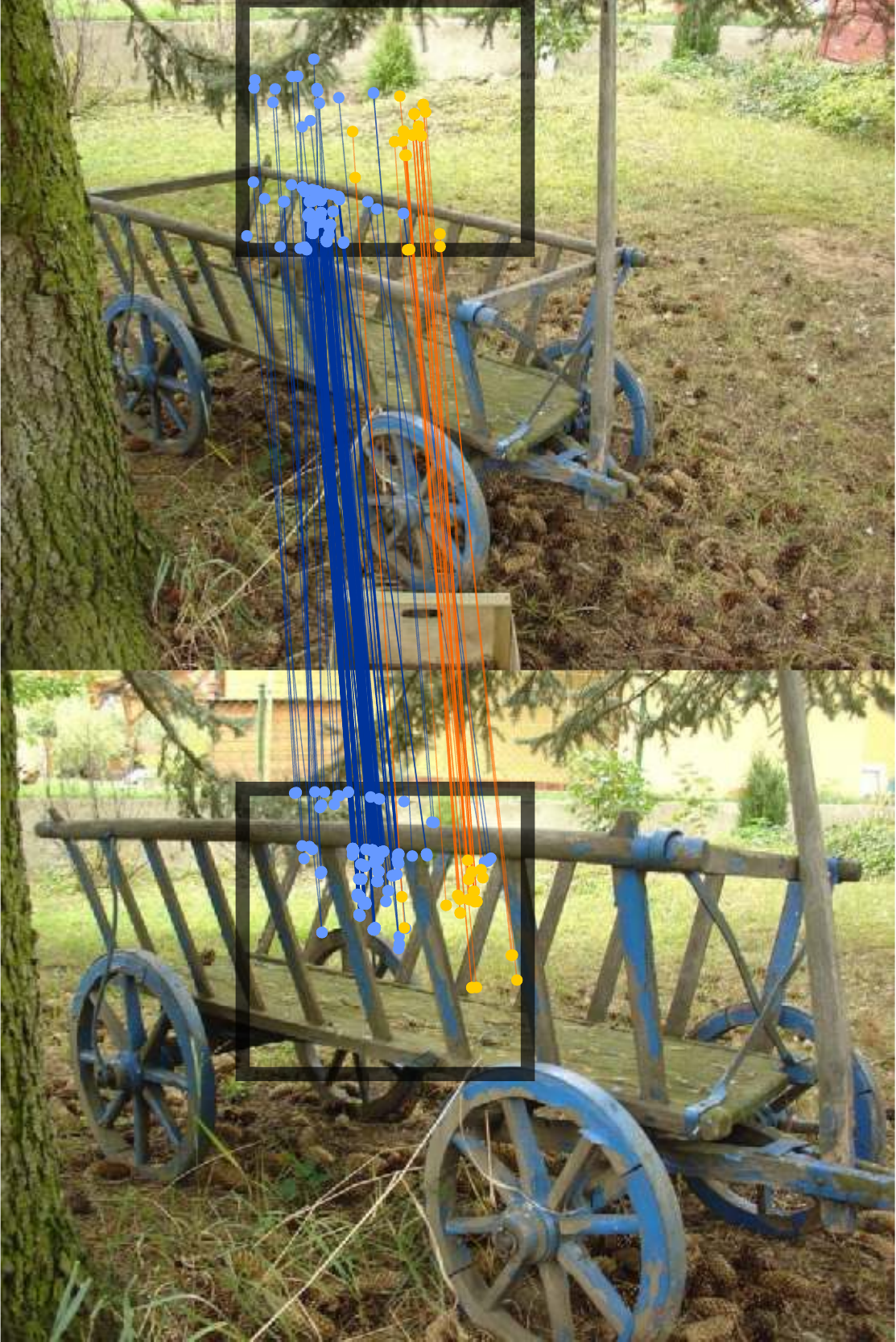}
	}
	\caption{\label{plane_exp_tile_best}
	\protect\subref{plane_exp} Match expansion and pruning of block pairs $(b^1_w,b^2_{w'})$ according to their homography $H_l$. Each column shows a different block pair, where blocks on the images are delimited by the black rectangles. The initial set of matches $\tilde{\mathcal{M}}_l$ is indicated in yellow by small markers, while matches in $\mathcal{P}_{l}$ are in blue. Matches removed according to the plane side, relative rotation and scale checks are in purple, red and green, respectively. Non-yellow matches are expanded matches, i.e. those satisfying the reprojection error according to $H_l$ (see Sec.~\ref{exp_prune}). \protect\subref{tile_best} A tile pair $(t^1_c,t^2_{c'})$, with tiles delimited by black rectangles.  The first and second best match set $\mathcal{H}^{1^\star}_v$ and $\mathcal{H}^{2^\star}_v$ for the given tile pair are in blue and orange, respectively (see Sec.~\ref{match_tile}). Best viewed in color and zoomed in.}\vspace{-1em}
\end{figure}

Similarly to the relative orientation, the relative scale $\psi=\frac{\sigma'}{\sigma}$ between the patches must be roughly preserved within the reprojection. As $\forall\mathbf{n}\in\mathcal{N}\Rightarrow\parallel\mathbf{n}-\mathbf{x}\parallel=1$ by definition, the relative scale after reprojection by $H$ is
\begin{equation}
	\psi^H_\mathbf{n}=\frac{\parallel H\mathbf{n}-H\mathbf{x}\parallel}{\parallel\mathbf{n}-\mathbf{x}\parallel}=\parallel H\mathbf{n}-H\mathbf{x}\parallel	
\end{equation}
In analogous way to the orientations, the compatibility for the scale is verified by
\begin{equation}
	\psi^{H_l}_\mathbf{n}\in\left[\frac{1}{t_\sigma},t_\sigma\right]	
\end{equation}
with $t_\sigma=3$ for at least one point $\mathbf{n}\in\mathcal{N}$ or for the inverse reprojection. Error thresholds $t_\theta$ and $t_\sigma$ are relatively high as the patch approximates a similarity while the actual local transformation is intended to be roughly a homography. Fig.~\ref{plane_exp} shows examples of both the expanded and refined matches.

\vspace{-0.5em}
\subsection{Selection of the best planes on image tiles}\label{match_tile}
\begin{figure}[t]
	\center\vspace{-1em}
	\subfloat[\small $\mathcal{H}^{q^\star}_{c\underline{c}'}$]{\label{t43}
		\includegraphics[width=0.14\textwidth]{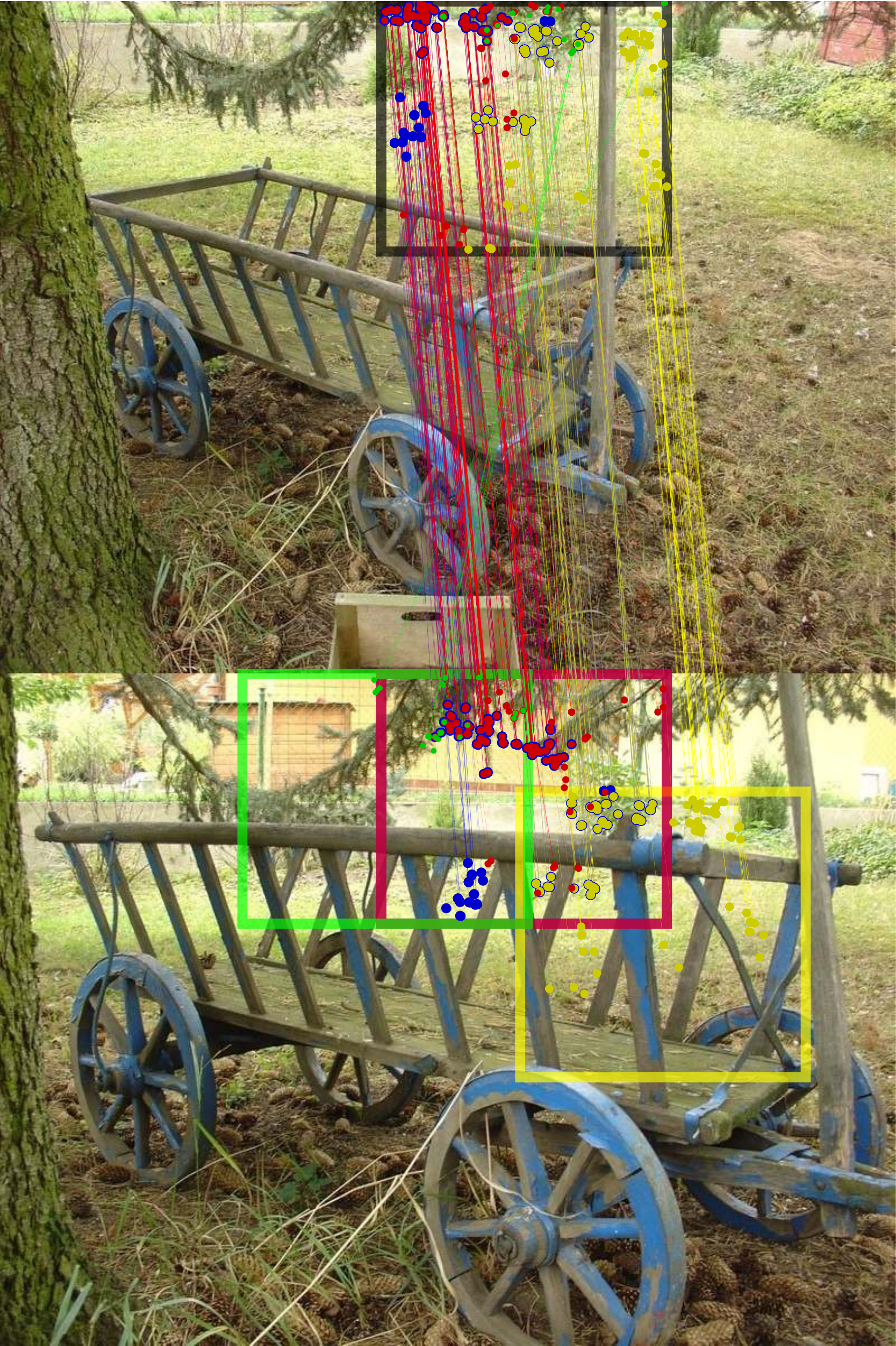}
	}
	\subfloat[\small $\mathcal{H}^{q^\star}_{\underline{c}c'}$]{\label{t44}
		\includegraphics[width=0.14\textwidth]{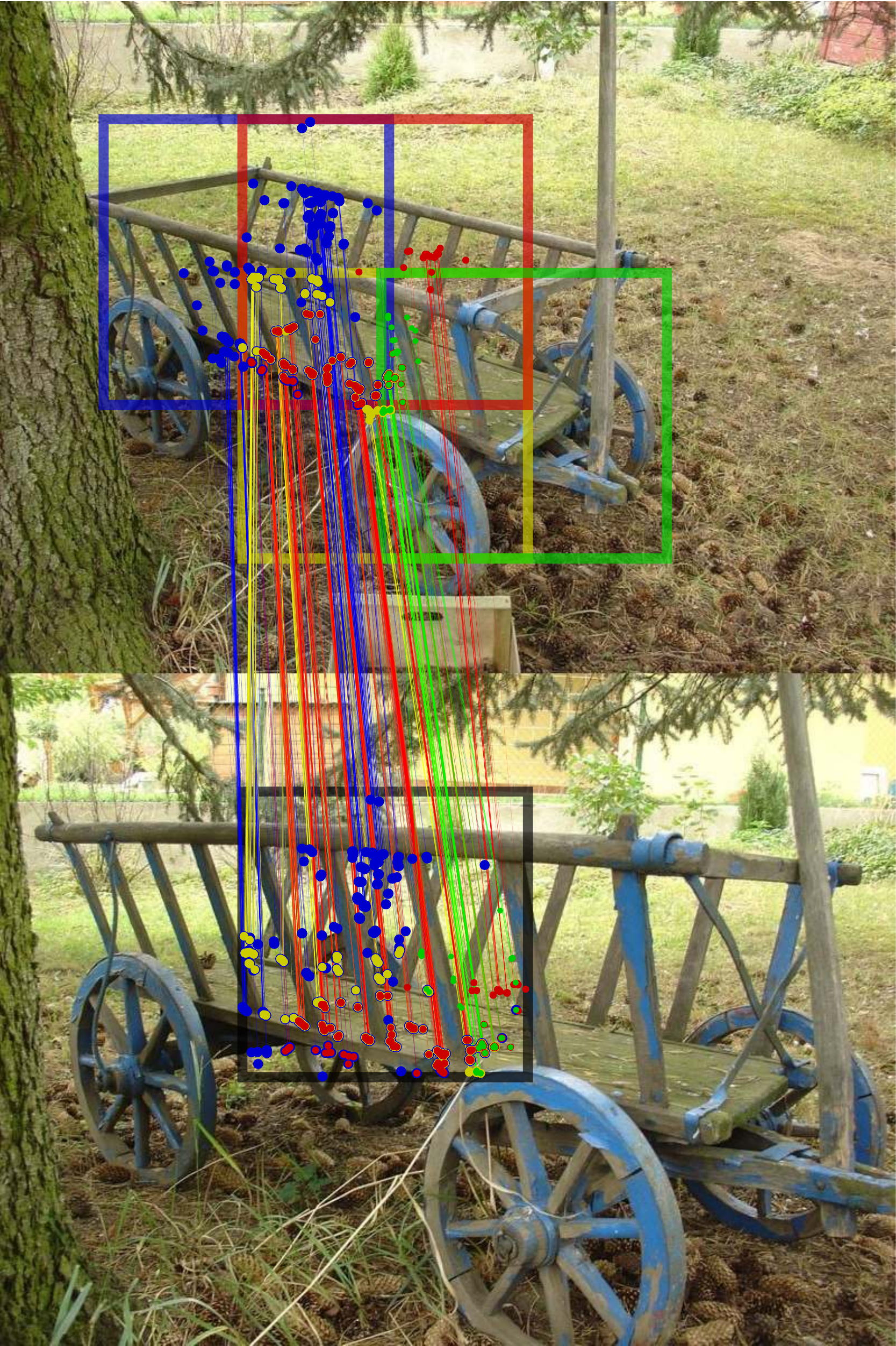}
	}
	\subfloat[\small $\mathcal{H}$]{\label{r1}
			\includegraphics[width=0.14\textwidth]{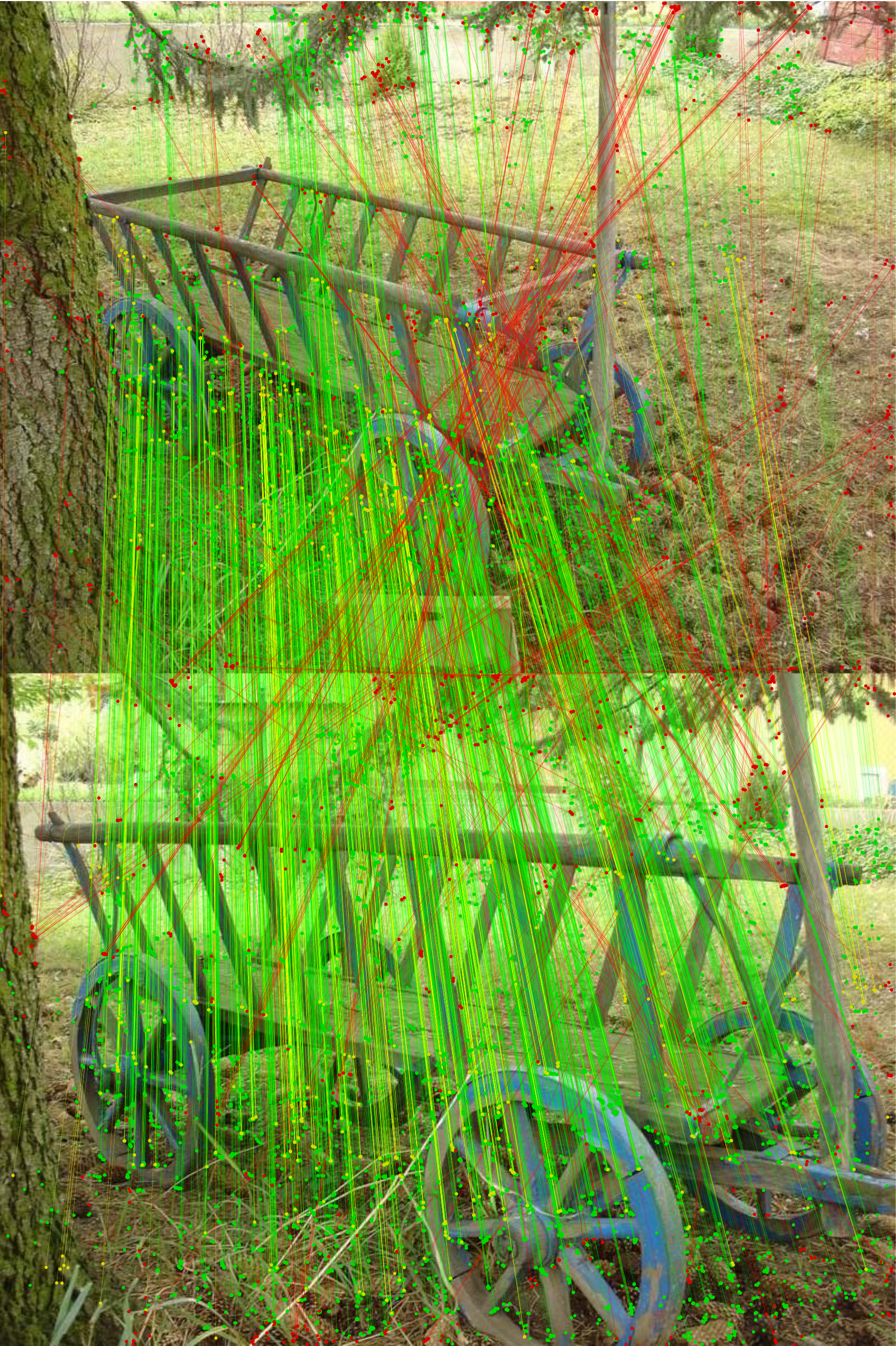}
	}
	\caption{\label{4tile_best}
	Examples of match sets \protect\subref{t43} $\mathcal{H}^{q^\star}_{c\underline{c}'}$ and \protect\subref{t44} $\mathcal{H}^{q^\star}_{\underline{c}c'}$, i.e. the four best match sets for a given tile in one of the two images, delimited by a black rectangle, among all the tiles in the other image. The rank index $q=1\ldots,4$ is indicated in blue, yellow, red and green, respectively (see Sec.~\ref{match_tile}). \protect\subref{r1} Matches in the superset $\mathcal{H}$ (green, yellow and red altogether) as the union of the four best match sets for each tile in both images, are successively filtered by the pairwise plane epipolar check to get $\mathcal{F}$ (green and yellow only) and greedy to obtain $\overline{\mathcal{F}}$ (green), which contains only two homographies per tile (see Sec.~\ref{epi_filt_sec}). Best viewed in color and zoomed in.}\vspace{-1em}
\end{figure}

The original images are divided also into overlapping blocks with the same size, stride and padding constraints employed in Sec.~\ref{multiscale_block}. These kind of blocks are named tiles to distinguish them from the previous blocks, as no scale factor is applied. Following an analogous notation, a generic tile for the $I_k$ image is indicated as $t^k_{ab}=t^k_c$ and the set of tiles as $\mathcal{T}_k=\{t^k_{ab}\}$. For a pair of tiles $(t^1_c,t^2_{c'})\in\mathcal{T}_1\times\mathcal{T}_2$ the set
\begin{equation}\label{tile_set}
	\mathcal{H}^l_v=\mathcal{H}^l_{cc'}=\{(p,p')\in\mathcal{P}_l:\mathbf{x}\in t^1_c, \mathbf{x}'\in t^2_{c'},\epsilon^{H_l}<t_\perp\}	
\end{equation}
defines the block matches on the two tiles compatible with the homography $H_l$, i.e. the matched keypoints lie onto the tiles $t^1_c$ and $t^2_{c'}$, where the linear index $v$ indicates the index pair $(c,c')$, and the reprojection error according to the homography is bounded by $t_\perp$. Successive filtering inspired by the extended descriptor matching strategy defined in~\cite{sgloh2} are then applied to get the best candidate sets of matches $\mathcal{H}^l_v$. The $q$-th best homography match set for a pair of tiles is found so as to maximize the coverage of the keypoints for that tiles. This is defined recursively as
\begin{equation}
	\mathcal{H}^{q^\star}_v=
		\begin{cases}
			\mathrm{argmax}_{\mathcal{H}^l_v,\forall l}|\mathcal{H}^l_v| & \text{if}\;q=1\\
			\mathrm{argmax}_{\mathcal{H}^l_v,\forall l}|\mathcal{H}^l_v-\bigcup_{q'=1}^{q-1}\mathcal{H}^{q'^\star}_v| &\text{otherwise}
		\end{cases}
\end{equation}
By this definition $\mathcal{H}^{1^\star}_v$ contains the matches associated with the planar homography $H_l$ that has more matches than others for the tile pair $v=(c,c')$, while the second best set $\mathcal{H}^{2^\star}_v$ includes, with respect to the remaining sets, more matches not already covered by the best set $\mathcal{H}^{1^\star}_v$. An example is shown in Fig.~\ref{tile_best}. For each tile pairs the best two homography sets are collected into the superset $\mathcal{H}^\star$, i.e.
\begin{equation}
	\mathcal{H}^\star=\{\mathcal{H}^{q^\star}_v:\forall v,q=1,2\}
\end{equation}
under the assumption that the scene can by locally modeled by two rough planes, one for the background and another for the foreground. Nevertheless, the current tile association is many-to-many and more than two homographies can be related to a tile. The $\mathcal{H}^\star$ is hence filtered so that at maximum four homographies are provided for a given tile. Defining 
\begin{equation}
\hspace{-0.5em}
\resizebox{0.95\hsize}{!}{
$
	\mathcal{H}^{q^\star}_{c\underline{c}'}=
	\begin{cases}
		\mathrm{argmax}_{\mathcal{H}^l_{cc'}\in\mathcal{H}^\star,\forall(l,c')}|\mathcal{H}^l_{cc'}| & \text{if}\;q=1\\
		\mathrm{argmax}_{\mathcal{H}^l_{cc'}\in\mathcal{H}^\star,\forall(l,c')}|\mathcal{H}^l_{cc'}-\bigcup_{q'=1}^{q-1}\mathcal{H}^{q'^\star}_{c\underline{c}'}| & \text{otherwise}
	\end{cases}
$	
}
\hspace{-1.5em}
\end{equation}
so that the best sets are searched on $I_2$ by spanning on $c'$ and similarly for $I_2$ by $\mathcal{H}^{q^\star}_{\underline{c}c'}$, the superset of homograhy matches $\mathcal{H}$ collects the four best tile sets for each image as 
\begin{equation}
	\mathcal{H}=\{\mathcal{H}^{q^\star}_{c\underline{c}'},\mathcal{H}^{q^\star}_{\underline{c}c'}:\forall v, q=1,\ldots,4\}
\end{equation}
Examples of match sets $\mathcal{H}^{q^\star}_{c\underline{c}'}$ and $\mathcal{H}^{q^\star}_{\underline{c}c'}$ for given tiles $t^1_c$ and $t^2_{c'}$ and of the superset $\mathcal{H}$ are shown in Fig.~\ref{4tile_best}. 

\vspace{-0.5em}
\subsection{Epipolar geometry from plane consistency}\label{epi_filt_sec}
As each match set $\mathcal{H}^l_v\in\mathcal{H}$ defines a homography, the union of matches of a pair of them can be used to get the underlining fundamental matrix of the scene or to prune matches. Let $(\mathcal{H}_d,\mathcal{H}_{d'})\in\mathcal{H}\times\mathcal{H},d\ne d'$ be a pair of tile match sets, and $F_{dd'}$ the fundamental matrix derived by the normalized eight-point algorithm~\cite{multiview} on the correspondences $(\mathbf{x},\mathbf{x}')\in\mathcal{H}_d\cup\mathcal{H}_{d'}$. The maximum epipolar distance for a match $(p,p')$ according to $F$ is
\begin{equation}\label{max_epi_error}
	\xi^F=\max\left(\frac{\mathbf{x}'^TF\mathbf{x}}{(F\mathbf{x})^2_1+(F\mathbf{x})^2_2},\frac{\mathbf{x}^TF^T\mathbf{x}'}{(F^T\mathbf{x}')^2_1+(F^T\mathbf{x}')^2_2}\right)
\end{equation}
where the subscript $n$ in the denominators indicates the $n$-th element of the argument vector. The $F_{dd'}$ fundamental matrices retain a match $(p,p')$ if
\begin{equation}
	\left(\sum_{F_{dd'}}\text{\textlbrackdbl}\xi^{F_{dd'}}< t_\perp\text{\textrbrackdbl}\right) >t_\xi |\mathcal{H}|
\end{equation}
where \textlbrackdbl$\cdot$\textrbrackdbl\phantom{,} means for the indicator function and $t_\xi=9$. The idea is that many planes can be acceptable and the derived fundamental matrices are roughly close to the correct one. The threshold $t_\xi |\mathcal{H}|$ is chosen since a match $(p,p')\in\mathcal{H}_d$ is an inlier for all the $F_{dd'}$ derived by it, which are $|\mathcal{H}|$ minus 1, and it is expected that there are about 9 similar homographies due to block overlaps, so that a match passes the check approximately at least 9 times the cardinality of $\mathcal{H}$. The set of all filtered matches by the fundamental matrix is denoted by $\mathcal{M}^+$ and the filtered match set of $\mathcal{H}^l_v$ is
\begin{equation}
	\mathcal{F}^l_v = \mathcal{H}^l_v \cap \mathcal{M}^+
\end{equation}
This filter makes a partion of the model space like J-linkage~\cite{j_linkage}, but the proposed filter excludes individual inconsistent matches in contrast to J-linkage or RANSAC which look  for the model that better fit data. 

The best mutual homography match sets are then recovered greedly from $\mathcal{F}$ such that no tiles has no more than two homographies, the background and foreground planes, and those with high cardinality are preferred. The match set $\mathcal{H}_d\in\mathcal{F}$ are sorted by their cardinality and in turn added to the superset $\overline{\mathcal{F}}$ if the addition does not violate the constraint that at maximum only two sets in $\overline{\mathcal{F}}$ can share the same tile. The obtained match sets are shown Fig.~\ref{r1}.

\vspace{-0.5em}
\subsection{Plane fusion and match refinement by global orientation}\label{global_ori}
For each $\mathcal{F}^l_v\in\overline{\mathcal{F}}$, $\mathcal{F}^{l^\star}_v$ is derived as the set union 
\begin{equation}
	\mathcal{F}^{l^\star}_v=\bigcup\,_{\forall l',\frac{|\mathcal{F}^l_v|\cap|\mathcal{F}^{l'}_v|}{|\mathcal{F}^l_v|\cup|\mathcal{F}^{l'}_v|}>t_{ov}}\mathcal{F}^{l'}_v
\end{equation}
where $l'$ spans on the homography set pair having an overlap, in terms of cardinality of matches shared with $\mathcal{H}^l_v$, greater than $t_{ov}=\frac{1}{3}$, see Fig.~\ref{r3}. Due to the overlap between blocks, this acts as a plane fusion and a local smoothing.

As for Sec.~\ref{block_match}, the global relative orientation for the set of matches $(p,p')\in\mathcal{F}^{l^\star}_v$ is estimated, denoted as $\theta^l_v$. The global relative orientation $\theta^+$ between the images is computed considering the corresponding bin with the maximum value on the set of the collected plane orientations $\{\theta^l_v\}$. Planes having inconsistent orientations with respect to $\theta^+$ are removed obtaining the superset
\begin{equation}
	\mathcal{F}^\theta=\{\mathcal{F}^{l^\star}_v:\forall(l,v),|\theta^+-\theta^l_v|<t_{\theta^+},\mathcal{F}^l_v\in\overline{\mathcal{F}}\}
\end{equation}
where $t_{\theta^+}=\frac{\pi}{4}=45^\circ$, i.e. the relative rotation of the patches given by the plane cannot differ more than $45^\circ$ from the global rotation $\theta^+$ between the images. Fig.~\ref{r2} shows the result of the above steps. 


\vspace{-0.5em}
\subsection{Tile matching refinement}\label{tile_ref}
\begin{figure}[t]
	\center
	\begin{minipage}{0.162\textwidth}		
		\subfloat[\small $\mathcal{F}^{l^\star}_v$]{\label{r3}
			\includegraphics[width=0.98\textwidth]{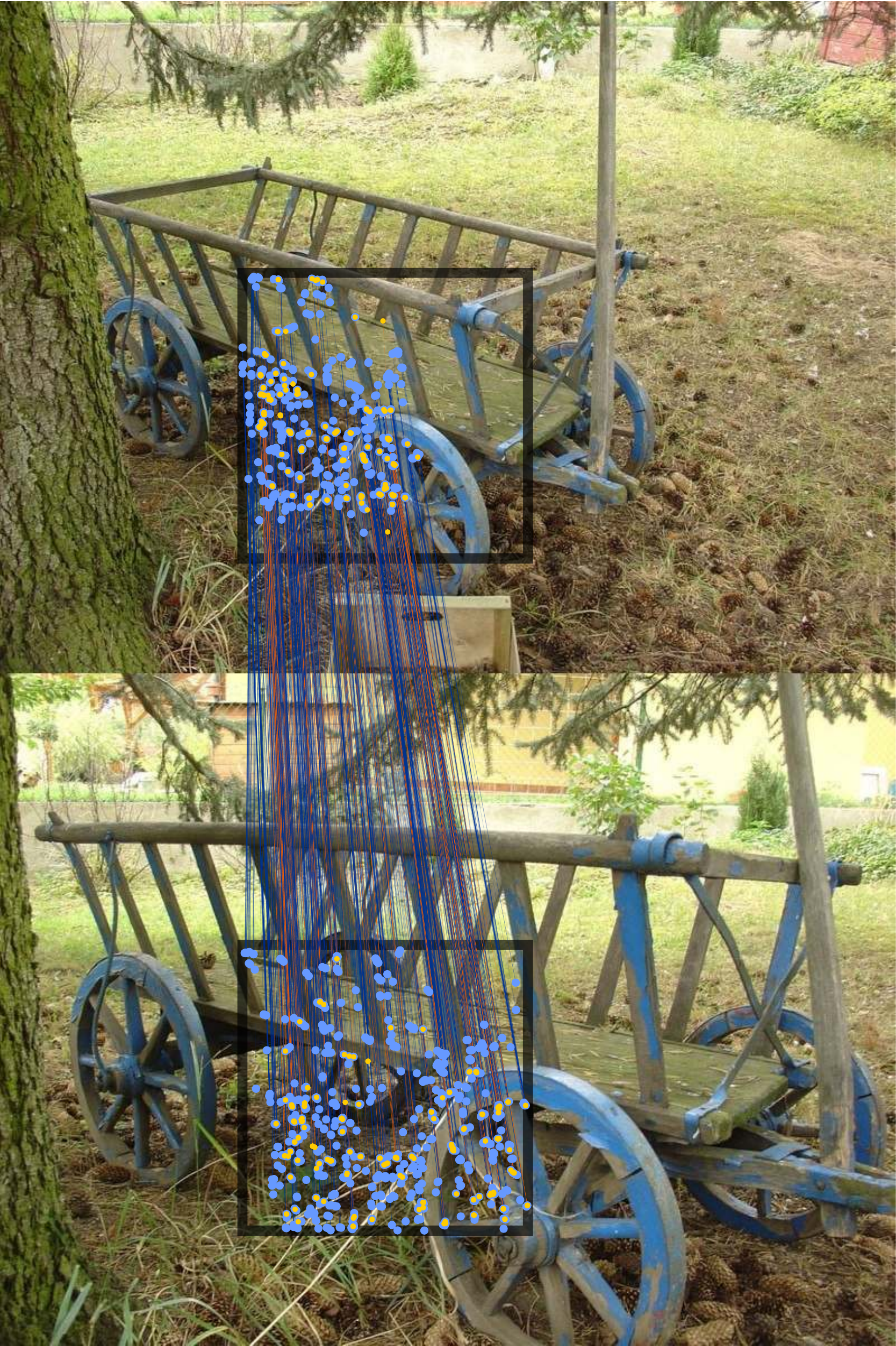}
		}
	\end{minipage}
	\begin{minipage}{0.162\textwidth}	
		\subfloat[\small $\mathcal{F}^\theta$]{\label{r2}
			\includegraphics[width=0.98\textwidth]{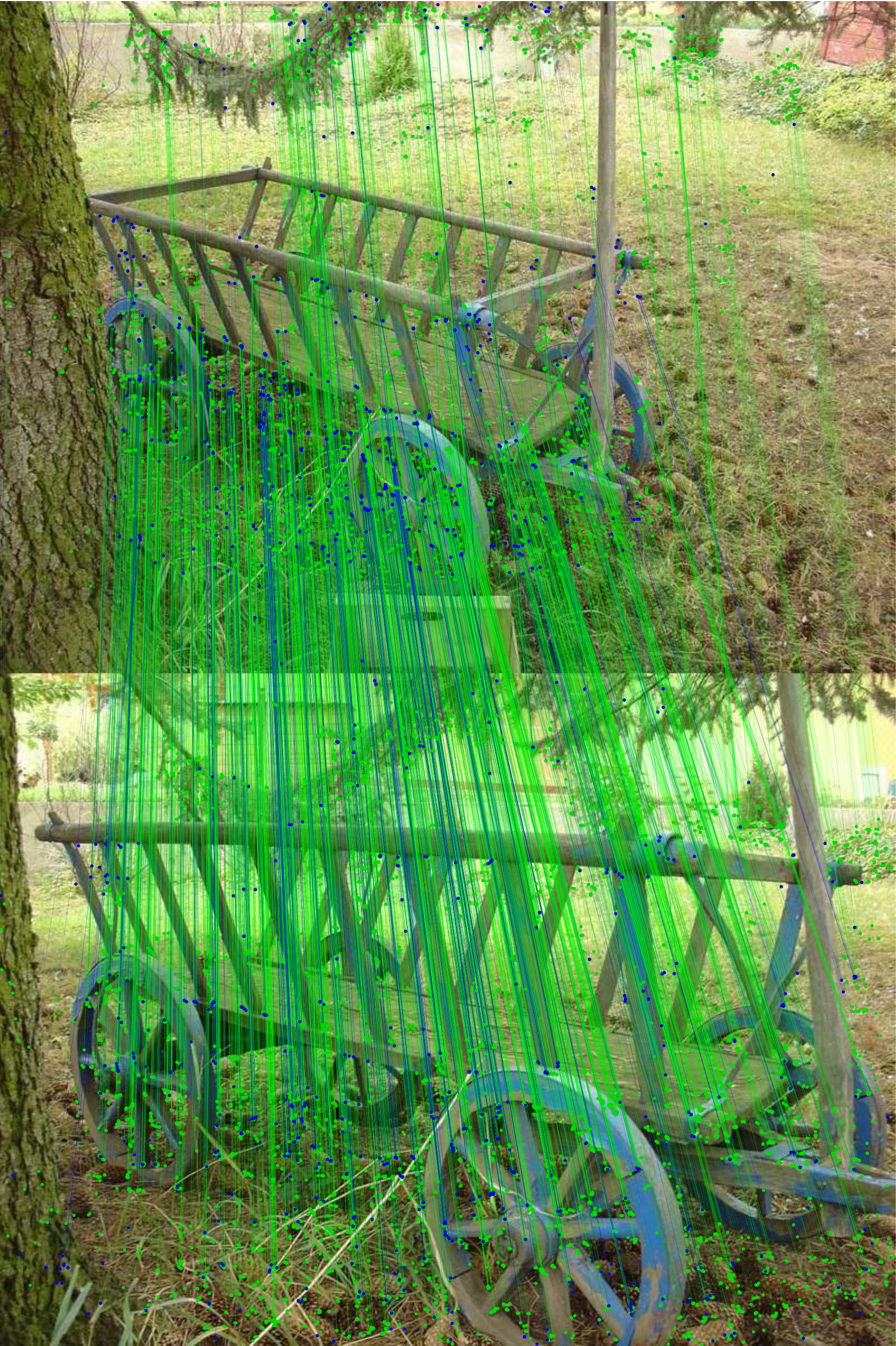}
		}
	\end{minipage}
	\begin{minipage}{0.151\textwidth}
		\subfloat[\small tile matches]{\label{f1}
			\includegraphics[width=0.82\textwidth]{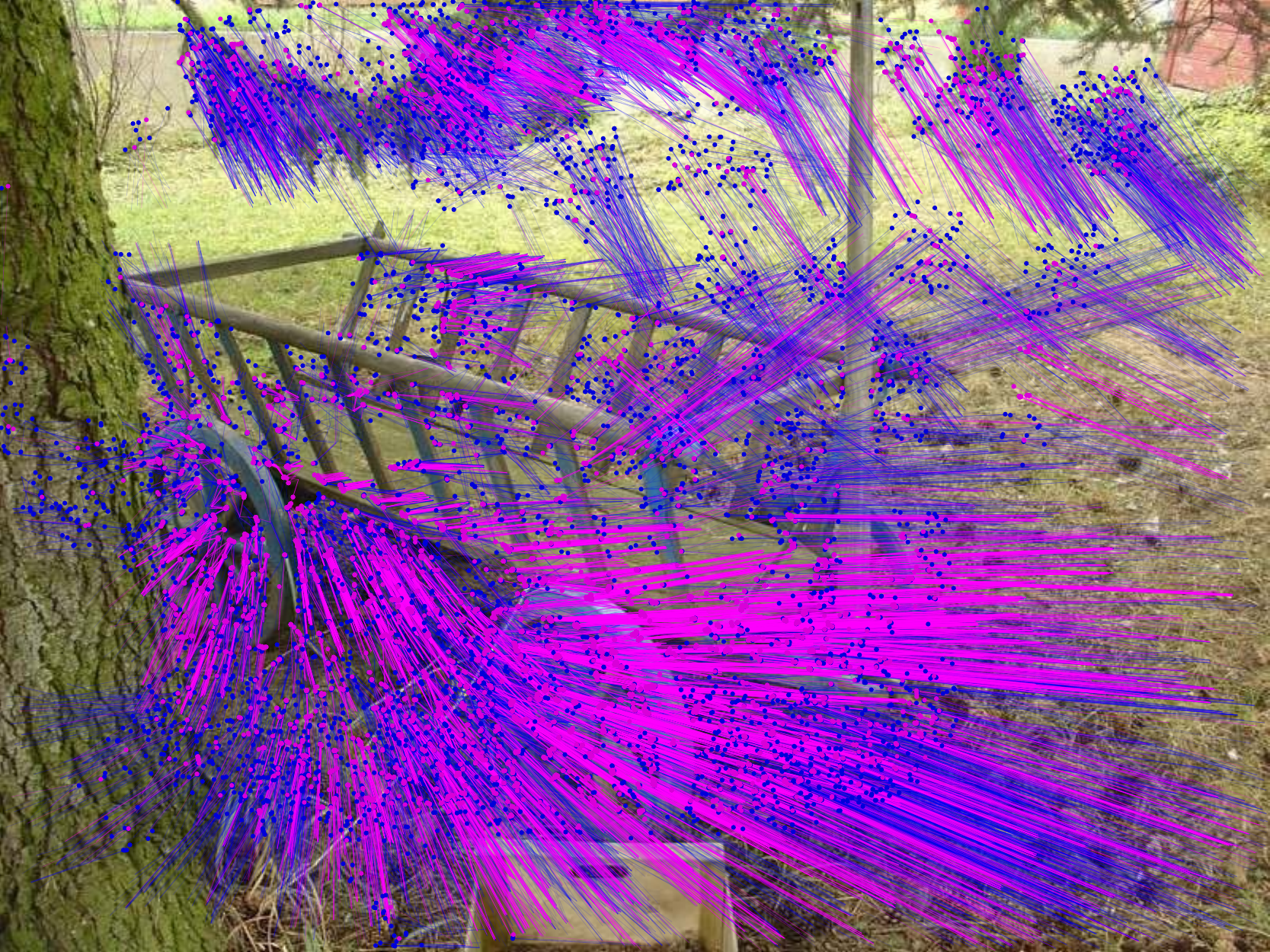}
		}\\
		\subfloat[\small global DTM]{\label{f2}
			\includegraphics[width=0.82\textwidth]{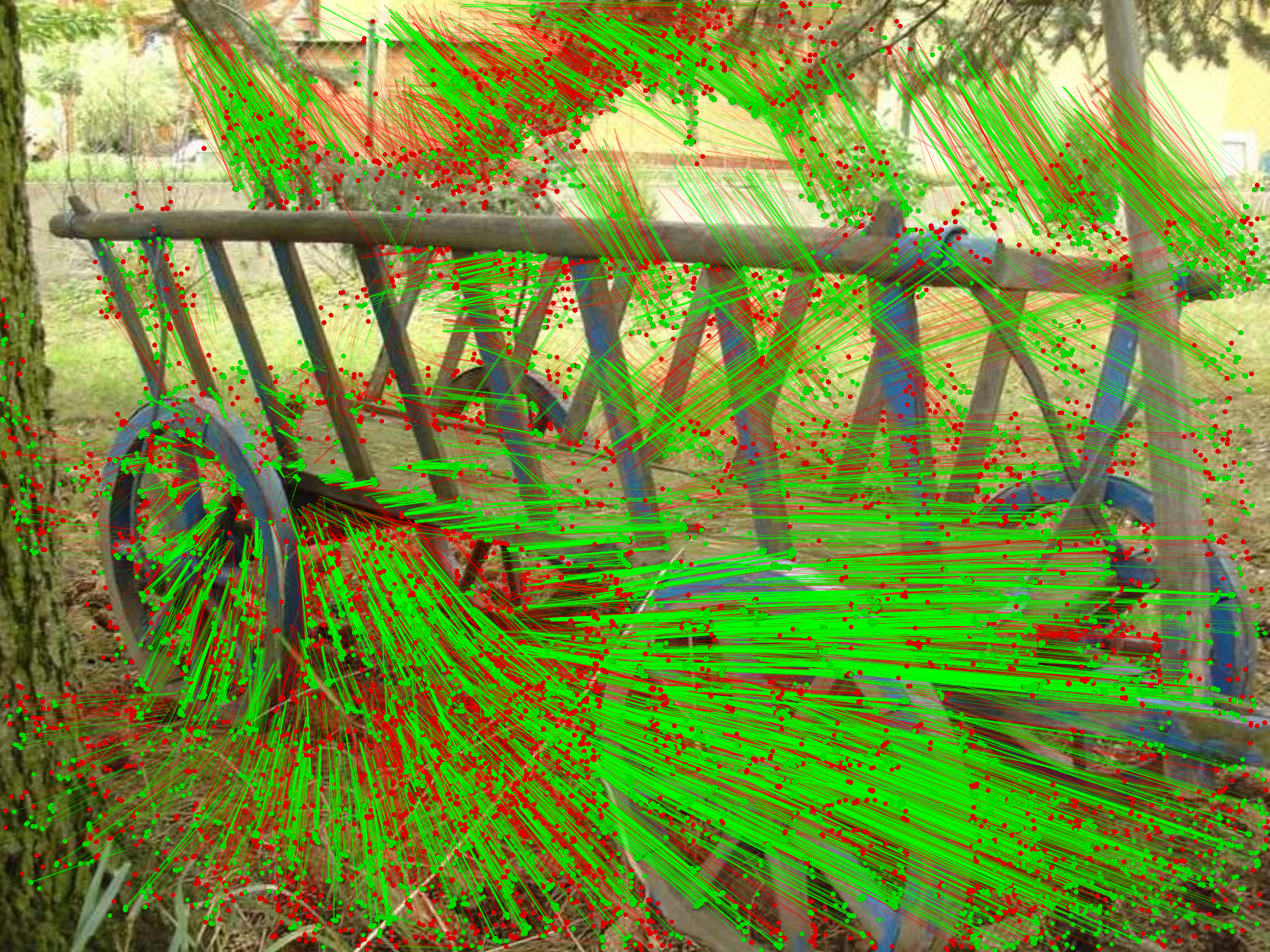}
		}
	\end{minipage}
	\caption{\label{epi_filt}
		\protect\subref{r3} A given match set within two tiles $\mathcal{F}^l_v$ (orange only) is merged with overlapping match sets, obtaining $\mathcal{F}^{l^\star}_v$ (blue and orange altogether).~\protect\subref{r2} Matches in $\mathcal{F}^\theta$ (green and blue altogether) obtained from $\mathcal{F}$ (green only) after plane fusion and discarding matches with inconsistent relative orientation with respect to the global relative image rotation (see Sec.~\ref{global_ori}).~\protect\subref{f1} Optical flow of block matches $(p,p')\in\mathcal{F}^\theta$ (purple) plus tile matches $(q,q')$ (blue) obtained by rerunning the base pipeline on tiles with the constrained flow according to the original block matches.~\protect\subref{f2} Final (green, the same of Fig.\protect\ref{slime_cart}) and discarded (red) optical flow of the matches after running the global DTM (see Sec.~\ref{tile_ref}). Best viewed in color and zoomed in.}\vspace{-1em}
\end{figure}

In the previous steps patches $(p,p')$ originated from blocks which are at coarse resolutions (see Sec.~\ref{multiscale_block}). In this final step keypoints are extracted at the effective image resolution from tiles and matched. To differentiate from the patches extracted in Sec.~\ref{block_match}, the generic patch pair generated from tiles is indicated as $(q,q')$. For each $\mathcal{F}^{l^\star}_v=\mathcal{F}^{l^\star}_{cc'}\in\mathcal{F}^\theta$ the Hz$^+$ pipeline without RANSAC is executed on the tile pairs $(t^1_c,t^2_{c'})$, imposing for each patch pair $(q,q')$ a relative orientation $\theta^l_v=\theta^l_{cc'}$, i.e. by setting $\theta=0$ and $\theta'=\theta^l_v$ according to Sec.~\ref{block_match}. Furthermore, matches $(q,q')$ are removed before running DTM as last step of the pipeline if does not exist another match among those contained in $\mathcal{F}^\theta$ (see Sec.~\ref{global_ori}) for which the flow is similar, i.e. if $\nexists(p,p')\in\mathcal{F}^{l^\star}_v, \mathcal{F}^{l^\star}_v\in\mathcal{F}^\theta$, such that
\begin{equation}
	\max(\parallel\mathbf{x}-\mathbf{y}\parallel,\parallel\mathbf{x}'-\mathbf{y}'\parallel)<t_\perp
\end{equation} 
where $(\mathbf{y},\mathbf{y}')$ are the keypoints associated to the patch $(q,q')$, see Fig.~\ref{f1}. The final set of matches is given by running DTM globally on the union of matches $(q,q')$ extracted from tiles as above together with the matches $(p,p')$ of $\mathcal{F}^\theta$. Fig.~\ref{f2} shows the result of this last step.

\section{Evaluation}\label{evaluation}
\subsection{Compared methods}
The compared methods include RootSIFT+FGINN as handcrafted baseline, Key.Net+HardNet+AffNet+OriNet+ AdaLAM and Hz$^+$+AffNet+OriNet+HardNet+DTM as hybrid pipelines, and DISK, SuperGlue, LoFTR, MatchFormer, QuadTree Attention, ECO-TR and DKM as fully end-to-end deep matching networks. Pipelines are identified by their first component, e.g. RootSIFT+FGINN is referenced as RootSIFT only. Besides the baseline, all the above methods achieved SOTA results in recent contests, such as the Image Matching Challenges (IMC) of the last years\footnote[2]{\label{imclink}\url{https://www.cs.ubc.ca/research/image-matching-challenge/2021/} \url{https://www.kaggle.com/competitions/image-matching-challenge-2022}}.

In addition to the Hz$^+$ pipeline it was designed for, Slime was tested paired with RootSIFT and Key.Net to further investigate its properties. The term ``Slimed'' preceding a pipeline means the use of Slime with the specific base pipeline and, if not explicitly stated, ``Slime'' refers as default to Slimed Hz$^+$. Notice that only the descriptor module AffNet+OriNet+HardNet, i.e. the feature extraction, is deep for Hz$^+$, slimed or not.

For RootSIFT, Key.Net and Hz$^+$, their upright counterparts, indicated by the ``$\Rsh$'' superscript, are also included to investigate rotation invariance in matching, by avoiding dominant orientation estimation for SIFT and by removing OriNet for the hybrid pipelines. Among the end-to-end deep methods, only LoFTR has a rotation invariant extension, denoted as SE2-LoFTR, included in the comparison. About 5\% of the image pairs included are non-upright, so that the overall performances of a method are not affected by them, but the analysis can provide insights of the possible boosts or issues.

Results have been obtained with or without postponing RANSAC, defining inliers according to Eq.~\ref{max_reproj} and Eq.~\ref{max_epi_error}, respectively in the case of planar and non-planar scenes, and evaluating different inlier thresholds.

Further details about the setup of the compared methods, as well as the code and datasets employed, including the implemented image matching methods, are available in SM\ref{sm_methods} while RANSAC implementation details are reported as SM\ref{sm_ransac}.

\vspace{-0.5em}
\subsection{Datasets}\label{dataset_desc}
\subsubsection{Planar dataset}
This includes 26 different scenes with six images pair each, plus a further single-pair scene. Excluding this latter pair, within each scene one image is fixed as reference, so that five image pairs are used for the evaluation. There are in total $141$ image pairs, of which about $9\%$ are designed to do not work with upright methods. The thumbnail gallery of these scenes is shown in Fig.~\ref{pd}.
\subsubsection{Non-planar dataset}
This includes 92 different scenes with up to three image pair for each scene, for a total of 142 image pairs, plus four upright image pairs obtained from the original non-upright image pairs. Fig.~\ref{npd} shows the thumbnail gallery of the scene of the non-planar dataset. Overall, the variety of scenes provides several challenging scenarios.

Both datasets extend those proposed in~\cite{dtm}. Please refer to SM\ref{sm_p_np} for additional details.

\vspace{-0.5em}
\subsection{Error metrics}\label{err_metric}
Three error metric have been defined. The first two, named coverage and precision focus on the individual matches, while the last one, named accuracy, on the scene transformation. 
\subsubsection{Coverage}\label{covs}
The coverage definition extends the recall so that close keypoints count less to avoid misleading results in case of accumulated keypoints into clusters. The raw coverage of an image is computed as the area covered by the keypoints of the correct matches, see Fig.~\ref{recall}. The raw image coverage is then normalized by the area of all the valid points that can be mapped from one image to the other, and the final coverage is lastly computed as the minimum of the normalized coverages between the two images.
\subsubsection{Precision}
The precision follows the standard definition, i.e. it is the percentage of the correct matches of a method.
\subsubsection{Accuracy}\label{accs}
\begin{figure}[t!]
	\center
	\resizebox{0.39\textwidth}{!}{	
		\begin{tabular}{rccc}
			\rotatebox{90}{\hspace{2.2em}planar} &	
			\includegraphics[width=0.0116\textwidth]{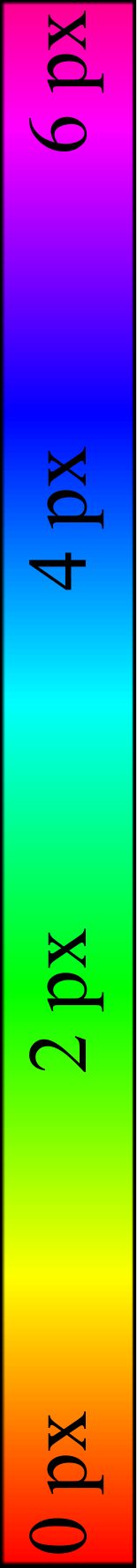} &
			\includegraphics[width=0.18\textwidth]{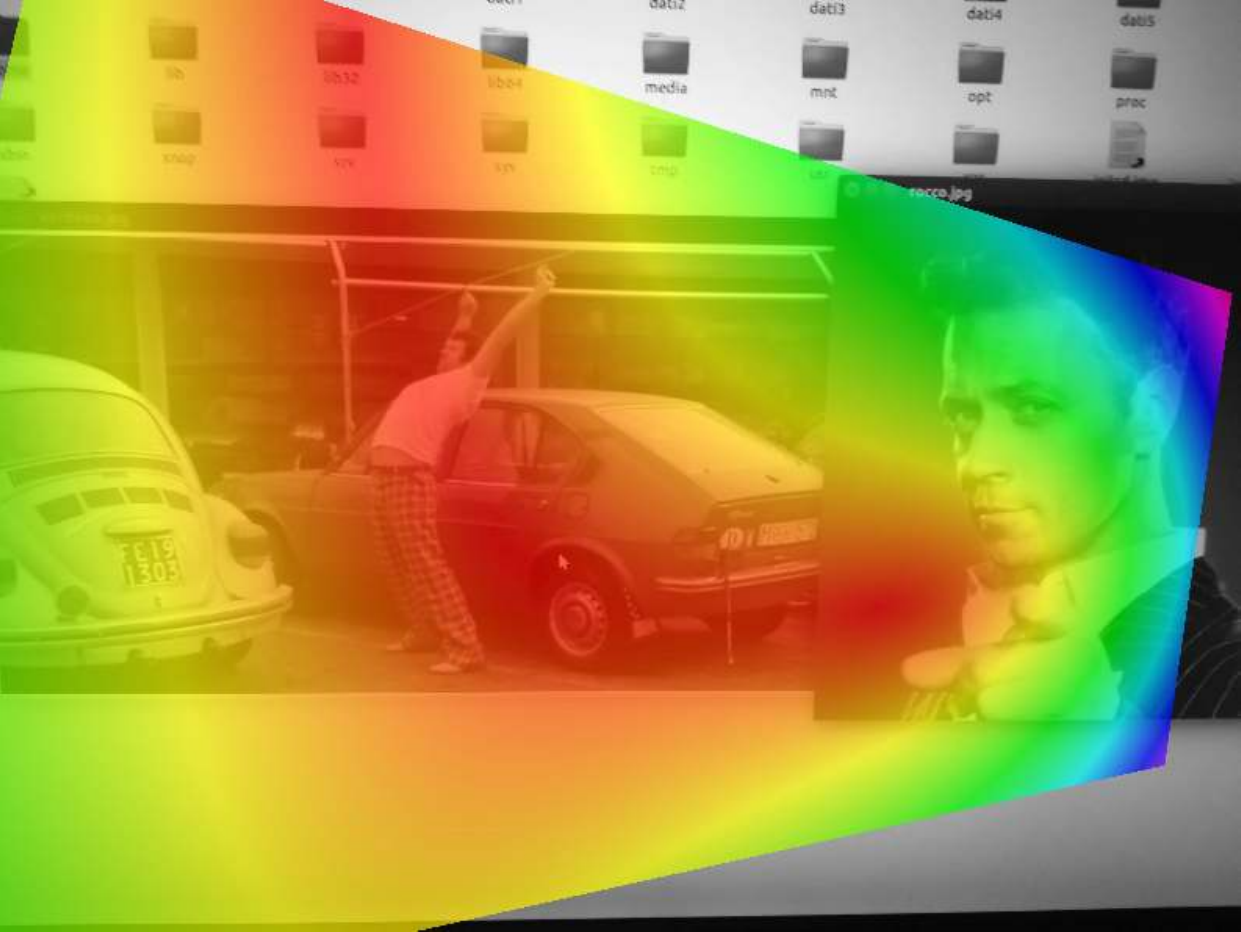} &
			\includegraphics[width=0.18\textwidth]{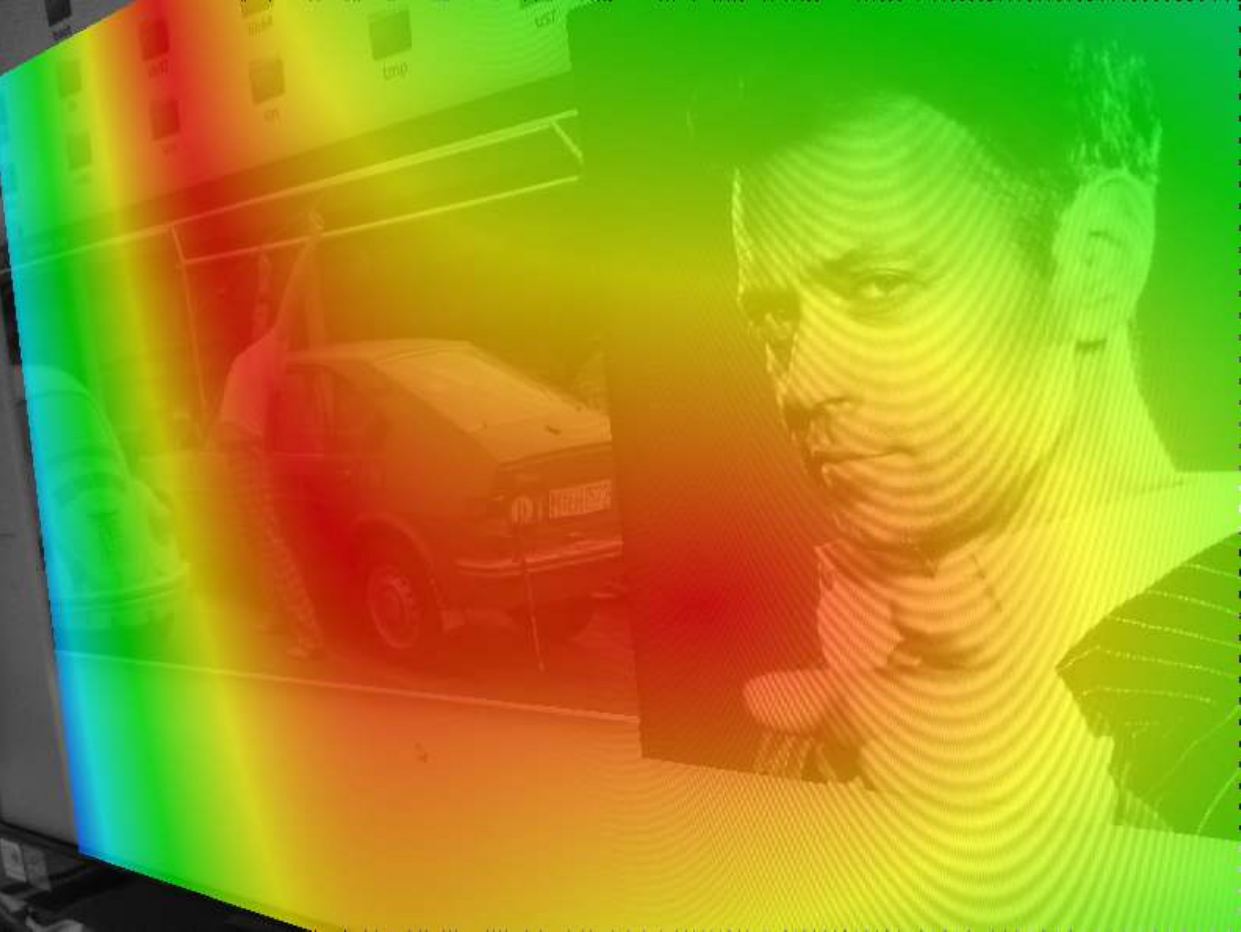} \\
			\rotatebox{90}{\hspace{1em}non-planar} &	
			\includegraphics[width=0.0116\textwidth,height=0.12\textwidth]{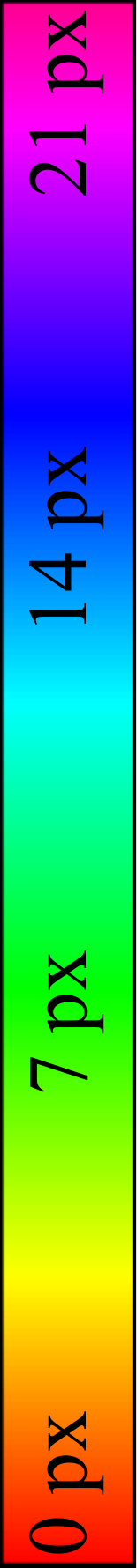} &
			\subfloat[$I_1$]{\label{anp1}
				\includegraphics[width=0.18\textwidth]{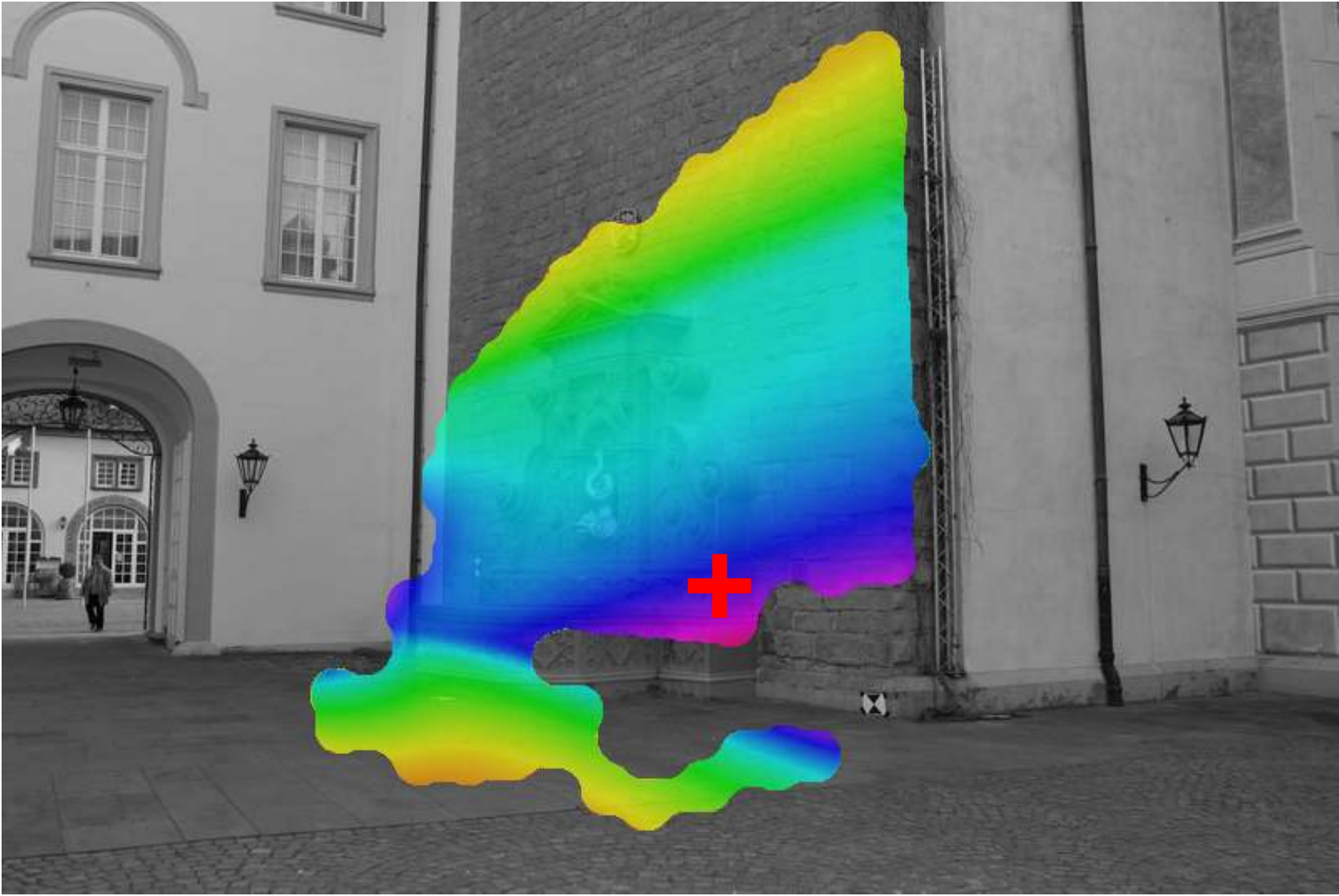}
			} &
			\subfloat[$I_2$]{\label{anp2}
				\includegraphics[width=0.18\textwidth]{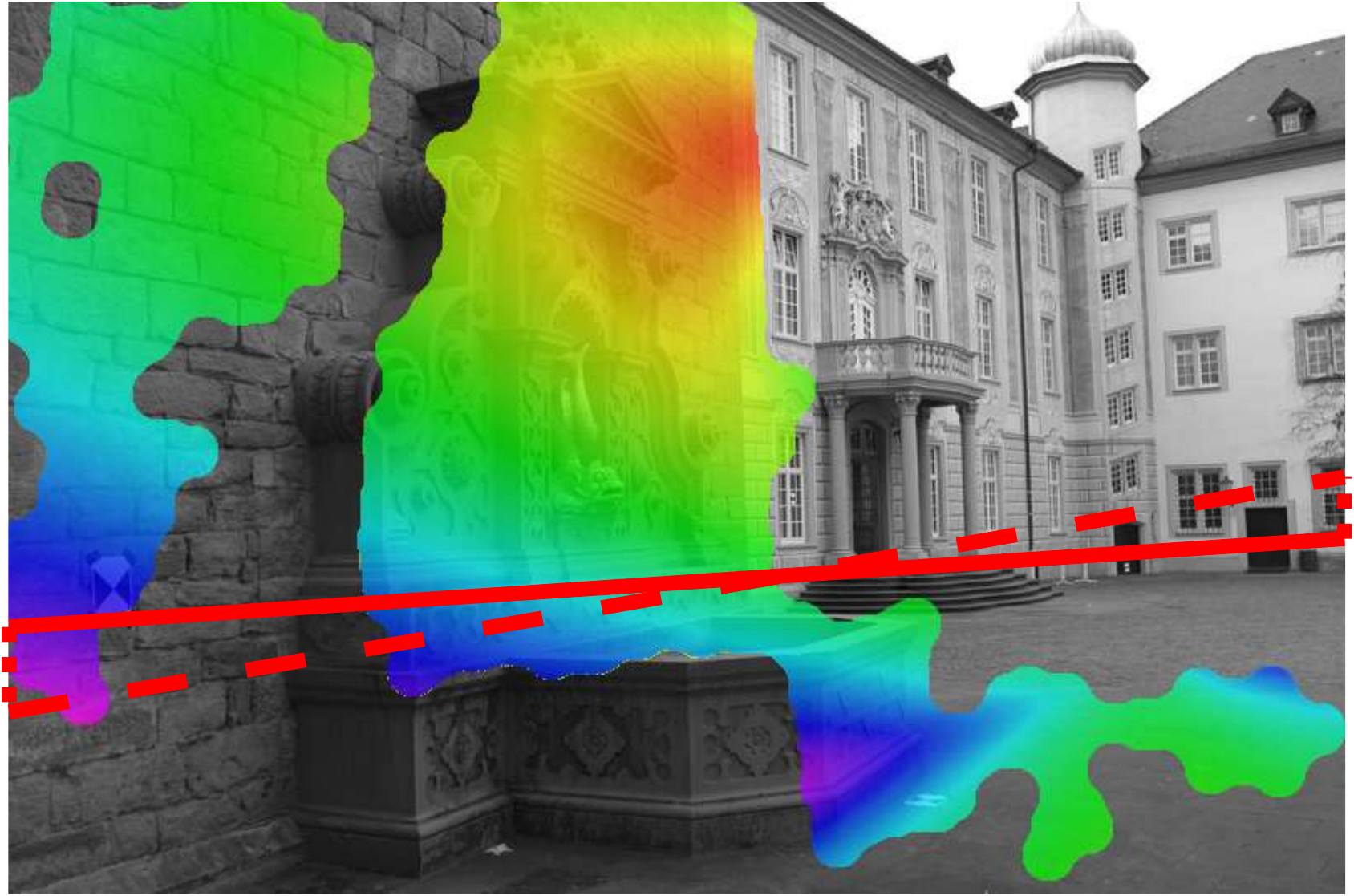}
			}	
		\end{tabular}	
	}
	\caption{\label{accuracy}
		Reprojection error maps for valid image points employed to compute the accuracy, input matches have been obtained by Slime without RANSAC. Non-valid image points are in gray, color bars refer to the measured errors. For the non-planar scene, the epipolar lines in $I_2$ corresponding to the point highlighted in red in $I_1$ obtained by $\tilde{F}$, $F_{GT}$  are indicated by a dashed, solid red lines, respectively. The error associated to the point is the area enclosed by the red polyline, including the dotted border red lines (see Sec.~\ref{accs}). Best viewed in color and zoomed in.}\vspace{-0.25em}
\end{figure}

The accuracy measures how well the model derived from the matches fits with the real scene orientation. For a planar image pair $\eta$, the average $E_{12}^\eta$ of reprojection errors $\parallel\tilde{H}\mathbf{x}-H_{GT}\mathbf{x}\parallel$ among the whole valid points $\mathbf{x}\in I_1$ is computed, where $\tilde{H}$ is the planar homography obtained from the matches of the method under evaluation and $H_{GT}$ is the GT homography. Likewise, $E_{21}^\eta$ is computed over the valid points $\mathbf{x}'\in I_2$ using the inverse homography. Following~\cite{imw2020}, the final accuracy is
\begin{equation}\label{acc_formula}
	\frac{1}{t_\perp N}\sum_{t=1}^{t_\perp}\sum_{\eta=1}^{N}\text{\textlbrackdbl}\max(E_{12}^\eta,E_{21}^\eta)<t\text{\textrbrackdbl}
\end{equation}
where $N$ is the total number of image pairs. In the case of non-planar scenes, $E^\eta_{12}$ compares the pointwise error of the fundamental matrix $\tilde{F}$ computed from the matches against the GT fundamental matrix $F_{GT}$, and likewise $E^\eta_{21}$. Specifically, the average is computed defining the error on a valid point $\mathbf{x}\in I_1$ as the minimum area given by intersecting the two semi-planes on $I_2$ defined by the epipolar lines $\tilde{F}\mathbf{x}$ and $F_{GT}\mathbf{x}$ as done in\cite{noransac}, but normalizing instead the area by the image diagonal to get a linear measure. The final accuracy over all the image pairs is aggregated as for Eq.~\ref{acc_formula}. Visual examples of the reprojection error maps are shown in Fig.~\ref{accuracy}.

A more detailed discussion about the defined metrics can be found in SM\ref{sm_metric}.

\begin{figure}[t]
	\center
	\subfloat[\small planar dataset]{\label{pd}
		\includegraphics[width=0.42\textwidth]{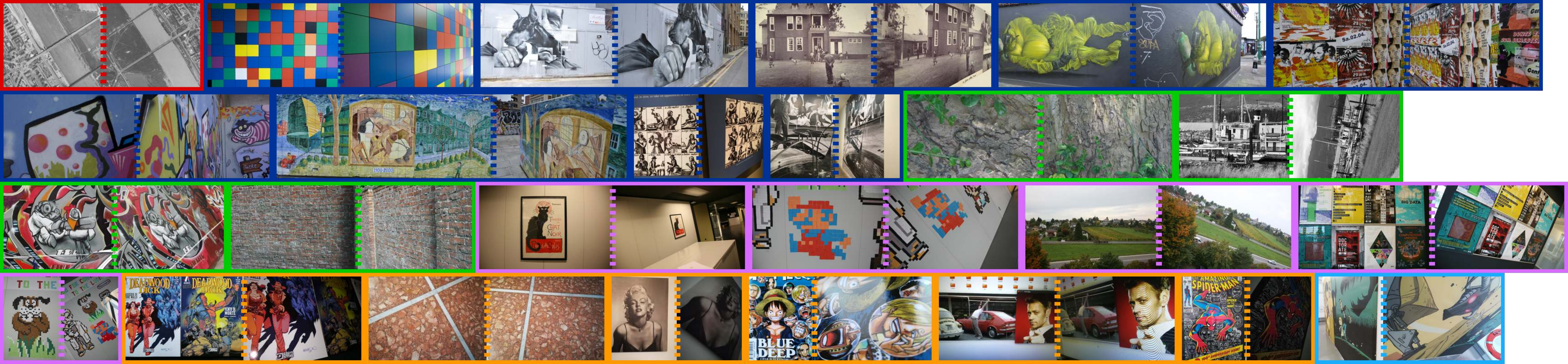}
	}
	\\
	\subfloat[\small non-planar dataset]{\label{npd}
		\includegraphics[width=0.42\textwidth]{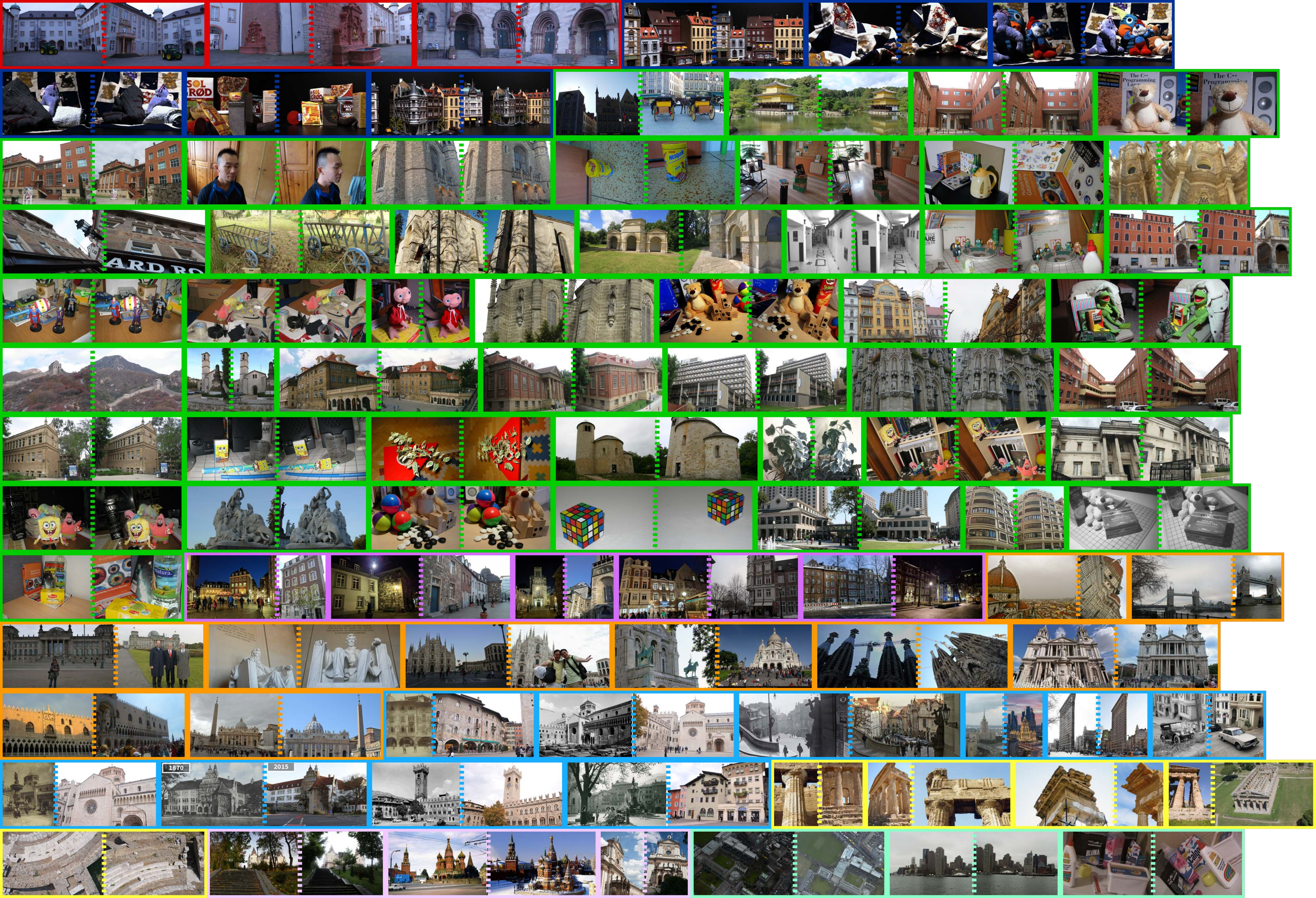}
	}
	\caption{\label{data_gallery}
		\protect\subref{pd} Examples of image pair for each scene in the planar dataset. Frame colors refer in order to the original dataset of the scene:~\cite{historical_aerial}, HPatches, Oxford,~\cite{learning_ori},~\cite{sgloh2} and this new one. \protect\subref{npd} Examples of image pair for each scene in the non-planar dataset. Frame colors refer in order to the original dataset of the scene: Stretcha, DTU, mixed datasets collected in~\cite{dtm}, Aachen Day-Night, IMC-PT,~\cite{low3d},~\cite{cultural_heritage},~\cite{wxbs} and various additional datasets (see Sec.~\ref{dataset_desc}). Best viewed in color and zoomed in.}\vspace{-1em}
\end{figure}

\vspace{-0.75em}
\subsection{Ground-truth estimation}
To compute the error metrics, GT must be available. For the planar dataset GT is obtained by hand-taken correspondences distributed so as to cover uniformly the whole overlapped area between the images. From these points the planar homography $H_{GT}$ is computed by DLT~\cite{multiview}. 
A match is considered correct if it lies on the plane and $\epsilon^{H_{GT}}<t_\perp$ as in Eq.~\ref{max_reproj}.

The GT fundamental matrices $F_{GT}$ for the non-planar dataset have been computed by hand-taken correspondences like in the case of the planar dataset. Following the same approach described in~\cite{dtm} matches are considered correct if $\varepsilon^{F_{GT}}<t_\perp$ as in Eq.~\ref{max_epi_error}, but it is also required that the match is compatible with the local neighborhood of hand-taken matches used to compute $F_{GT}$. Due to the presence of semi-dense methods in the evaluation, with respect to the original dataset more hand-taken matches have been added. The acquisition of the hand-taken GT matches is further discussed in SM\ref{sm_gt}.

\vspace{-0.25em}
\subsection{Standard pose-based benchmarks}\label{standard_bench}
Megadepth~\cite{megadepth} and ScanNet~\cite{scannet} are used in this case, respectively for the outdoor and indoor evaluation, following the protocol reported in\mbox{\cite{loftr}} which considers the Area Under the Curve (AUC) according to the pose error computed as the maximum within the rotation and translation angular errors. There are 1500 tested image pairs for each datasets, beloging to only two scenes in the case of MegaDepth and to about one hundred scenes for ScanNet. As for the other above setup and unlike previous works, RANSAC post-processing is added before the pose estimation, see SM\ref{sm_standard}.

\vspace{-1em}
\subsection{Results}\label{results}
Table~\ref{result_plot} reports the coverage, precision and accuracy metrics of the compared methods on the planar and non-planar datasets, referring to the best RANSAC setup with respect to the accuracy maximization for the non-planar scenes. The number of times a method fails is also reported, distinguishing between the case when no matches are provided and the case when only wrong matches are produced, being the former situation more preferable. Bars are normalized within the minimum and maximum column values and darker colors mean better values. Some exemplar visual results for challenging matching pairs are shown in Fig.~\ref{planar_nonplanar_match}. Detailed tabular results according to different RANSAC inlier thresholds can be found in SM\ref{sm_res1}.

\subsubsection{Planar dataset}\label{psec}
According to Table~\ref{result_plot} the coverage of Slimed Hz$^+$ is almost the same of that provided by LoFTR-based architectures, i.e. LoFTR, SE2-LoFTR, QuadTree Attention and Matchformer. Together with ECO-TR, all the above methods get a coverage of about 35\%, surpassed only by DKM which achieves a coverage around 45\%. In order, the coverages of SuperGlue, DISK, Hz$^+$ and Key.Net, the last two slimed or not, are placed in a range between 25-30\%, while RootSIFT has the lowest coverage around 10\% increased to about 20\% when slimed. RANSAC decreased the coverage by about 5\% with respect to raw matches, see SM\ref{sm_res1}.

\begin{figure}[t]
	\center
	\resizebox{0.36\textwidth}{!}{
		\begin{tabular}{ccc}	
			\rotatebox{90}{\hspace{2.2em}planar} &
			\includegraphics[width=0.18\textwidth]{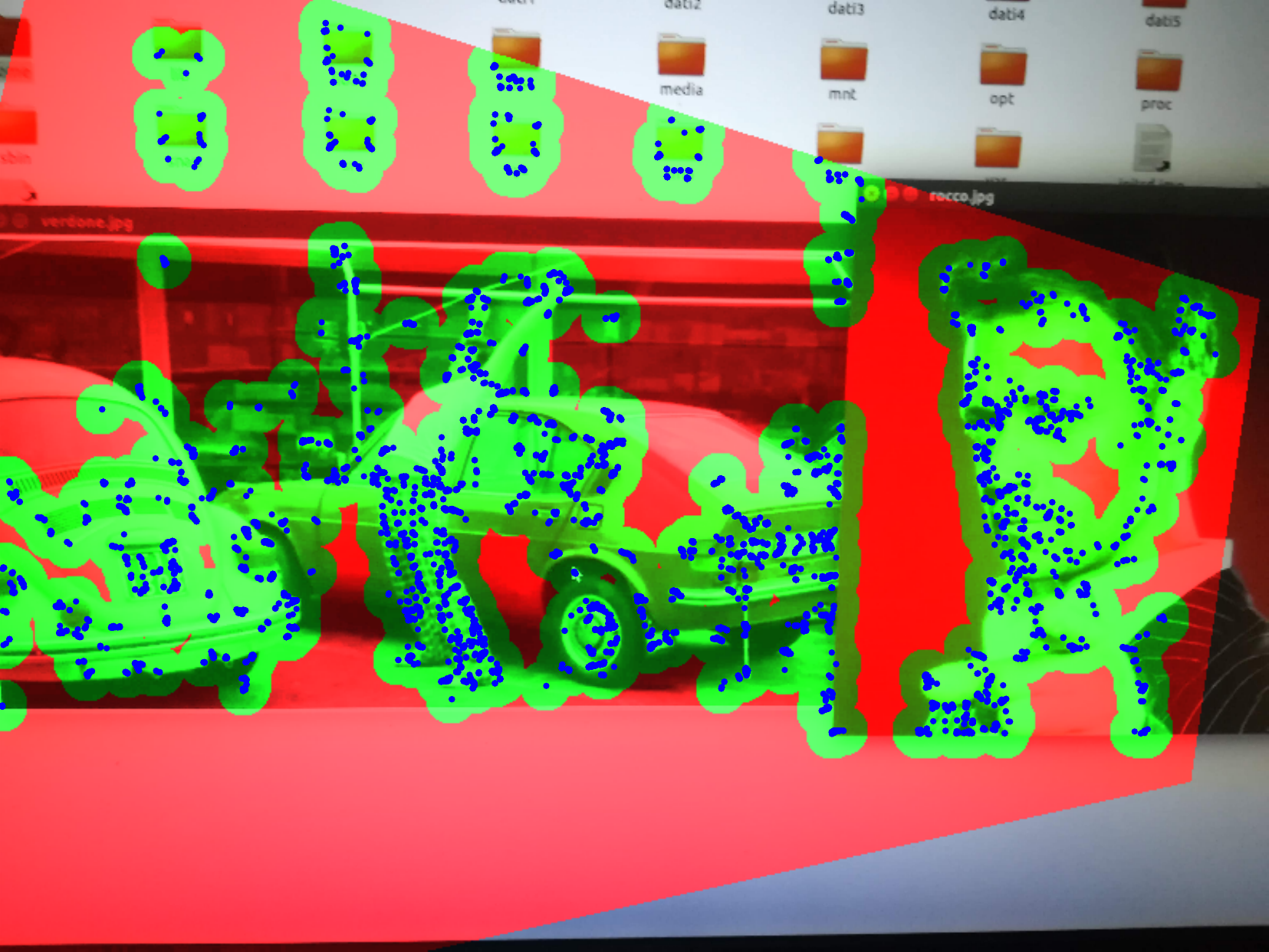} &
			\includegraphics[width=0.18\textwidth]{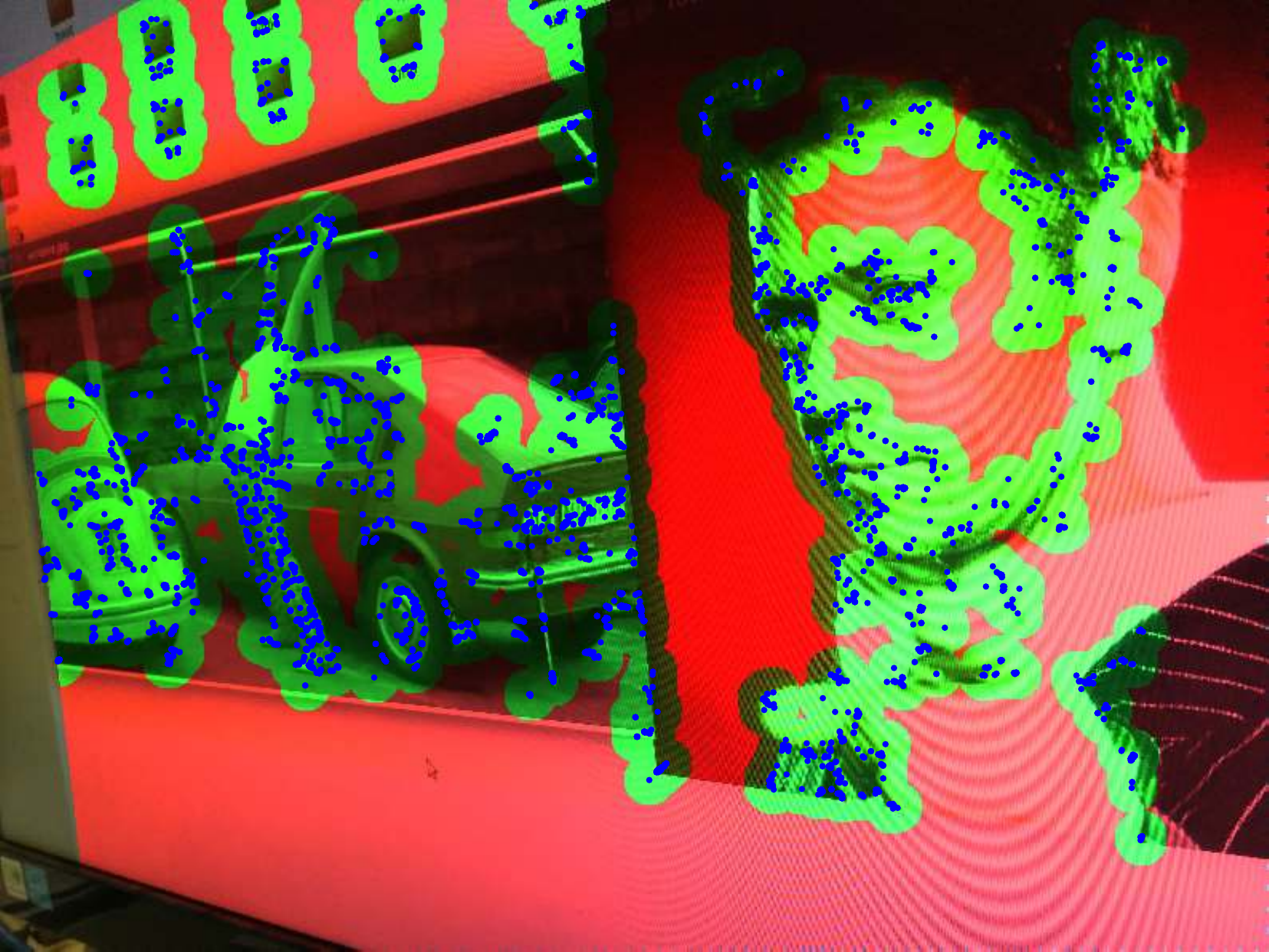} \\
			\rotatebox{90}{\hspace{1em}non-planar} &
			\subfloat[$I_1$]{\label{rnp1}
				\includegraphics[width=0.18\textwidth]{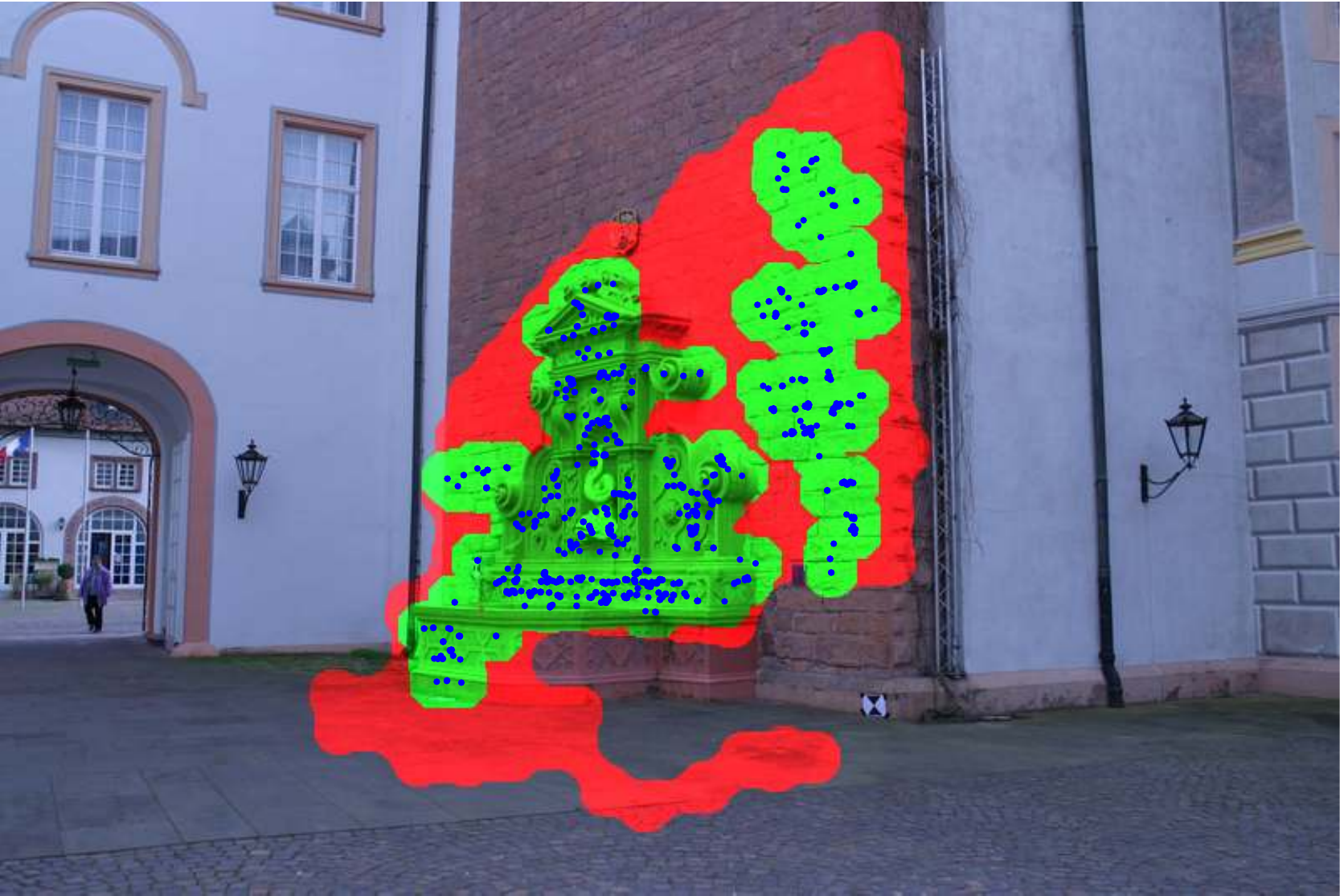}
			} &
			\subfloat[$I_2$]{\label{rnp2}
				\includegraphics[width=0.18\textwidth]{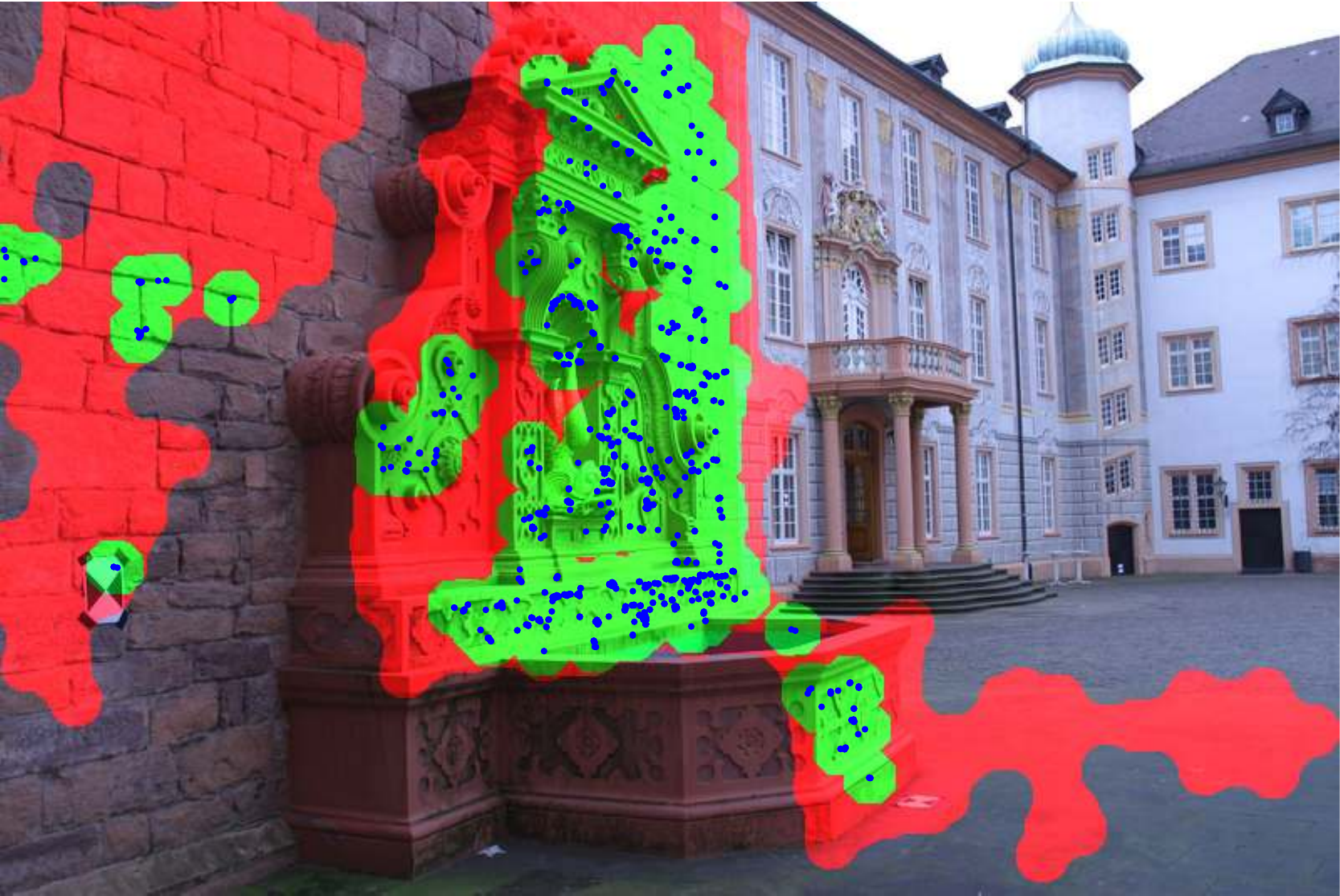}
			}	
		\end{tabular}
	}
	\caption{\label{recall}
		The raw image coverage region on each image, shown in green, is obtained by expanding through morphological dilation the keypoints of the correct matches, indicated in blue. The image region of the valid points used as normalization factor is given by the union of red and green areas. In the example input matches have been obtained by Slime without RANSAC (see Sec.~\ref{covs}). Best viewed in color and zoomed in.}\vspace{-0.25em}
\end{figure}

Concerning the precision, all the methods are over 80\%, with the exception of the base RootSIFT. RootSIFT, slimed or not, and DISK appear to be the methods which rely more on RANSAC to boost their results, see SM\ref{sm_res1}. Key.Net and Hz$^+$, both slimed or not, work slightly better than other methods in terms of match precision. Slimed RootSIFT boosts the base RootSIFT precision by a 15\%, while Slimed Key.Net slightly decreases the precision of the base Key.Net pipeline.

Regarding the accuracy, the rank is similar to that obtained in the case of the precision when RANSAC is used. Otherwise, LoFTR-based methods, and more in particular RootSIFT, DISK and Hz$^+$, get much lower accuracy values. These results are reasonable since both Key.Net, which uses AdaLAM, and the Slime framework implement a local plane-based matching strategy. Likely, SuperGlue, ECO-TR and DKM implicitly employ a similar mechanism. Slimed Hz$^+$, Key.Net and Hz$^+$ achieve the highest accuracy level with respect to the others since they are able to deal with non-upright images. Their upright counterparts and the remaining methods are unable to deal with this specific transformation of the image.

Without considering the ten truly non-upright image pairs, RootSIFT, ECO-TR and DISK are the methods which fail often, as one can observe by inspecting the two rightmost columns for the planar benchmark. For ECO-TR these failures are due to out-of-memory GPU issues, hence the count refers mainly to pairs with no matches. Interestingly, only Hz$^+$, slimed or not, is able to find correct matches for all the image in the planar dataset, and in general Slimed Hz$^+$ provides a higher absolute number of correct matches than the base Hz$^+$ (see Fig.~\ref{planar_nonplanar_match}). Notice also that Slimed RootSIFT decreases and moves the count of wrong only match pairs to pairs with no matches, implying that Slime solution is more robust. Differently, Slimed Key.Net seems to generally decrease the original performances of its base matching pipeline. 

\subsubsection{Non-planar dataset}\label{npsec}
As shown in Table~\ref{result_plot}, the coverage relative rank for the non-planar dataset is in accordance to the planar case, but is slightly lower with respect to the planar case in terms of absolute values, with the exception of DISK, LoFTR-based architectures and DKM. Slimed Hz$^+$ coverage is better than that of other hybrid and fully handcrafted evaluated methods, DISK and SuperGlue, and close to that of LoFTR-base methods. The coverage difference between raw and RANSAC-filtered matches, unlike the planar case, is almost absent, see SM\ref{sm_res1}.

\begin{table*}[t!]
	\centering
	\caption{\label{result_plot}
		Comparative results for the planar and non-planar datasets (see Sec.~\ref{results}). Best viewed in color.}	
	\renewcommand{\arraystretch}{0}
	\setlength{\tabcolsep}{4pt}
	\centering
	\resizebox{0.89\textwidth}{!}{
		\begin{tabular}{r<{}c<{}l<{}l<{}l<{}l<{}l<{}l<{}l<{}l<{}l<{}l}
			\toprule
			& & \multicolumn{5}{c}{Planar} & \multicolumn{5}{c}{Non-planar}\\
			\cmidrule(lr){3-7}\cmidrule(lr){8-12}
			& \shortstack[c]{RANSAC\\thr. (px)} & \shortstack[c]{Coverage\\(\%)} & \shortstack[c]{Precision\textcolor{white}{y}\\(\%)} & \shortstack[c]{Accuracy\\(\%)} & \shortstack[l]{No match\textcolor{white}{y}\\pairs} & \shortstack[l]{Wrong only\\pairs} & \shortstack[c]{Coverage\\(\%)} & \shortstack[c]{Precision\textcolor{white}{y}\\(\%)} & \shortstack[c]{Accuracy\\(\%)} & \shortstack[l]{No match\textcolor{white}{y}\\pairs} & \shortstack[l]{Wrong only\\pairs} \\[5pt]
			\midrule
			RootSIFT\hphantom{$^\Rsh$} & 2 & \Chart{8.19}{0.045}{blue}{25}{12}\hphantom{999999} & \Chart{64.30}{0.468}{red}{34}{12}\hphantom{99999} & \Chart{50.40}{0.569}{teal}{34}{12}\hphantom{99999} & \Chart{0}{0.000}{orange}{62}{12}\hphantom{99999999.} & \Chart{41}{0.868}{purple}{25}{12}\hphantom{9999999.} & \Chart{4.99}{0.045}{blue}{25}{12}\hphantom{999999} & \Chart{53.02}{0.362}{red}{25}{12}\hphantom{99999} & \Chart{17.12}{0.239}{teal}{25}{12}\hphantom{99999} & \Chart{0}{0.000}{orange}{62}{12}\hphantom{99999999.} & \Chart{42}{0.952}{purple}{25}{12}\hphantom{9999999.} \\
			\rowcolor{gray!15}RootSIFT$^\Rsh$ & 2 & \Chart{8.54}{0.053}{blue}{25}{12} & \Chart{60.87}{0.420}{red}{25}{12} & \Chart{47.80}{0.540}{teal}{34}{12} & \Chart{2}{0.045}{orange}{62}{12} & \Chart{45}{0.952}{purple}{25}{12} & \Chart{5.56}{0.058}{blue}{25}{12} & \Chart{55.21}{0.394}{red}{25}{12} & \Chart{24.11}{0.336}{teal}{25}{12} & \Chart{0}{0.000}{orange}{62}{12} & \Chart{41}{0.930}{purple}{25}{12} \\
			Slimed RootSIFT\hphantom{$^\Rsh$} & 2 & \Chart{18.60}{0.280}{blue}{25}{12} & \Chart{89.94}{0.819}{red}{53}{12} & \Chart{69.36}{0.783}{teal}{53}{12} & \Chart{14}{0.317}{orange}{62}{12} & \Chart{0}{0.000}{purple}{62}{12} & \Chart{12.74}{0.221}{blue}{25}{12} & \Chart{70.60}{0.621}{red}{44}{12} & \Chart{33.47}{0.467}{teal}{34}{12} & \Chart{27}{0.918}{orange}{25}{12} & \Chart{7}{0.159}{purple}{62}{12} \\
			\rowcolor{gray!15}Slimed RootSIFT$^\Rsh$ & 2 & \Chart{14.06}{0.178}{blue}{25}{12} & \Chart{68.84}{0.530}{red}{34}{12} & \Chart{52.96}{0.598}{teal}{34}{12} & \Chart{42}{0.952}{orange}{25}{12} & \Chart{1}{0.021}{purple}{62}{12} & \Chart{10.62}{0.173}{blue}{25}{12} & \Chart{65.63}{0.548}{red}{34}{12} & \Chart{27.31}{0.381}{teal}{25}{12} & \Chart{28}{0.952}{orange}{25}{12} & \Chart{12}{0.272}{purple}{62}{12} \\
			Key.Net\hphantom{$^\Rsh$} & 2 & \Chart{28.04}{0.493}{blue}{34}{12} & \Chart{98.58}{0.938}{red}{62}{12} & \Chart{82.84}{0.936}{teal}{62}{12} & \Chart{2}{0.045}{orange}{62}{12} & \Chart{0}{0.000}{purple}{62}{12} & \Chart{20.68}{0.402}{blue}{25}{12} & \Chart{89.10}{0.894}{red}{62}{12} & \Chart{51.87}{0.724}{teal}{44}{12} & \Chart{8}{0.272}{orange}{62}{12} & \Chart{2}{0.045}{purple}{62}{12} \\
			\rowcolor{gray!15}Key.Net$^\Rsh$ & 2 & \Chart{26.34}{0.454}{blue}{34}{12} & \Chart{90.22}{0.823}{red}{53}{12} & \Chart{77.07}{0.871}{teal}{62}{12} & \Chart{6}{0.136}{orange}{62}{12} & \Chart{7}{0.148}{purple}{62}{12} & \Chart{21.29}{0.416}{blue}{25}{12} & \Chart{88.48}{0.885}{red}{62}{12} & \Chart{50.46}{0.704}{teal}{44}{12} & \Chart{8}{0.272}{orange}{62}{12} & \Chart{2}{0.045}{purple}{62}{12} \\
			Slimed Key.Net\hphantom{$^\Rsh$} & 3 & \Chart{31.32}{0.567}{blue}{34}{12} & \Chart{96.35}{0.907}{red}{62}{12} & \Chart{79.67}{0.900}{teal}{62}{12} & \Chart{3}{0.068}{orange}{62}{12} & \Chart{2}{0.042}{purple}{62}{12} & \Chart{23.19}{0.459}{blue}{34}{12} & \Chart{88.63}{0.887}{red}{62}{12} & \Chart{43.11}{0.601}{teal}{34}{12} & \Chart{0}{0.000}{orange}{62}{12} & \Chart{0}{0.000}{purple}{62}{12} \\
			\rowcolor{gray!15}Slimed Key.Net$^\Rsh$ & 3 & \Chart{28.65}{0.506}{blue}{34}{12} & \Chart{91.41}{0.839}{red}{53}{12} & \Chart{74.04}{0.836}{teal}{53}{12} & \Chart{5}{0.113}{orange}{62}{12} & \Chart{6}{0.127}{purple}{62}{12} & \Chart{22.91}{0.452}{blue}{34}{12} & \Chart{84.61}{0.828}{red}{53}{12} & \Chart{43.74}{0.610}{teal}{44}{12} & \Chart{0}{0.000}{orange}{62}{12} & \Chart{8}{0.181}{purple}{62}{12} \\
			Hz$^+$\hphantom{$^\Rsh$} & 2 & \Chart{26.33}{0.454}{blue}{34}{12} & \Chart{98.99}{0.943}{red}{62}{12} & \Chart{83.07}{0.938}{teal}{62}{12} & \Chart{0}{0.000}{orange}{62}{12} & \Chart{0}{0.000}{purple}{62}{12} & \Chart{19.74}{0.380}{blue}{25}{12} & \Chart{87.58}{0.872}{red}{62}{12} & \Chart{53.01}{0.740}{teal}{53}{12} & \Chart{0}{0.000}{orange}{62}{12} & \Chart{6}{0.136}{purple}{62}{12} \\
			\rowcolor{gray!15}Hz$^{+\Rsh}$ & 2 & \Chart{25.00}{0.424}{blue}{25}{12} & \Chart{95.35}{0.893}{red}{62}{12} & \Chart{79.29}{0.896}{teal}{62}{12} & \Chart{0}{0.000}{orange}{62}{12} & \Chart{5}{0.106}{purple}{62}{12} & \Chart{20.24}{0.392}{blue}{25}{12} & \Chart{88.17}{0.880}{red}{62}{12} & \Chart{52.88}{0.738}{teal}{44}{12} & \Chart{0}{0.000}{orange}{62}{12} & \Chart{4}{0.091}{purple}{62}{12} \\
			Slimed Hz$^+$\hphantom{$^\Rsh$} & 3 & \Chart{34.87}{0.647}{blue}{44}{12} & \Chart{99.49}{0.950}{red}{62}{12} & \Chart{82.74}{0.935}{teal}{62}{12} & \Chart{0}{0.000}{orange}{62}{12} & \Chart{0}{0.000}{purple}{62}{12} & \Chart{31.14}{0.639}{blue}{44}{12} & \Chart{91.69}{0.932}{red}{62}{12} & \Chart{53.20}{0.742}{teal}{53}{12} & \Chart{0}{0.000}{orange}{62}{12} & \Chart{1}{0.023}{purple}{62}{12} \\
			\rowcolor{gray!15}Slimed Hz$^{+\Rsh}$ & 3 & \Chart{32.23}{0.587}{blue}{34}{12} & \Chart{93.18}{0.864}{red}{62}{12} & \Chart{77.40}{0.874}{teal}{62}{12} & \Chart{7}{0.159}{orange}{62}{12} & \Chart{2}{0.042}{purple}{62}{12} & \Chart{29.79}{0.609}{blue}{44}{12} & \Chart{89.39}{0.898}{red}{62}{12} & \Chart{53.52}{0.747}{teal}{53}{12} & \Chart{0}{0.000}{orange}{62}{12} & \Chart{4}{0.091}{purple}{62}{12} \\
			DISK\hphantom{$^\Rsh$} & 2 & \Chart{22.57}{0.369}{blue}{25}{12} & \Chart{80.52}{0.690}{red}{44}{12} & \Chart{63.97}{0.723}{teal}{44}{12} & \Chart{0}{0.000}{orange}{62}{12} & \Chart{23}{0.487}{purple}{53}{12} & \Chart{22.47}{0.442}{blue}{34}{12} & \Chart{80.23}{0.763}{red}{53}{12} & \Chart{44.57}{0.622}{teal}{44}{12} & \Chart{0}{0.000}{orange}{62}{12} & \Chart{13}{0.295}{purple}{62}{12} \\
			\rowcolor{gray!15}SuperGlue\hphantom{$^\Rsh$} & 2 & \Chart{20.18}{0.316}{blue}{25}{12} & \Chart{90.69}{0.830}{red}{53}{12} & \Chart{78.91}{0.891}{teal}{62}{12} & \Chart{3}{0.068}{orange}{62}{12} & \Chart{10}{0.212}{purple}{62}{12} & \Chart{23.14}{0.458}{blue}{34}{12} & \Chart{90.10}{0.909}{red}{62}{12} & \Chart{60.41}{0.843}{teal}{53}{12} & \Chart{0}{0.000}{orange}{62}{12} & \Chart{4}{0.091}{purple}{62}{12} \\
			LoFTR\hphantom{$^\Rsh$} & 2 & \Chart{26.70}{0.462}{blue}{34}{12} & \Chart{85.06}{0.752}{red}{53}{12} & \Chart{65.39}{0.739}{teal}{53}{12} & \Chart{0}{0.000}{orange}{62}{12} & \Chart{17}{0.360}{purple}{62}{12} & \Chart{32.97}{0.681}{blue}{44}{12} & \Chart{88.22}{0.881}{red}{62}{12} & \Chart{55.84}{0.779}{teal}{53}{12} & \Chart{0}{0.000}{orange}{62}{12} & \Chart{6}{0.136}{purple}{62}{12} \\
			\rowcolor{gray!15}SE2-LoFTR\hphantom{$^\Rsh$} & 2 & \Chart{32.69}{0.597}{blue}{34}{12} & \Chart{93.09}{0.862}{red}{62}{12} & \Chart{76.31}{0.862}{teal}{62}{12} & \Chart{0}{0.000}{orange}{62}{12} & \Chart{7}{0.148}{purple}{62}{12} & \Chart{31.37}{0.645}{blue}{44}{12} & \Chart{88.05}{0.879}{red}{62}{12} & \Chart{58.58}{0.817}{teal}{53}{12} & \Chart{0}{0.000}{orange}{62}{12} & \Chart{2}{0.045}{purple}{62}{12} \\
			MatchFormer\hphantom{$^\Rsh$} & 2 & \Chart{27.56}{0.482}{blue}{34}{12} & \Chart{85.83}{0.763}{red}{53}{12} & \Chart{64.82}{0.732}{teal}{44}{12} & \Chart{0}{0.000}{orange}{62}{12} & \Chart{17}{0.360}{purple}{62}{12} & \Chart{34.68}{0.720}{blue}{44}{12} & \Chart{91.86}{0.935}{red}{62}{12} & \Chart{57.72}{0.805}{teal}{53}{12} & \Chart{0}{0.000}{orange}{62}{12} & \Chart{2}{0.045}{purple}{62}{12} \\
			\rowcolor{gray!15}QuadTree Att.\hphantom{$^\Rsh$} & 2 & \Chart{34.18}{0.631}{blue}{44}{12} & \Chart{87.69}{0.788}{red}{53}{12} & \Chart{74.99}{0.847}{teal}{53}{12} & \Chart{0}{0.000}{orange}{62}{12} & \Chart{13}{0.275}{purple}{62}{12} & \Chart{36.05}{0.751}{blue}{53}{12} & \Chart{91.87}{0.935}{red}{62}{12} & \Chart{68.26}{0.952}{teal}{62}{12} & \Chart{0}{0.000}{orange}{62}{12} & \Chart{4}{0.091}{purple}{62}{12} \\
			ECO-TR\hphantom{$^\Rsh$} & 2 & \Chart{34.48}{0.638}{blue}{44}{12} & \Chart{82.87}{0.722}{red}{44}{12} & \Chart{68.79}{0.777}{teal}{53}{12} & \Chart{17}{0.385}{orange}{62}{12} & \Chart{5}{0.106}{purple}{62}{12} & \Chart{31.76}{0.654}{blue}{44}{12} & \Chart{85.76}{0.845}{red}{53}{12} & \Chart{54.79}{0.764}{teal}{53}{12} & \Chart{15}{0.510}{orange}{53}{12} & \Chart{1}{0.023}{purple}{62}{12} \\
			\rowcolor{gray!15}DKM\hphantom{$^\Rsh$} & 2 & \Chart{42.13}{0.810}{blue}{53}{12} & \Chart{90.78}{0.831}{red}{53}{12} & \Chart{81.09}{0.916}{teal}{62}{12} & \Chart{0}{0.000}{orange}{62}{12} & \Chart{11}{0.233}{purple}{62}{12} & \Chart{43.81}{0.927}{blue}{62}{12} & \Chart{88.27}{0.882}{red}{62}{12} & \Chart{66.85}{0.933}{teal}{62}{12} & \Chart{0}{0.000}{orange}{62}{12} & \Chart{3}{0.068}{purple}{62}{12} \\
			\bottomrule
		\end{tabular}
	}\vspace{-1em}
\end{table*}

Concerning the precision, values are almost equal to the planar case, but Key.Net and Hz$^+$ are slightly pushed backward while MatchFormer and QuadTree Attention are pushed forward. Notice that the number of non-upright image pairs is almost negligible, only four pairs, and hence does not affect the global behavior in the dataset.

In general the accuracy with respect to the planar case is decreased, especially for raw matches without RANSAC. This is expected due to the increased complexity of the scenes. QuadTree Attention and DKM achieve the top accuracy, followed by SuperGlue and the remaining LoFTR-based architectures. Hz$^+$, slimed or not, and Key.Net achieve similar values and are positioned next. Key.Net is better than the slimed version, while for RootSIFT the opposite holds. In the non-planar case the difference between precision and accuracy is more marked than in the planar case. Overall, the accuracy amplifies the relative gap in terms of precision between the methods. In some sense, the given precision measure focuses on coarse correspondences, while the accuracy on a fine localization of matches. Notice that SuperGlue, DKM, ECO-TR, Key.Net and Slime obtain in order the best accuracy values without RANSAC, see SM\ref{sm_res1}.

\begin{figure*}[t]
	\center
	\resizebox{0.99\textwidth}{!}{
		\begin{tabular}{ccc|ccc|ccc}
			\multicolumn{3}{c|}{ZeroCalcare} & \multicolumn{3}{c|}{There} & \multicolumn{3}{c}{SunSeason}\\
			\includegraphics[height=0.175\textwidth]{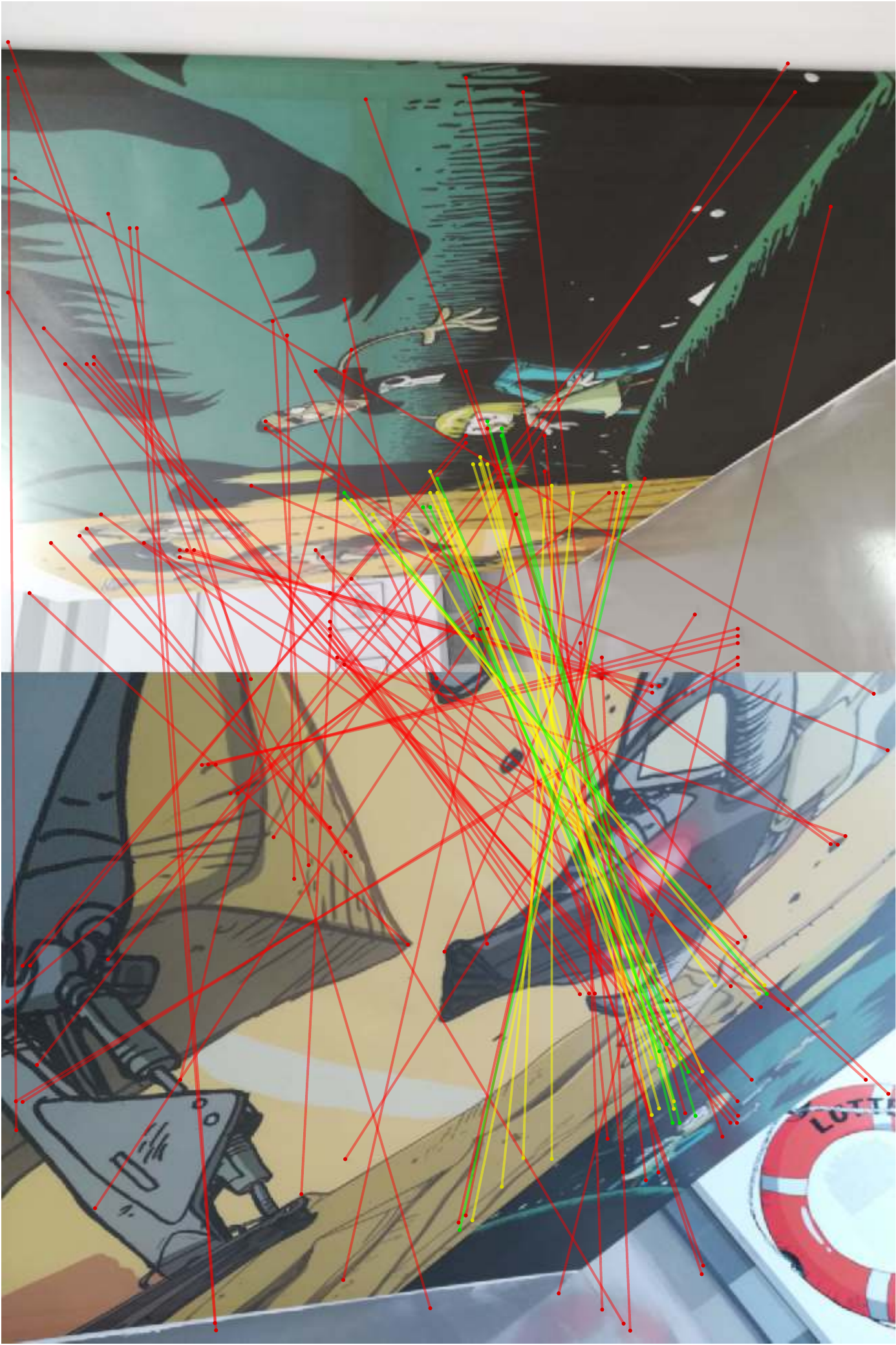} &
			\includegraphics[height=0.175\textwidth]{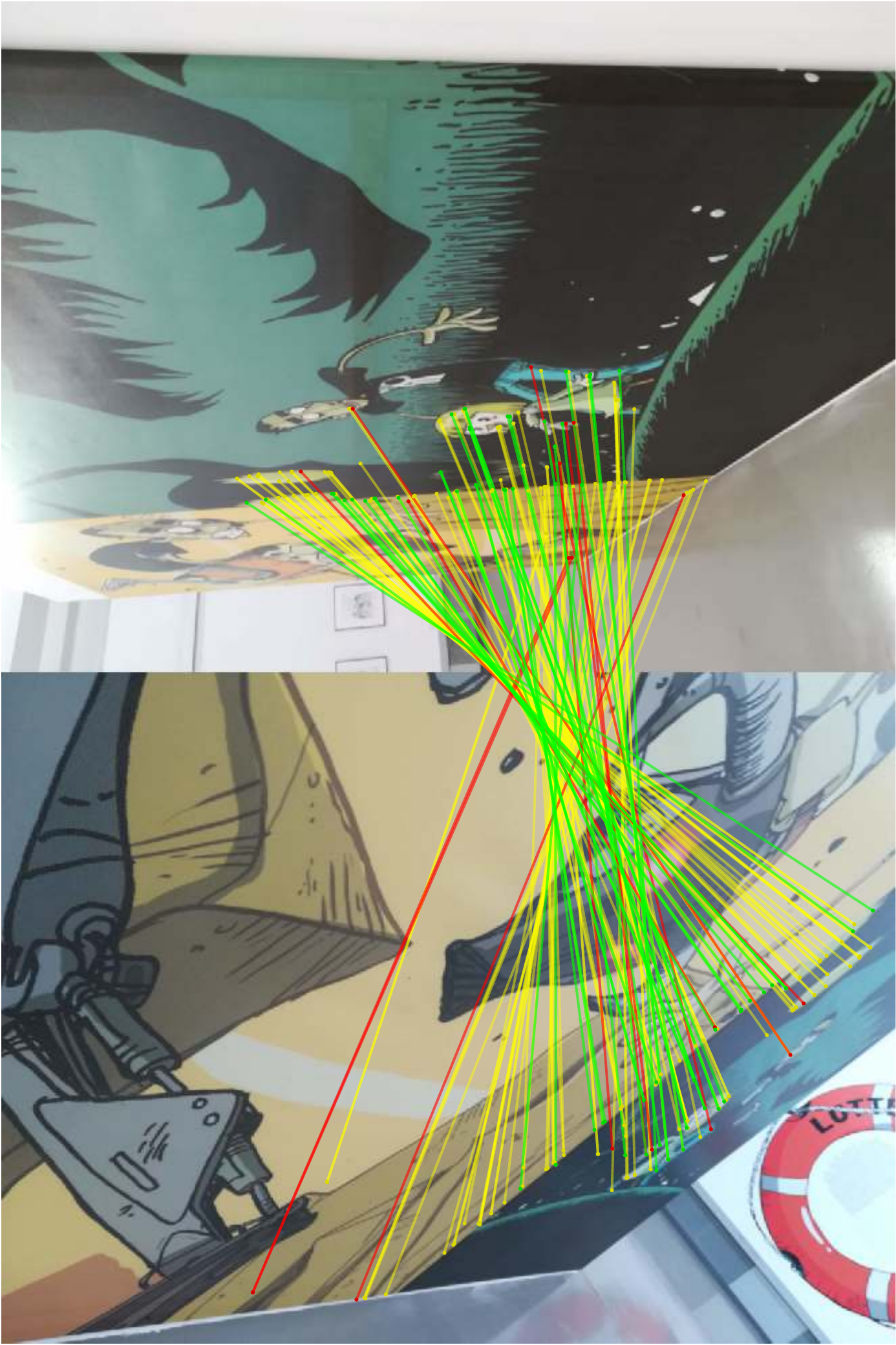} &
			\includegraphics[height=0.175\textwidth]{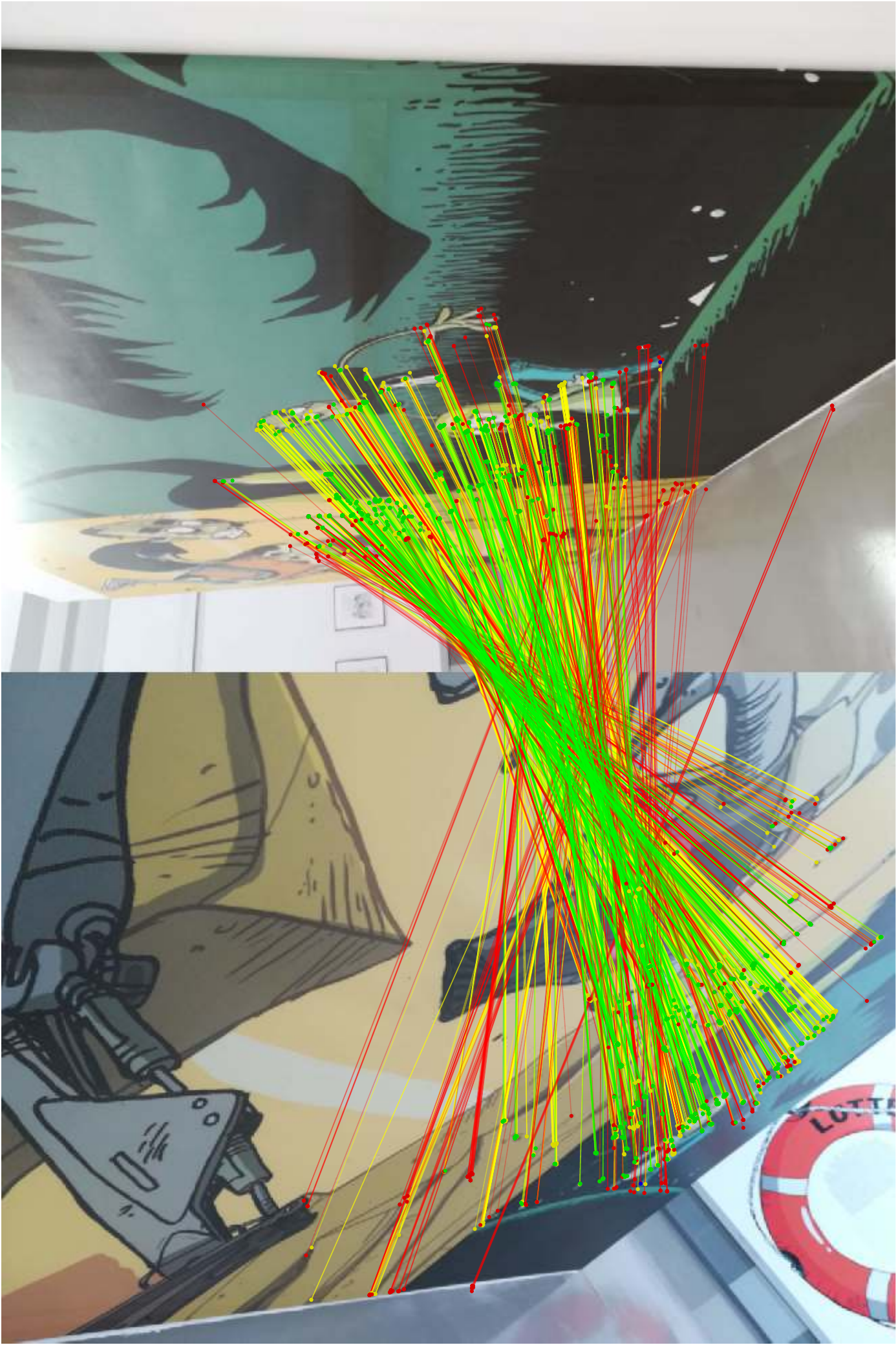} &
			\includegraphics[height=0.175\textwidth]{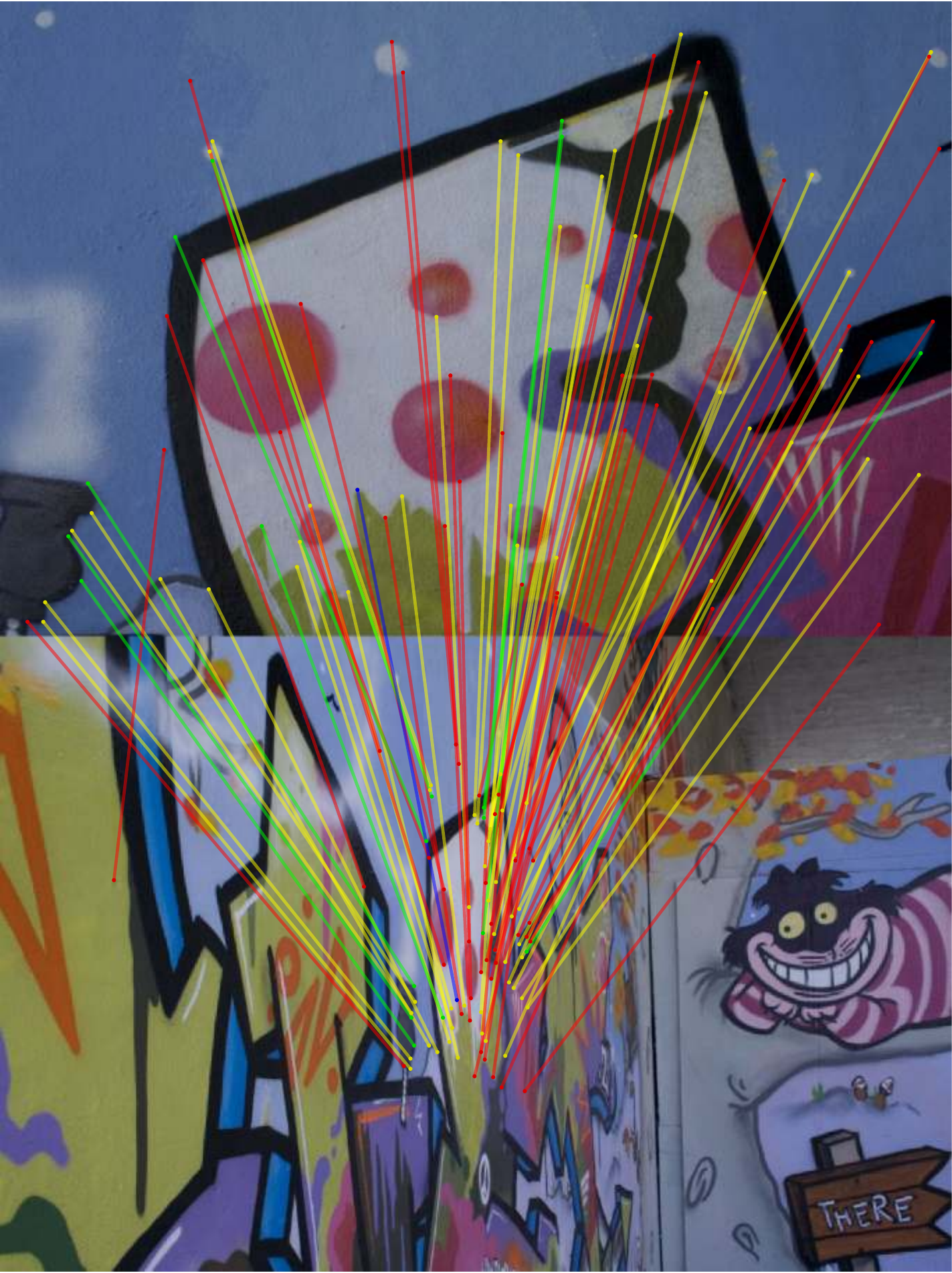} &
			\includegraphics[height=0.175\textwidth]{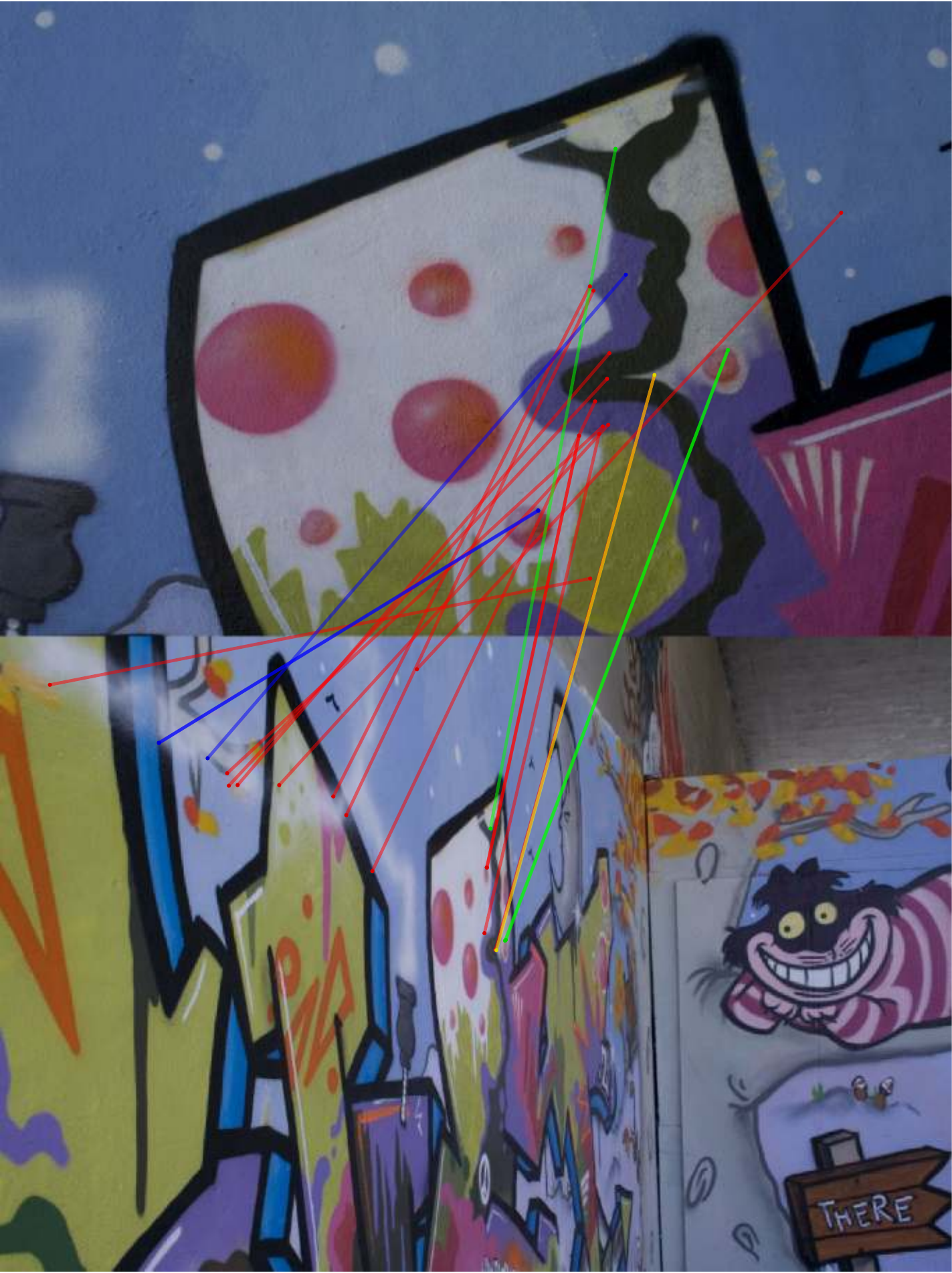} &
			\includegraphics[height=0.175\textwidth]{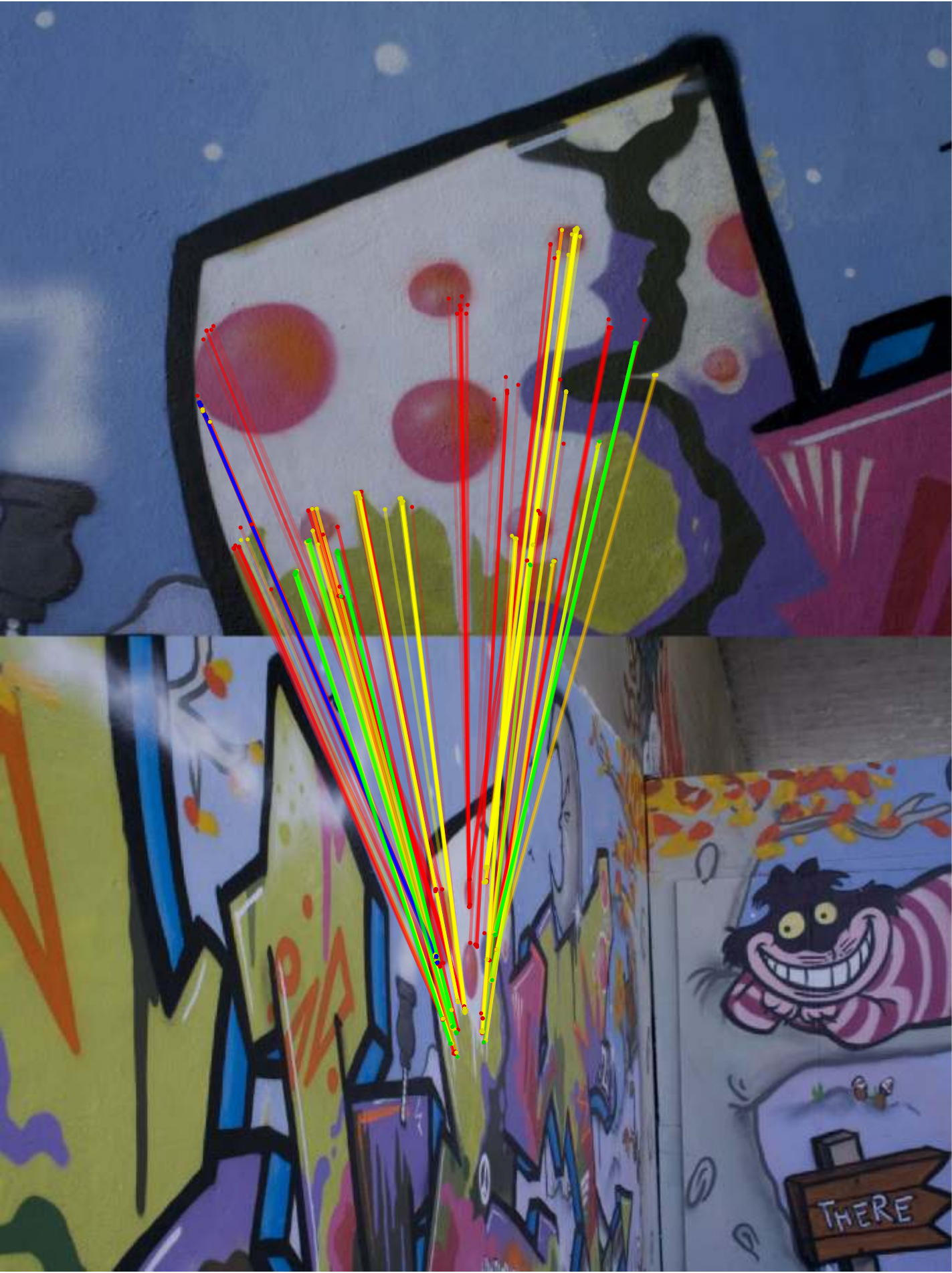} &
			\includegraphics[height=0.175\textwidth]{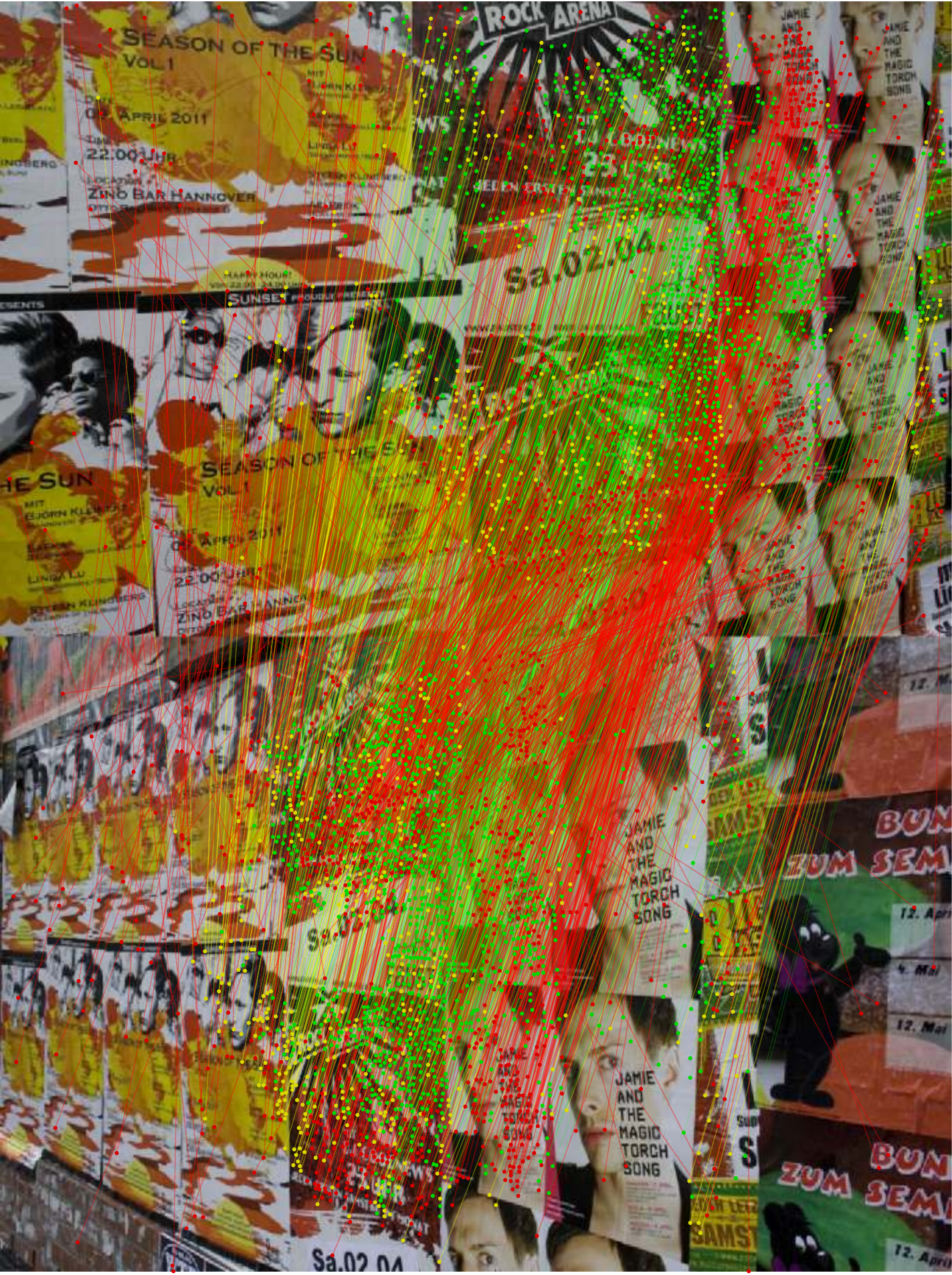} &
			\includegraphics[height=0.175\textwidth]{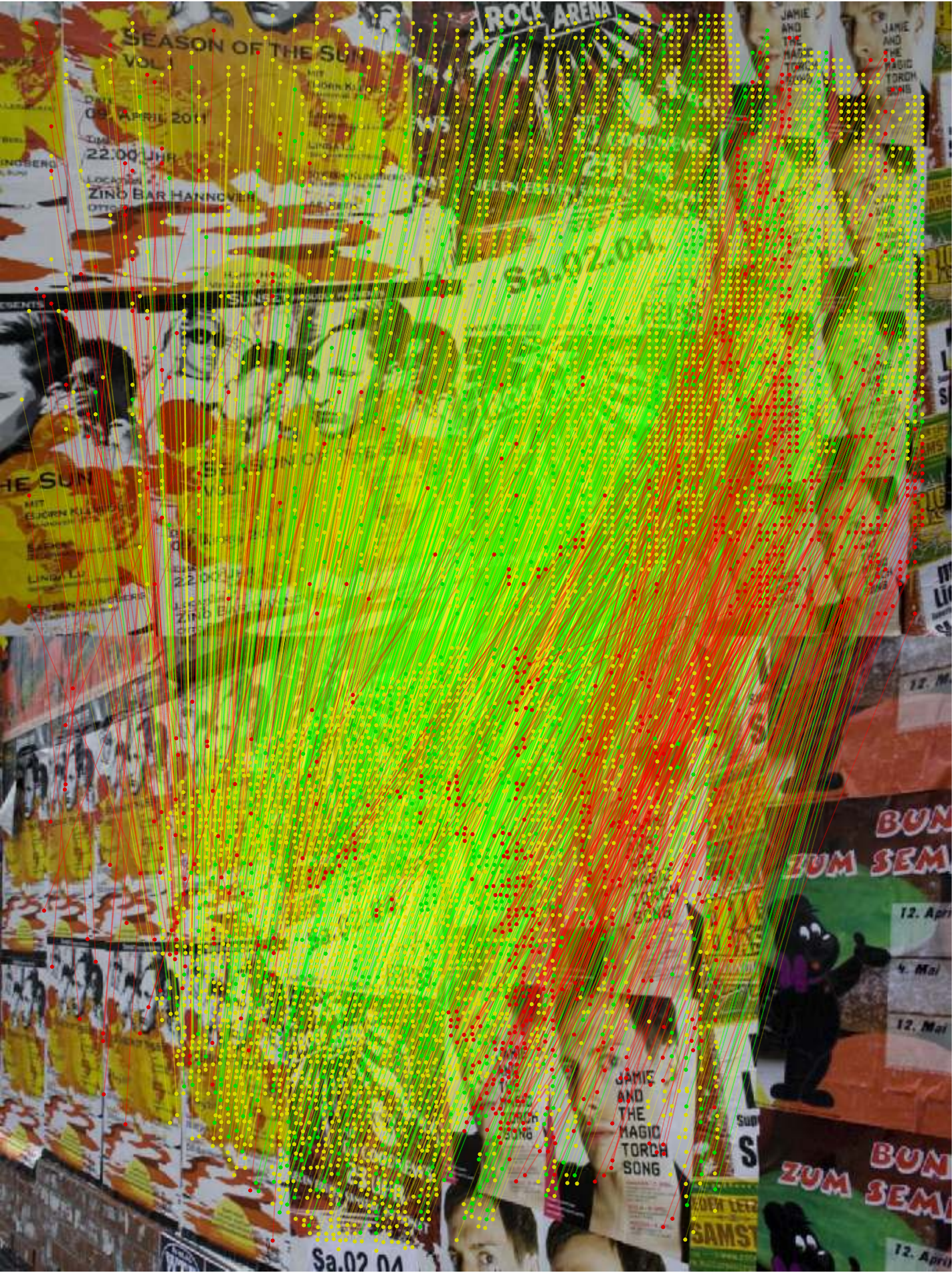} &
			\includegraphics[height=0.175\textwidth]{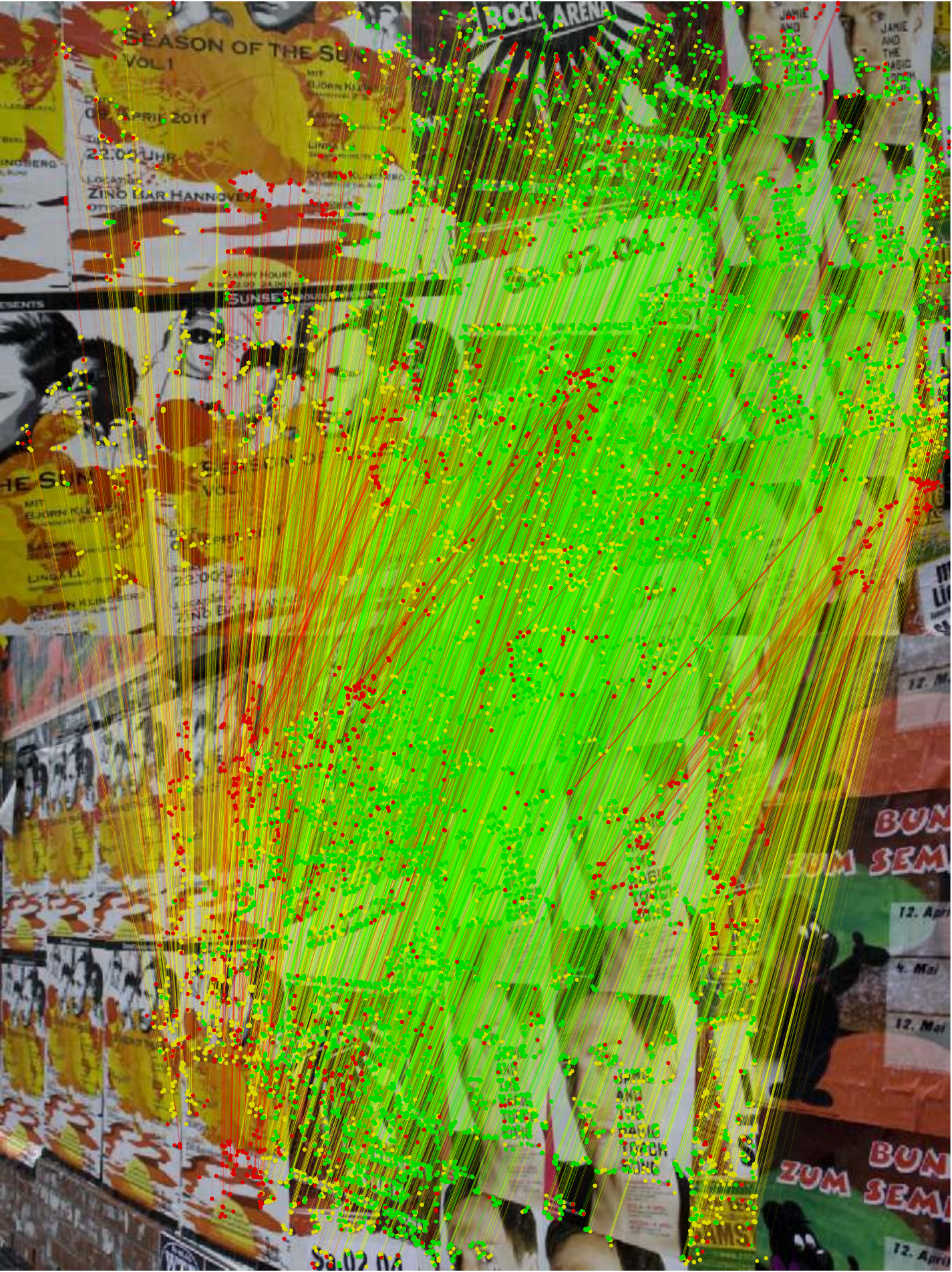} \\
			MatchFormer & Key.Net & Slime & SuperGlue & Hz$^+$ & Slime & DISK & LoFTR & Slime \\
			\midrule
			\multicolumn{3}{c|}{Temple} & \multicolumn{3}{c|}{DayNight} & \multicolumn{3}{c}{NettunoDuomo}\\
			\includegraphics[angle=90,width=0.13\textwidth]{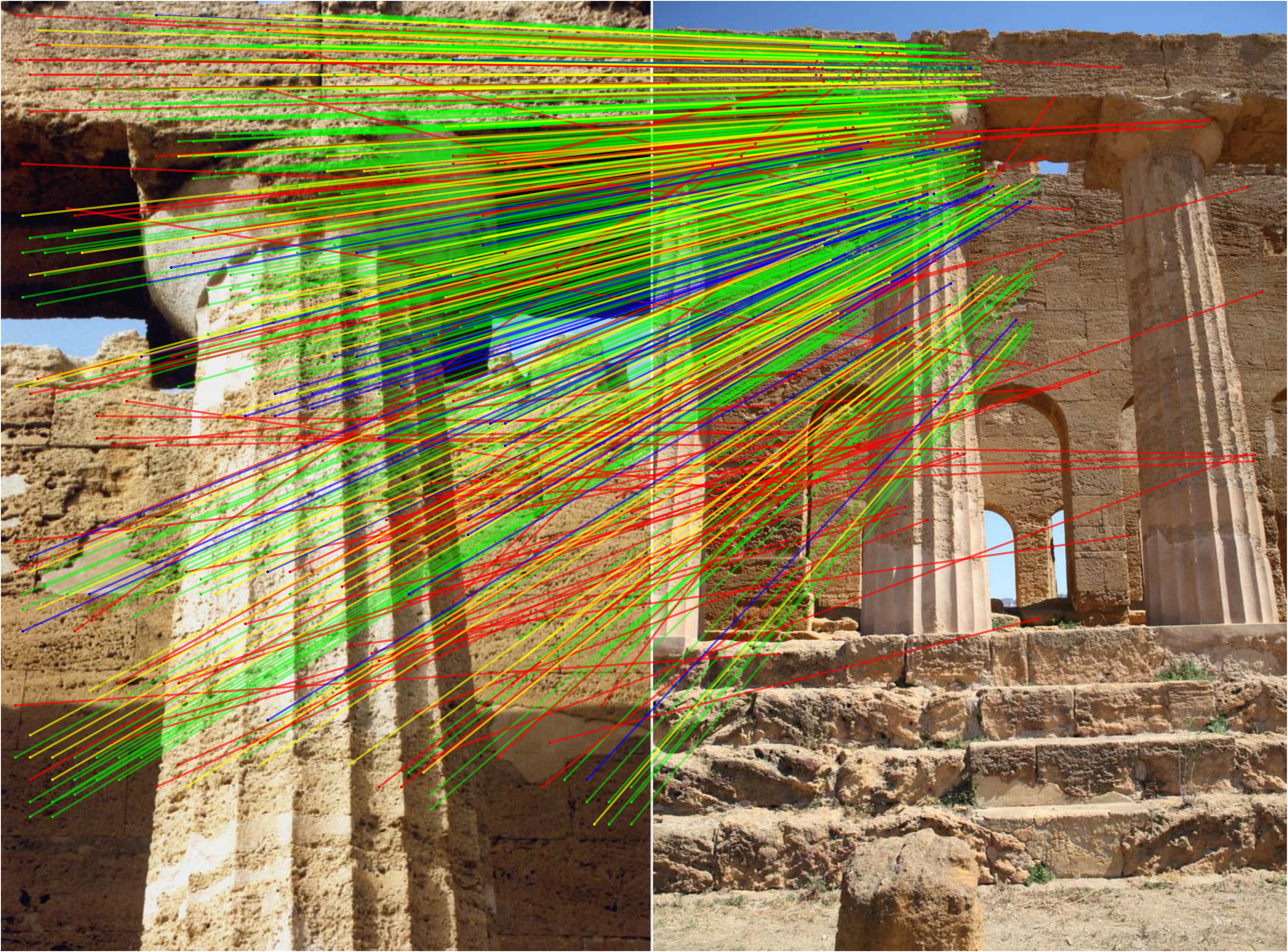} &
			\includegraphics[angle=90,width=0.13\textwidth]{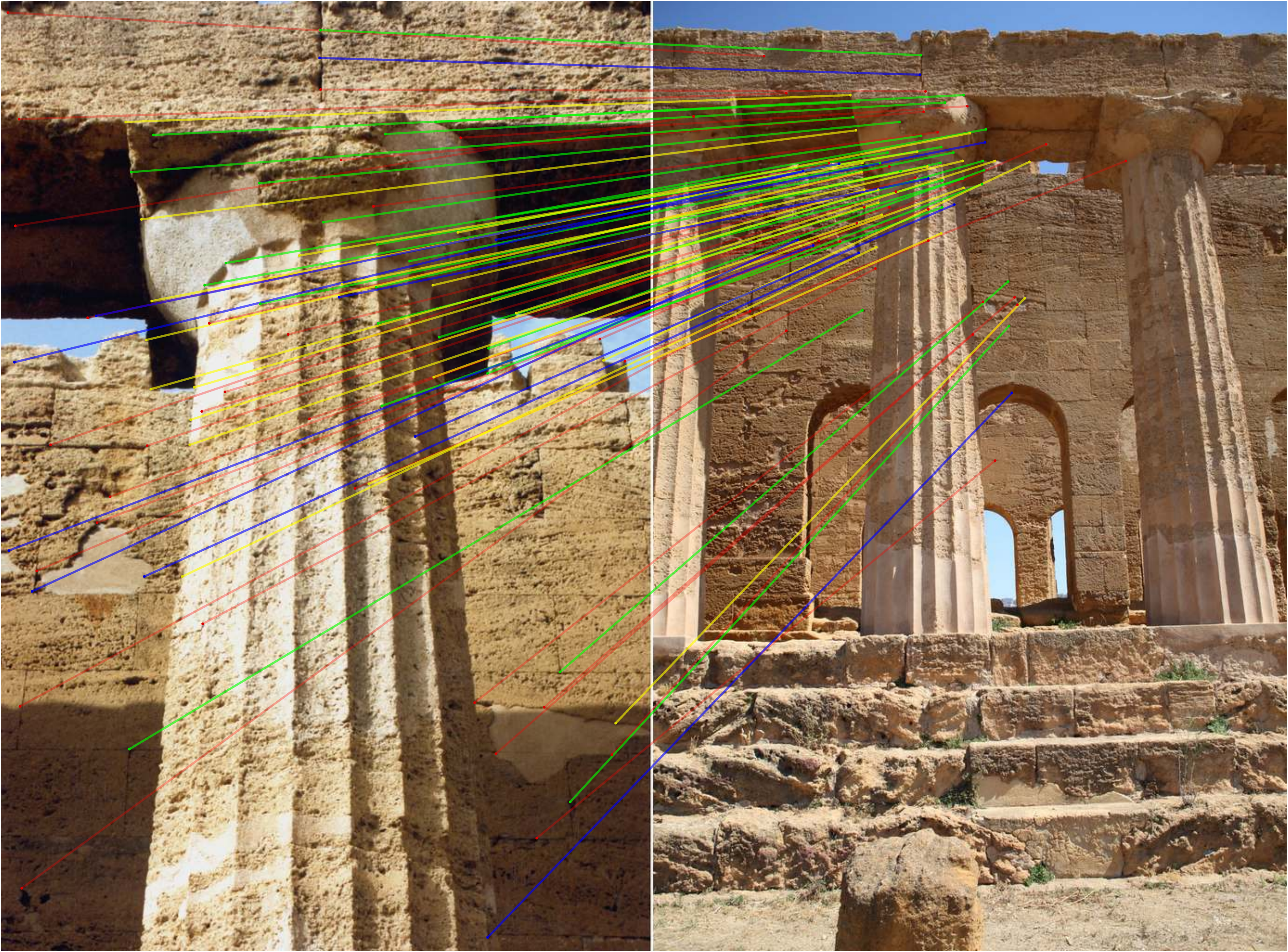} &
			\includegraphics[angle=90,width=0.13\textwidth]{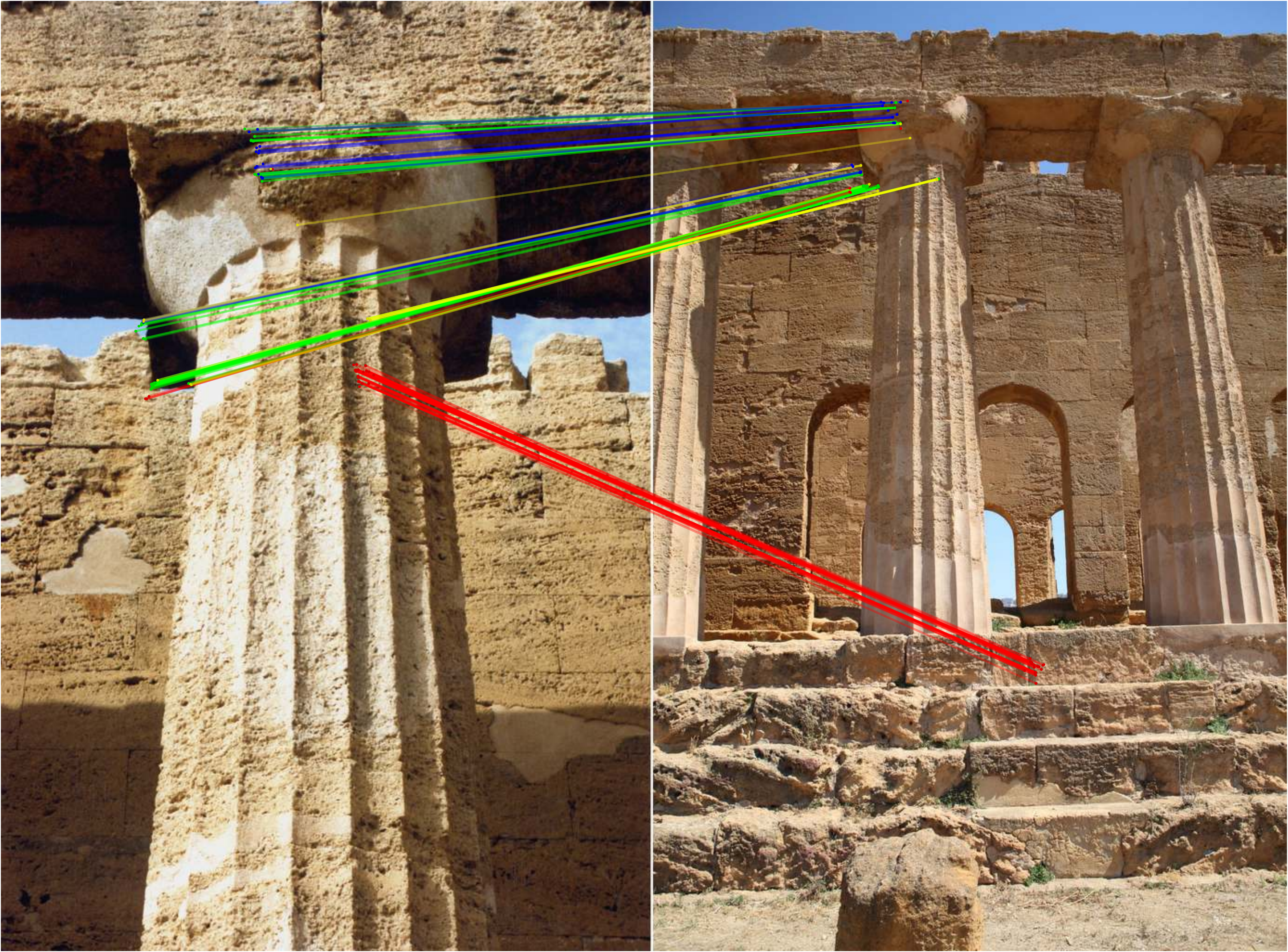} &
			\includegraphics[height=0.175\textwidth]{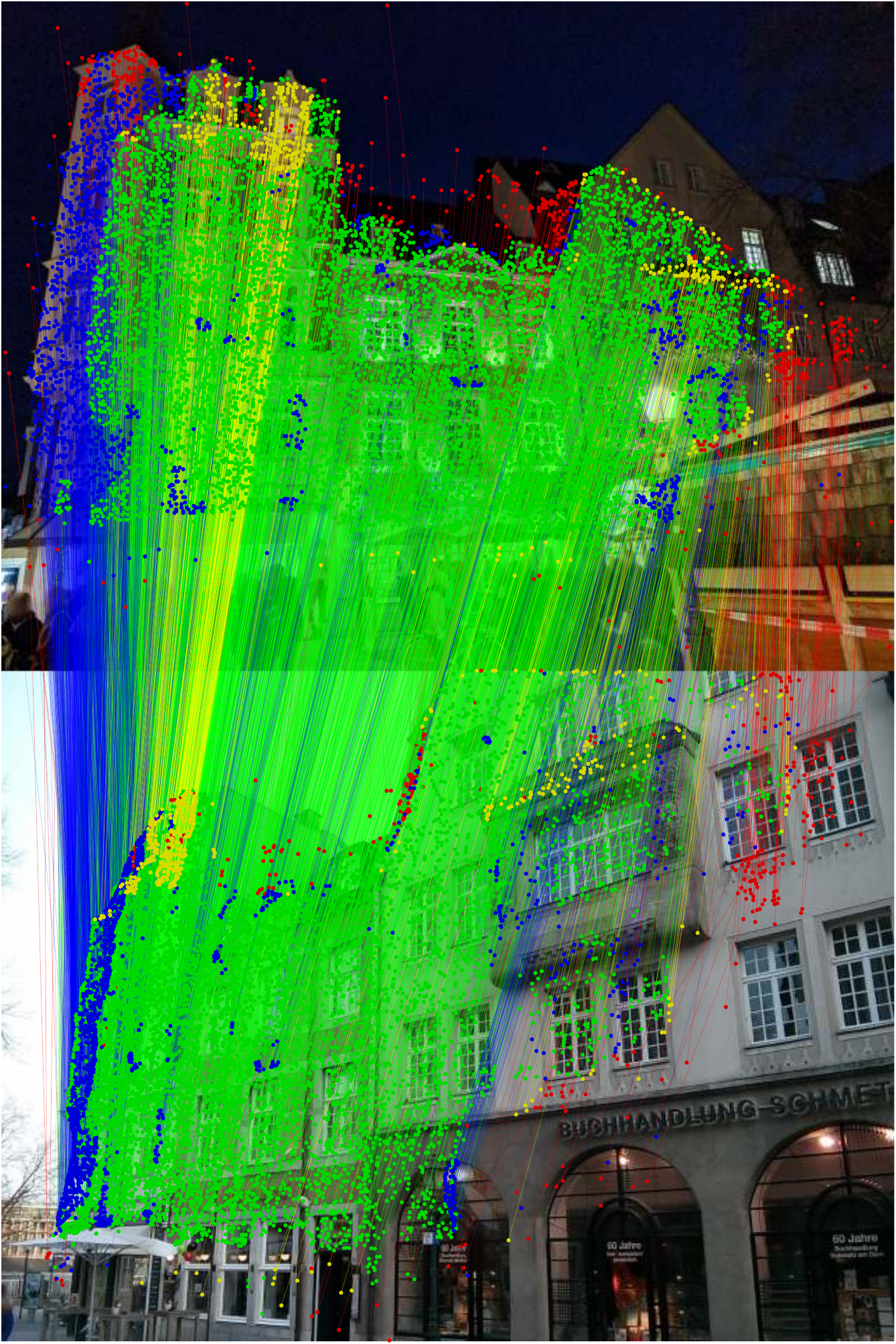} &
			\includegraphics[height=0.175\textwidth]{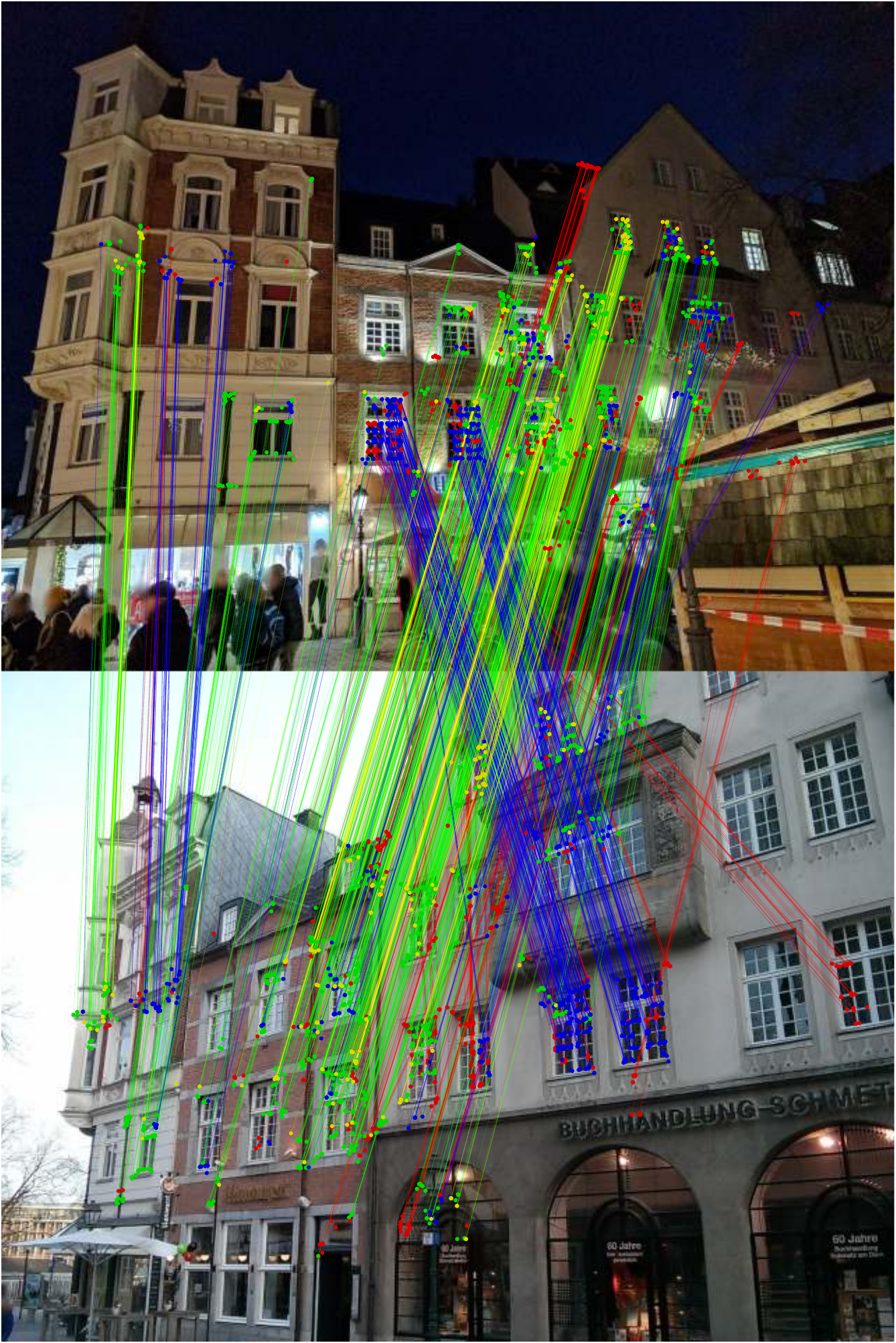} &
			\includegraphics[height=0.175\textwidth]{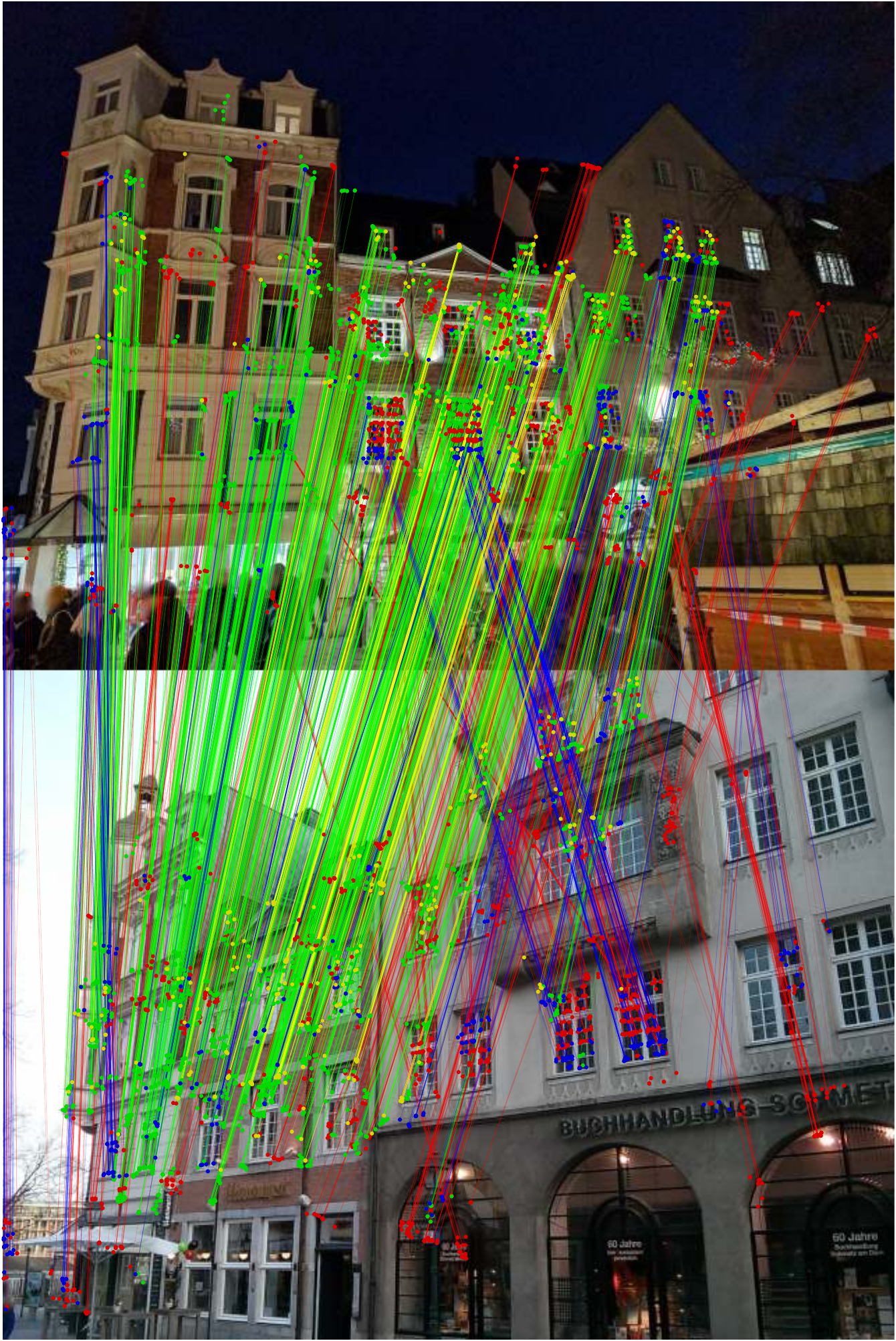} &
			\includegraphics[height=0.175\textwidth]{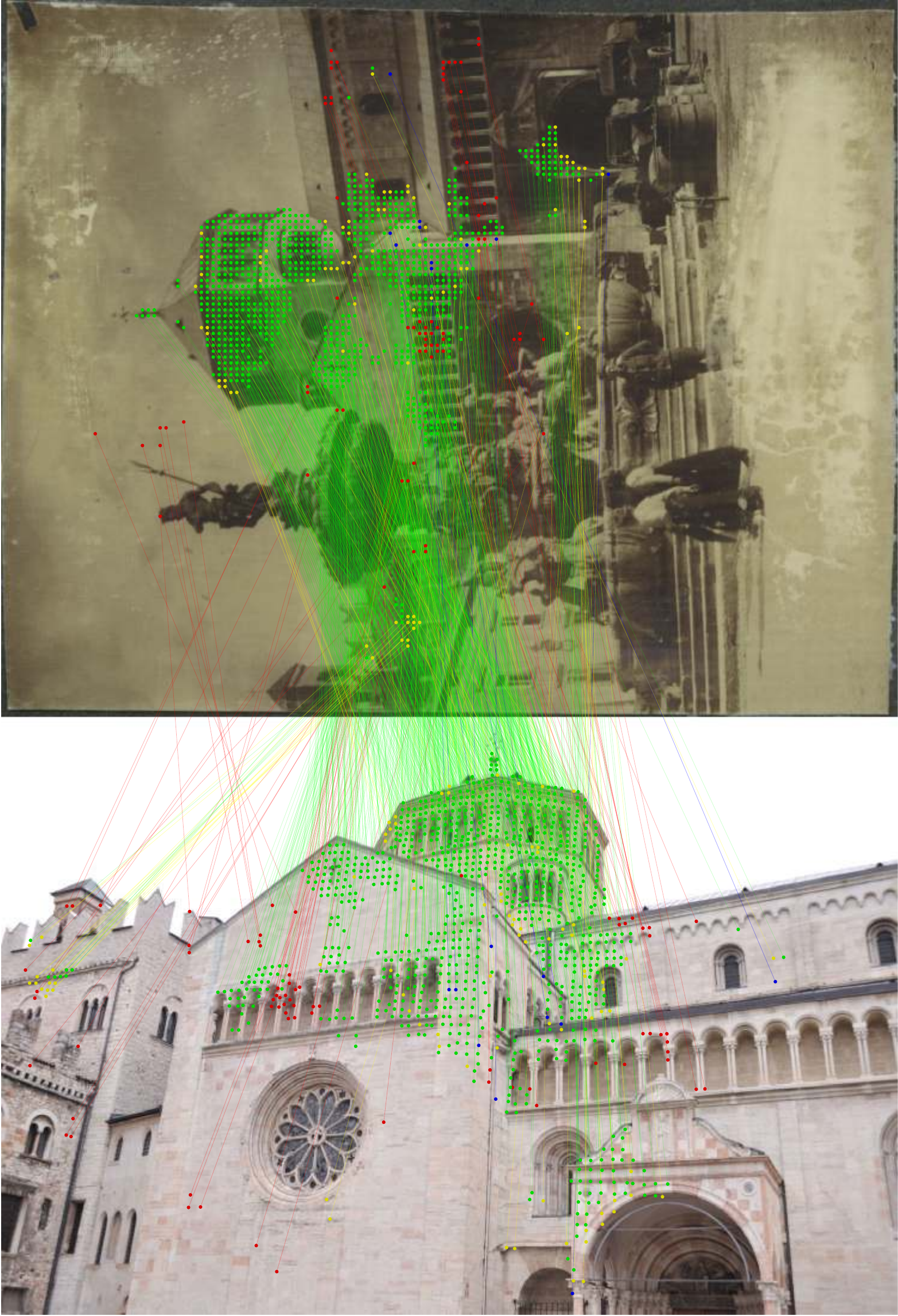} &
			\includegraphics[height=0.175\textwidth]{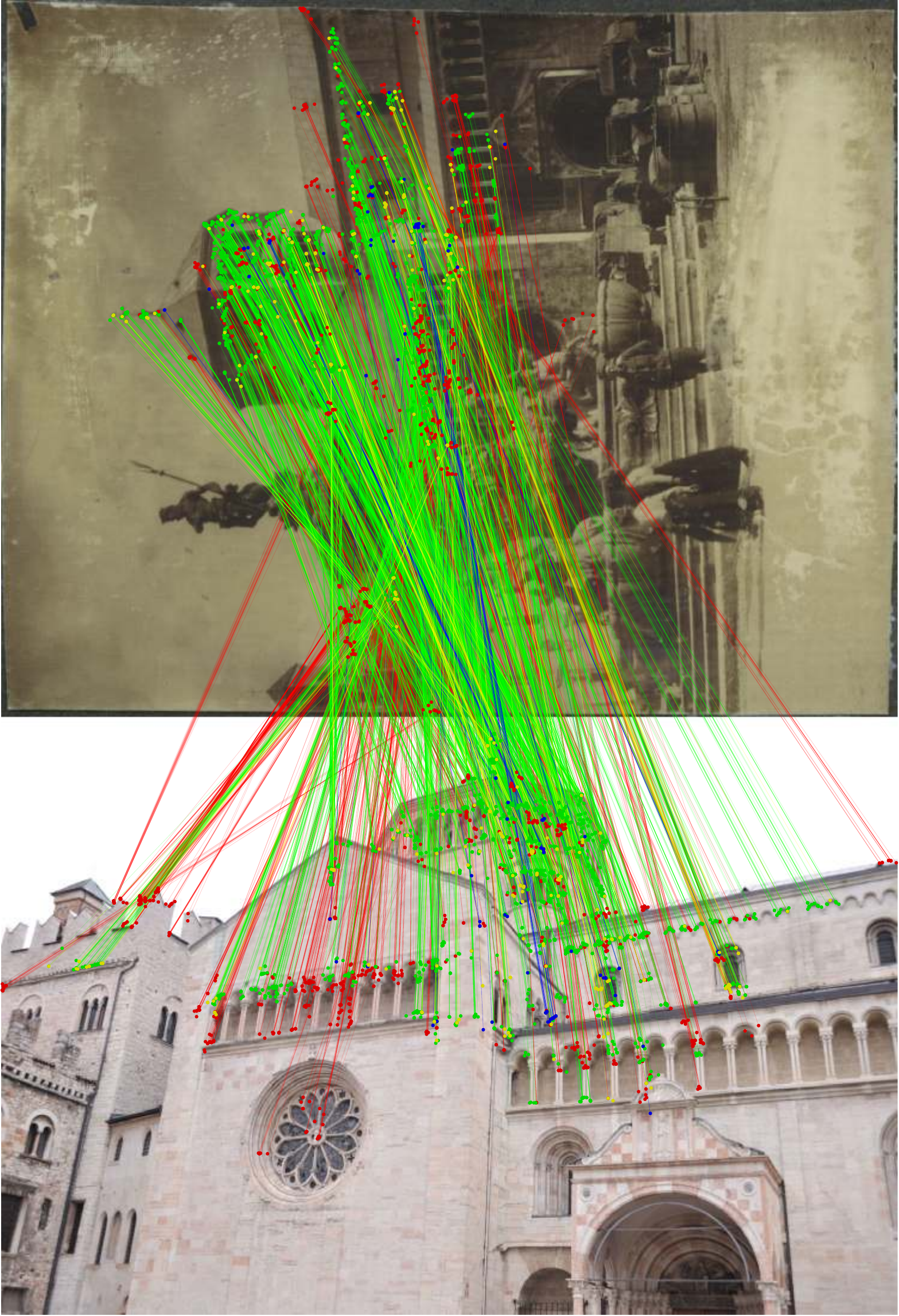} &
			\includegraphics[height=0.175\textwidth]{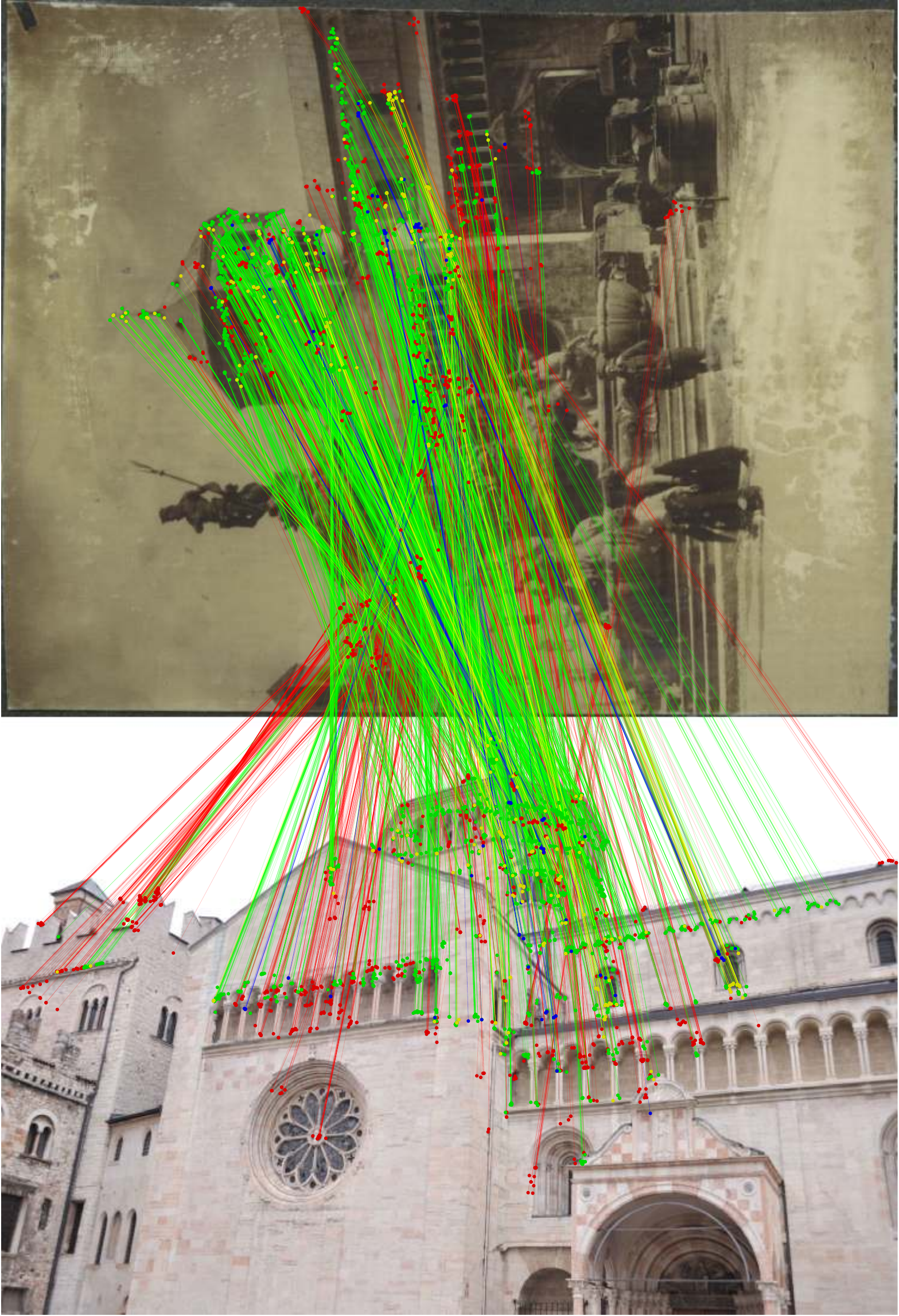} \\
			QuadTree Att. & SuperGlue & Slime$^\Rsh$ & DKM & Slime & Slime$^\Rsh$ & QuadTree. Att. & Slime & Slime$^\Rsh$		
		\end{tabular}
	}
	\caption{\label{planar_nonplanar_match}
		Visual results for challenging planar and non-planar pairs, the original top images of ZeroCalcare and NettunoDuomo have been rotated anticlockwise by 90$^\circ$ for an optimal fit in the frame. Correct matches are in green and yellow, wrong matches in red and blue. Matches after RANSAC best setup are in green and blue. In the planar case (top row), both LoFTR and QuadTree Attention cannot get correct matches for the ZeroCalcare and There pairs, and only Slime, Hz$^+$ and SuperGlue get correct matches for the There pair. The SunSeason pair presents ambiguous matches due to repeated structures which are resolved by RANSAC (see Sec.~\ref{psec}). In the non-planar case (bottom row), Slime is unable to retrieve any correct match only for the Temple pair, but orientation priors allows its upright counterpart to find some correct matches. For the DayNight pair Hz$^+$ can get only one correct match, while Slime based on it gives good results. Again, for the DayNight and NettunoDuomo pairs, orientation priors allow Slime$^\Rsh$ to detect more matches than Slime (see Sec.~\ref{npsec}). Best viewed in color and zoomed in.}\vspace{-0.5em}
\end{figure*}

About the failure cases, only ECO-TR, Key.Net and Slimed RootSIFT output no matches. For ECO-TR and Slimed RootSIFT the same motivations discussed above hold, for Key.Net this is due to its inability to find a good initialization for the specific input pairs. Slime strategy, which can be assimilated to fit multiple local models as for the AdaLAM employed with Key.Net, appears more robust considering this aspect. About the outputs with only wrong matches, RootSIFT and DISK, the latter with a consistent difference from the former, are the methods which fail more. All upright approaches, including end-to-end networks, fail on the four rotated pairs. While some accidental raw matches can be found, these are not robust and are removed by RANSAC, for instance in the case of DKM, see SM\ref{sm_res1}. Slimed Hz$^+$ has only one failure case, resolved by simply adding upright priors, since its upright implementation succeeds for the same scene (see Fig.~\ref{planar_nonplanar_match}), while Hz$^+$ has six failures. Moreover, Slimed Key.Net has no failure cases on the opposite of the base Key.Net which got in total ten failures.

\subsubsection{MegaDepth and ScanNet}\label{mdepth_scannet}
Table~\ref{megadepth_scannet_other} reports the AUC for MegaDepth and ScanNet at different error thresholds. Bars and colors are normalized within each dataset. Only results with Hz$^+$ are reported in the case of Slime since it was globally the best in the above precedent evaluations. Actually, before the OpenCV essential matrix estimation internal RANSAC, matches have been pre-filtered according to the fundamental matrix RANSAC implemented for Slime. Reported values are the best ones obtained, see SM\ref{sm_res2} for further details.

For Megadepth AUC proportionally increases according to the pose error threshold and in terms of rank ordering is quite similar to that defined for the non-planar dataset accuracy and in further benchmarks, see SM\ref{sm_res3}. Note that values can slightly vary with respect to those reported in~\cite{loftr,superglue,dkm}, probably due small differences in the setup and model weights. Upright Key.Net and Hz$^+$ counterparts provide better results than the general versions, maybe due to the nature of the two MegaDepth scenes, presenting only upright images which seem overall easier than those in the non-planar dataset when establishing matches. This is consistent with the slightly worse results of Slimed Hz$^+$ with respect to the base Hz$^+$, as the former outputs more but less accurate matches due to the lacking of any keypoint refinement in Slime, so the noise in the final pose estimation may increase.

For ScanNet AUC still increases proportionally with the pose error threshold, but values for deep approaches are lower than those reported in other works since, in order to have a fair comparison with non-deep approach, the same outdoor weights used for the evaluation on the other above datasets have been employed. Heuristics could be designed for non-deep approaches to exploit the specific constraints in terms of depth, baseline and rotation characterizing these indoor images, which are presumably learn from the training data by deep architectures. Anyways, while accuracy relative order is consistent with those on the datasets above, there is a more remarkable gap between recent end-to-end networks and other methods on ScanNet, greater than 10\% for the AUC@20$^\circ$, due to motion blur and lack of textures\hspace{-0.1em} in the scenes. \hspace{-0.1em}In this respect, Slimed Hz$^+$ is able to reduce by more than half the gap of the base Hz$^+$, approaching SuperGlue results and highlighting the matching robustness of Slime as further validated in SM\ref{sm_res4}.

\subsubsection{To summarize on Slime}
Overall, the Slime framework digs feasible raw matches at the expense of their localization accuracy. Keypoints are taken as extracted by the detector with no refinements along the process. This leads to stable and robust clusters of matches, increasing the current coverage and precision as defined in Sec.~\ref{err_metric}. Also, this indicates the ability for Slime to discover and retain matches in challenging scenarios, as suggested by the results on ScanNet (see Sec.~\ref{mdepth_scannet}), supplying at the same time to the lack of good seed matches in the guided filtering of the correspondences. As drawback, more poorly-localized matches are included when computing the pose, hence for non-complex scenes the estimation can be degraded by a limited amount of noise.  

Slime wrong matches are mainly due to the raw keypoint localization since the generated optical flow is feasible, or ambiguous and repeated structures as can happen for LoFTR. In the former case, the estimated pose is contaminated by noise, in the latter case the fusion of local clues with a global RANSAC makes the correct solution to prevail (see Fig.~\ref{planar_nonplanar_match}).

By its design, Slime suffers with tight-constrained base pipelines as it is eager of raw matches. This is reflected by the worse relative performances when combined with AdaLAM than the case it is used in conjuction with Hz$^+$ and RootSIFT, and in the slightly higher RANSAC post-processing threshold requirement. Nevertheless, Slime was able to limit Key.Net failure cases as seen for the non-planar dataset in Sec.~\ref{npsec}.

\subsubsection{Running times}
Experiments have been done with Ubuntu 20.04 running on an Intel Core I9 10900K with 64GB of RAM and equipped with a NVIDIA GeForce RTX 2080 Ti GPU. Excluding RANSAC post-processing, all methods run within 5 s on this middle-high level system with the exception of ECO-TR, requiring 25 s, Hz$^+$ and Slime. The average running time for Hz$^+$ goes from 7 s to 35 s, respectively on ScanNet ($640\times480$ px images) and MegaDepth (roughly $1200\times1024$ px images). The corresponding execution time for Slime goes from 1 to 3 min, which is relatively high, and is close to that of COTR, the original non-optimized ECO-TR architecture, which takes about 3 min for 8K matches~\cite{ecotr}. 

Nevertheless, it should be pointed out that both Hz$^+$ and Slime code is mainly written in Matlab, also mixing other languages without optimizations, with the aim of providing flexibility when experimenting and managing pipeline modules as standalone. Unlike handcrafted methods, deep networks are implicitly and transparently better optimized by standard libraries like PyTorch or TensorFlow.

Besides code porting, overhead removal, parallel multi-thread or GPU execution and further optimizations, it must be observed that code execution is proportional to the number of matches involved in the computation, up to 8K for Hz$^+$ and  around 64K for Slime. Many of the keypoints of these matches are almost duplicates due to the overlap between blocks or tiles, so their computation could be optimized. Indeed, as long as matches remain until the final pipeline stages, the execution slows down, meaning that easier image pairs take more time. Clustering together close matches and replacing them with a single weighted match (see, Sec.~\ref{block_match}) is a promising solution, leaved as future work, within the same spirit of~\cite{lightglue}. Please refer to SM\ref{sm_time} for more details on the Slime computational time.

\section{Conclusion and Future Work}\label{conclusions}
\subsection{Slime} 
This paper proposes Slime, a handcrafted image matching framework exploiting only local and global raw planar constraints to progressively expand and filter matches. Slime presents some similarities with AdaLAM. In particular, they both proceed by aggregating local cluster of matches, subject respectively to planar and affine constraints. Nevertheless, Slime initialization is more robust than that of AdaLAM and leads to less failure cases.

\begin{table}[t]
	\renewcommand{\arraystretch}{0}
	\setlength{\tabcolsep}{5pt}
	\centering
	\caption{AUC for MegaDepth and ScanNet (see Sec.~\ref{mdepth_scannet}).\\ Best viewed in color.}\label{megadepth_scannet_other}
	\resizebox{0.48\textwidth}{!}{
		\begin{tabular}{r<{}c<{}c<{}c<{}c<{}c<{}c}
			\toprule
			& \multicolumn{3}{c}{MegaDepth} & \multicolumn{3}{c}{ScanNet}\\
			\cmidrule(lr){2-4}\cmidrule(lr){5-7}
			& @5$^\circ$ (\%) & @10$^\circ$ (\%) & @20$^\circ$ (\%) & @5$^\circ$ (\%) & @10 (\%) & @20$^\circ$ (\%)\\
			\midrule
			RootSIFT\hphantom{$\Rsh$} & \Chart{27.63}{0.050}{blue}{25}{12}\hphantom{9999.} & \Chart{40.58}{0.261}{blue}{25}{12}\hphantom{9999.} & \Chart{52.18}{0.451}{blue}{34}{12}\hphantom{9999.} & \Chart{2.75}{0.043}{red}{25}{12}\hphantom{99999.} & \Chart{6.26}{0.097}{red}{25}{12}\hphantom{99999.} & \Chart{11.57}{0.179}{red}{25}{12}\hphantom{9999.} \\
			\rowcolor{gray!15}RootSIFT$^\Rsh$ & \Chart{27.37}{0.045}{blue}{25}{12}\hphantom{9999.} & \Chart{39.91}{0.251}{blue}{25}{12}\hphantom{9999.} & \Chart{52.20}{0.452}{blue}{34}{12}\hphantom{9999.} & \Chart{3.18}{0.049}{red}{25}{12}\hphantom{99999.} & \Chart{7.38}{0.114}{red}{25}{12}\hphantom{99999.} & \Chart{13.56}{0.210}{red}{25}{12}\hphantom{9999.} \\
			Key.Net\hphantom{$\Rsh$} & \Chart{40.94}{0.267}{blue}{25}{12}\hphantom{9999.} & \Chart{59.02}{0.563}{blue}{34}{12}\hphantom{9999.} & \Chart{73.59}{0.801}{blue}{53}{12}\hphantom{9999.} & \Chart{9.69}{0.150}{red}{25}{12}\hphantom{99999.} & \Chart{20.95}{0.325}{red}{25}{12}\hphantom{9999.} & \Chart{33.65}{0.522}{red}{34}{12}\hphantom{9999.} \\
			\rowcolor{gray!15}Key.Net$^\Rsh$ & \Chart{44.01}{0.318}{blue}{25}{12}\hphantom{9999.} & \Chart{61.71}{0.607}{blue}{44}{12}\hphantom{9999.} & \Chart{75.57}{0.834}{blue}{53}{12}\hphantom{9999.} & \Chart{9.91}{0.154}{red}{25}{12}\hphantom{99999.} & \Chart{20.94}{0.325}{red}{25}{12}\hphantom{9999.} & \Chart{34.19}{0.530}{red}{34}{12}\hphantom{9999.} \\
			Hz$^+$\hphantom{$\Rsh$} & \Chart{45.52}{0.342}{blue}{25}{12}\hphantom{9999.} & \Chart{62.50}{0.620}{blue}{44}{12}\hphantom{9999.} & \Chart{75.56}{0.834}{blue}{53}{12}\hphantom{9999.} & \Chart{10.77}{0.167}{red}{25}{12}\hphantom{9999.} & \Chart{22.38}{0.347}{red}{25}{12}\hphantom{9999.} & \Chart{34.69}{0.538}{red}{34}{12}\hphantom{9999.} \\
			\rowcolor{gray!15}Hz$^{+\Rsh}$ & \Chart{46.88}{0.365}{blue}{25}{12}\hphantom{9999.} & \Chart{63.96}{0.644}{blue}{44}{12}\hphantom{9999.} & \Chart{77.02}{0.857}{blue}{62}{12}\hphantom{9999.} & \Chart{10.58}{0.164}{red}{25}{12}\hphantom{9999.} & \Chart{22.01}{0.341}{red}{25}{12}\hphantom{9999.} & \Chart{34.97}{0.543}{red}{34}{12}\hphantom{9999.} \\
			Slimed Hz$^+$\hphantom{$\Rsh$} & \Chart{42.49}{0.293}{blue}{25}{12}\hphantom{9999.} & \Chart{60.14}{0.581}{blue}{34}{12}\hphantom{9999.} & \Chart{74.09}{0.809}{blue}{53}{12}\hphantom{9999.} & \Chart{14.69}{0.228}{red}{25}{12}\hphantom{9999.} & \Chart{29.28}{0.454}{red}{34}{12}\hphantom{9999.} & \Chart{44.35}{0.688}{red}{44}{12}\hphantom{9999.} \\
			\rowcolor{gray!15}Slimed Hz$^{+\Rsh}$ & \Chart{41.80}{0.281}{blue}{25}{12}\hphantom{9999.} & \Chart{59.43}{0.570}{blue}{34}{12}\hphantom{9999.} & \Chart{73.98}{0.808}{blue}{53}{12}\hphantom{9999.} & \Chart{14.10}{0.219}{red}{25}{12}\hphantom{9999.} & \Chart{27.82}{0.432}{red}{34}{12}\hphantom{9999.} & \Chart{42.02}{0.652}{red}{44}{12}\hphantom{9999.} \\
			DISK\hphantom{$\Rsh$} & \Chart{41.57}{0.278}{blue}{25}{12}\hphantom{9999.} & \Chart{57.79}{0.543}{blue}{34}{12}\hphantom{9999.} & \Chart{70.56}{0.752}{blue}{53}{12}\hphantom{9999.} & \Chart{9.59}{0.149}{red}{25}{12}\hphantom{99999.} & \Chart{20.30}{0.315}{red}{25}{12}\hphantom{9999.} & \Chart{34.02}{0.528}{red}{34}{12}\hphantom{9999.} \\
			\rowcolor{gray!15}SuperGlue\hphantom{$\Rsh$} & \Chart{45.67}{0.345}{blue}{25}{12}\hphantom{9999.} & \Chart{62.79}{0.625}{blue}{44}{12}\hphantom{9999.} & \Chart{76.99}{0.857}{blue}{62}{12}\hphantom{9999.} & \Chart{14.93}{0.232}{red}{25}{12}\hphantom{9999.} & \Chart{30.86}{0.479}{red}{34}{12}\hphantom{9999.} & \Chart{47.94}{0.744}{red}{53}{12}\hphantom{9999.} \\
			LoFTR\hphantom{$\Rsh$} & \Chart{39.30}{0.241}{blue}{25}{12}\hphantom{9999.} & \Chart{57.32}{0.535}{blue}{34}{12}\hphantom{9999.} & \Chart{72.40}{0.782}{blue}{53}{12}\hphantom{9999.} & \Chart{18.37}{0.285}{red}{25}{12}\hphantom{9999.} & \Chart{35.45}{0.550}{red}{34}{12}\hphantom{9999.} & \Chart{51.97}{0.806}{red}{53}{12}\hphantom{9999.} \\
			\rowcolor{gray!15}SE2-LoFTR\hphantom{$\Rsh$} & \Chart{50.44}{0.423}{blue}{25}{12}\hphantom{9999.} & \Chart{67.11}{0.695}{blue}{44}{12}\hphantom{9999.} & \Chart{79.60}{0.900}{blue}{62}{12}\hphantom{9999.} & \Chart{18.83}{0.292}{red}{25}{12}\hphantom{9999.} & \Chart{36.45}{0.565}{red}{34}{12}\hphantom{9999.} & \Chart{53.76}{0.834}{red}{53}{12}\hphantom{9999.} \\
			MatchFormer\hphantom{$\Rsh$} & \Chart{43.86}{0.315}{blue}{25}{12}\hphantom{9999.} & \Chart{60.94}{0.594}{blue}{34}{12}\hphantom{9999.} & \Chart{74.80}{0.821}{blue}{53}{12}\hphantom{9999.} & \Chart{16.44}{0.255}{red}{25}{12}\hphantom{9999.} & \Chart{32.81}{0.509}{red}{34}{12}\hphantom{9999.} & \Chart{49.63}{0.770}{red}{53}{12}\hphantom{9999.} \\
			\rowcolor{gray!15}QuadTree Att.\hphantom{$\Rsh$} & \Chart{51.12}{0.434}{blue}{34}{12}\hphantom{9999.} & \Chart{67.76}{0.706}{blue}{44}{12}\hphantom{9999.} & \Chart{80.52}{0.915}{blue}{62}{12}\hphantom{9999.} & \Chart{21.21}{0.329}{red}{25}{12}\hphantom{9999.} & \Chart{39.05}{0.606}{red}{44}{12}\hphantom{9999.} & \Chart{55.70}{0.864}{red}{62}{12}\hphantom{9999.} \\
			DKM\hphantom{$\Rsh$} & \Chart{56.43}{0.521}{blue}{34}{12}\hphantom{9999.} & \Chart{71.87}{0.773}{blue}{53}{12}\hphantom{9999.} & \Chart{82.96}{0.955}{blue}{62}{12}\hphantom{9999.} & \Chart{24.37}{0.378}{red}{25}{12}\hphantom{9999.} & \Chart{44.27}{0.687}{red}{44}{12}\hphantom{9999.} & \Chart{61.52}{0.954}{red}{62}{12}\hphantom{9999.} \\
			\bottomrule
		\end{tabular}	
	}\vspace{-1.5em}
\end{table}

Slime greatly improves the coverage of Hz$^+$ and its ability to get robust matches in complex scenes. However, as rough matches are included, in non-complex scenes the level of noise in the pose estimation can slightly increase. After all, if the keypoint pixel localization is accurate, only a limited number of matches are required to get the scene configuration, which is the SuperGlue strategy. Future work will investigate solutions to this problem introducing keypoint refinement within Slime, as the strategy recently presented in~\cite{miho}.

In the light of this, Slime can be associated to the coarse level of LoFTR, and shows that it is still possible to provide an explainable handcrafted solution to the matching problem. It is not a case that Slime and LoFTR can have similar failure cases on fake planes caused by repeated or similar patterns. Future research will focus on solutions to this kind of scenes.

Slime framework analysis also shows that matching pipelines need a feedback system linking the different modules in order to improve the detected matches. Slime accomplishes this by progressively moving through its planar intermediate representation from local to global matching, while recent end-to-end deep image matching networks use attention and transformers. Incorporating the planar constraints and filtering strategies adopted by Slime into deep image matching can be a future line of research.

The feedback mechanism is also employed by Slime for rotation invariance. The relative orientation of the plane clusters are estimated, and matches are filtered and expanded in accordance, progressively in a coarse-to-fine manner. Current deep solutions consider patch orientation only individually, as in SE2-LoFTR, and would benefits of the Slime approach. 

Lastly, Slime roughly estimates epipolar geometry by pairwise planes and remove matches on the basis of their stability to the epipolar constraints found in this process. This approach relies on RANSAC ideas but focuses on the individual match robustness on close epipolar configuration and not on the model with the highest consensus. Future work will also further investigate this strategy.

\vspace{-0.5em}
\subsection{Comparative evaluation}
As further contribution, a comparative evaluation considering the latest SOTA is proposed. Recall is replaced by coverage for taking into account semi-dense methods, while precision and accuracy mainly measure the quality of matches at the coarse and fine levels, respectively. The analyzed datasets consider both planar and non-planar scenes, non-upright image pairs and a wide range of different scenarios including temporal, viewpoint and illumination changes.  

According to this evaluation, Slime can be an effective handcrafted method to reach the same level of coverage of semi-dense approaches, i.e LoFTR-based ones, ECO-TR and DKM. Other sparse approaches, including handcrafted and hybrid pipelines together with the end-to-end deep networks SuperGlue and DISK, notability differ in values. This is more evident in the case of non-planar scenes.

In the planar case Key.Net, Hz$^+$ and Slime give overall the best matches, in term of both precision and accuracy. On the other hand, in the non-planar case QuadTree Attention and DKM provide the best matches. Still on non-planar images, SuperGlue and semi-dense methods reach globally better accuracy levels than other sparse approaches. These results are aligned with those on MegaDepth and ScanNet.

The comparative analysis also shows that strong viewpoint changes mixed with other image transformations, historical temporal changes and repeated patterns remain challenging conditions to be faced for which only partial solutions are currently available. Moreover, there is not yet a stable approach to make end-to-end matching method capable to naturally deal with non-upright image pairs. Finally, RANSAC post-processing is still mandatory to get a robust and stable solution. 

The outcomes of this analysis can be employed to identify further research directions and problems worth to be investigated. As future work, the evaluation datasets can be extended to include more challenging scenarios, as well as additional recent image matching approaches.\vspace{-0.5em}


\bibliographystyle{IEEEtran}
\bibliography{slime_abbr}


\vfill
\pagebreak

\renewcommand\appendixname{Supplemental Material (SM)}
\appendices

\section*{Supplemental Material (SM)}

\subsection{High-resolution figures}\label{high_res_fig}
Paper high-resolution figures can be found \href{https://unipa-my.sharepoint.com/:f:/g/personal/fabio_bellavia_unipa_it/ErqK2tFObGhBgMXc92Jt018BOOgPpw_uBO76wxOK8AywMw?e=nrS4p0}{here}. Please note that the quality of the rendering, especially for what concerns transparencies, depends on the viewer employed.

\subsection{Compared method setups}\label{sm_methods}
Authors' original implementations are used for the evaluation with the exception of RootSIFT, Key.Net and LoFTR, whose code mainly come from the \href{https://kornia.github.io/}{Kornia library}. All methods use the default settings and outdoor model weights.

Slime framework parameters are chosen experimentally in accordance to previous works, i.e.~\cite{sgloh2} for what concerns angular thresholds and orientation histogram binning, and~\cite{dtm,hz_pipeline} for what concerns spatial thresholds. Base pipelines employed in Slime also use their default settings. Unlike Slimed Hz$^+$, which runs at the end of the pipeline DTM globally, Slime Key.Net does not run the equivalent AdaLAM globally as last stage, since it was observed that this is mostly equivalent to running the base Key.Net alone but with tight threshold values due to the tighter constraints of AdaLAM.

Input images for MatchFormer and QuadTree Attention are resized so that the maximum size does not exceed 1200 px to avoid GPU out-of-memory issues, and a padding is added for the end-to-end methods requiring input image size as a multiple of 16. The NVIDIA GeForce RTX 2080 Ti GPU employed in the evaluation runs frequently in out-of-memory for ECO-TR, but no simple general solutions to avoid this situation has been found.

One of the aim of this paper is to show that also non-deep and hybrid pipelines are able to work in challenging scenarios. That said, although a long-standing tradition of non-deep and hybrid matching pipelines exist, incorporating several keypoint detectors, descriptors, and local constraint filters like LPM and GMS, in the view of previous analyses~\cite{dtm,imw2020} and of the image matching challenges\hyperref[imclink]{\footnotemark[2]}, their results do not generally surpassed those of the Hz$^+$ and Key.Net pipelines. For these reason, besides the baseline RooSIFT, these two hybrid pipelines are the only two considered in the comparison.

Concerning handcrafted keypoint extractors, detectors based on DoG, e.g. SIFT, and on the Harris corners are those providing according to the author's experience better matchability and repeatabiltiy in case of perspective distortions and illumination changes. Specifically, DoGs provide a slight better keypoint localization while Harris corners are slightly more discriminant. Roughly speaking, they both detect similar regions, but the former pushes the keypoint location more towards the flatness area of the patch, which is more stable yet less discriminant.

Regarding patch pre-processing before computing the descriptor, affine patch normalization marks the difference with the common scale and rotation patch normalization of SIFT in scenes with strong viewpoint changes, otherwise they obtain quite similar results.
 
Concerning the descriptor, other standalone best solutions are in line with HardNet, which also provides a very robust patch affine normalization, and none of them is actually handcrafted. Being descriptor extractors designed to find features which depend on the patch data, the author does not personally think that this is in contrast with the aim of designing an image matching method as-handcrafted-as-possible, yet future work in this sense are planned.

Finally, about local spatial filters, according to the author's direct experience in~\cite{dtm}, the performances of top methods are in line with those of AdaLAM, excluding DTM which has less constraints, higher recall and maybe a slightly lower precision. Nevertheless these DTM characteristics fit well with the Slime philosophy of elaborating raw matches.    

The code employed in the evaluation is available \href{https://unipa-my.sharepoint.com/:f:/g/personal/fabio_bellavia_unipa_it/EktcfLJkX_1Es2J7m6G3BiIBZQlmXFCzftcMmMuoUGPBHQ?e=fDdmjx}{here}.

\subsection{RANSAC implementations}\label{sm_ransac}
The RANSACs employed inside Slime to filter planar homographies and to post-process all image matching methods are based on the \href{https://www.peterkovesi.com/matlabfns/}{Matlab code by Peter Kovesi}.
\subsubsection{Planar homography estimation} This RANSAC is used internally by Slime. The minimum and maximum number of sampled models is set to 100 and 10K, respectively. A planar homography $H_i$ is computed at each RANSAC iteration $i$ by the normalized DLT algorithm from four sampled correspondences. If the second minimum eigenvalue $\epsilon_D$ of the DLT matrix is less than 0.05, $H_i$ is marked as degenerate and the rest of the iteration skipped. The above condition implies that the rank of the associated homogeneous system is more close to $n-2$ than $n-1$; notice that $\epsilon_D$ can be an absolute value since points have been normalized. Otherwise, inliers are defined following Eq.~\ref{max_reproj} and filtered according to the plane side, see Sec.~\ref{exp_prune}.

If the number of filtered inliers for the $H_i$ model is greater than the current best one, the maximum iteration is updated accordingly. The probability of at least one sample free by outliers is fixed to $p=0.9$. In case the maximum number of iterations is reached, the loop is stopped. The homography corresponding to the model with the highest number of inliers is retained and refined by DLT using the whole inlier set in the least square sense, re-updating the final list of inliers accordingly.
\subsubsection{Fundamental matrix estimation} This is the post-processing RANSAC used after each image matching method. The minimum and maximum number of sampled models is set to 200 and 10K, respectively. At each iteration a fundamental matrix $F_i$ is computed using the eight-point algorithm sampling from the match set. If the second minimum eigenvalue of the associated DLT matrix is less than 0.05, $F_i$ is marked as degenerate. In this case, a planar homography $H_i$ is extracted from the eight sampled correspondences. If $H_i$ is also degenerate the rest of the iteration is skipped. Otherwise, inliers are computed as above in case of the planar model, or according to the maximum epipolar error defined as in Eq.~\ref{max_epi_error}.

The best homography and fundamental matrix are both retained and independently updated at each iteration. When a new top-scored inlier number is found, the maximum iteration is updated accordingly. When the maximum iteration is exceeded, starting from the best homography model a further RANSAC is executed by sampling two correspondences from the out-of-plane matches so as to get a fundamental matrix~\cite{multiview}. The top model in terms of inliers between the two models, i.e. the best one from $F_i$ and the one from the best $H_i$ plus two correspondences, are then refined and the best overall model is provided as final output.

\subsection{Datasets}\label{sm_p_np}
\subsubsection{Planar dataset} This dataset includes 26 different scenes with six images pair each, plus a further scene with only two aerial historical images with high temporal variation from~\cite{historical_aerial}, see Fig.~\ref{pd}. Excluding this latter pair, within each scene one image is fixed as reference, so that five image pairs are used for the evaluation. This dataset extends the one proposed in~\cite{dtm} with scenes from HPatches.

Specifically, ten scenes come from HPatches~\cite{hpatches} and present only viewpoint changes, four scenes come from the Oxford dataset, two of them including viewpoint changes and the others scale and rotation changes simultaneously, five scenes are from~\cite{learning_ori} and present a combination of viewpoint changes and rotation, six scenes come from~\cite{sgloh2} and include viewpoint changes concurrently with illumination, blur, aliasing, scale and rotation changes, and the last scene is new and presents high viewpoint and rotation variations together.

Ten images pairs of these $26\times5+1=131$ pairs are non-upright and their upright counterparts are provided too. There are in total $131+10=141$ image pairs, of which about $8\%$ are designed to do not work with upright methods. Image resolutions range from $640\times480$ px to $1024\times768$ px. The image pairs of the planar dataset are available \href{https://unipa-my.sharepoint.com/:f:/g/personal/fabio_bellavia_unipa_it/ErBa9ao5L5hNlQ_LWvO9C4EBqysAenXz2NBBDH-ueH0bfw?e=P2CERp}{here}.
\subsubsection{Non-planar dataset} This dataset includes 92 different scenes with up to three image pair for each scene, for a total of 142 image pairs, plus four upright image pairs obtained from the original non-upright image pairs as for the planar dataset, see Fig.~\ref{npd}. The non-planar dataset extends the one employed in~\cite{dtm}, already containing among the others scenes from the Stretcha and DTU datasets, and others scene quite known in the computer vision community, with more than 30 further scenes. These included scenes come mainly from the Aachen Day-Night dataset, the IMC-PT datasets, from the challenging scene of~\cite{wxbs}, from the complex historical image pairs of~\cite{low3d} and from other scenes of interest for cultural heritage of~\cite{cultural_heritage}.

After a preliminary evaluation, the four Turntable scenes originally included in~\cite{dtm} were removed since obtained by moving the turntable leaving the background fixed, thus providing an inconsistent stereo setup. With respect to the past, the background inconsistency cannot be neglected because recent semi-dense matching methods, such as LoFTR and DKM, cover the background with matches that cannot be checked according to the stereo configuration of the scene.

Overall, the variety of scenes provides several challenging scenarios, with a low overlap with the training data of the evaluated deep methods, mostly limited to the IMC-PT and MegaDepth datasets, hence better able to provide insight of the generic real performances. Furthermore, since only used for evaluation and not for training, the analysis provided by this dataset is more general. The non-planar dataset images are available \href{https://unipa-my.sharepoint.com/:f:/g/personal/fabio_bellavia_unipa_it/EiZnOZXPZpZJlNuP6WJvuDMBgUH4SbYDOMEUTG7Zqar6Wg?e=czKe0h}{here}.

\subsection{Error metrics}\label{sm_metric}
\subsubsection{Coverage} The coverage definition extends the recall so that close keypoints count less to avoid misleading results in case of accumulated keypoints into clusters. The raw coverage of an image is computed as the area covered by the keypoints of the correct matches, see Fig.~\ref{recall}. This area is obtained by considering the binary mask of the correct keypoints and then applying a morphological dilation with a disk of radius $t_\perp$. The raw image coverage is normalized by the area of all the valid points that can be mapped from one image to the other.

In case of planar scenes, the normalization factor is computed by checking the boundary constraints when reprojecting image points, while in the case of non-planar scenes it is approximated by considering the raw image coverage of the union of the whole correct matches among all the evaluated methods plus the hand-taken matches use to compute the GT as discussed later. The final coverage is the minimum of the normalized coverages between the two images.  
\subsubsection{Accuracy} 
\begin{figure}[h!]
	\center
	\includegraphics[width=0.3\textwidth]{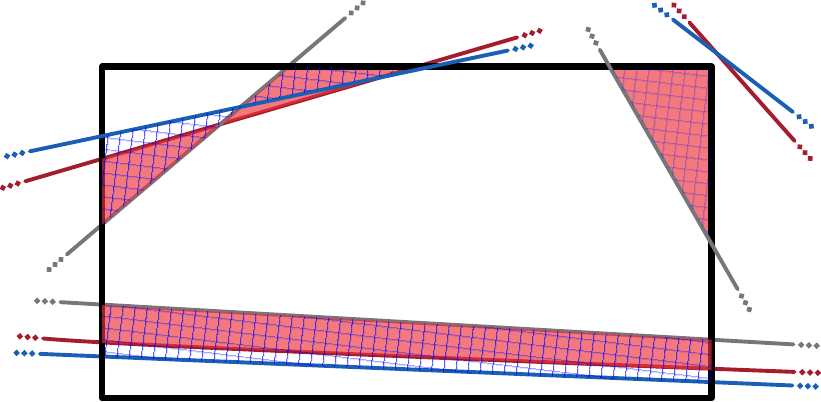} \\
	\caption{\label{epi_err_}
		Given pairs of lines, e.g. those constituted by the three gray and red line couples in the image, the proposed accuracy considers the normalized area on the image enclosed by them, indicated in red. These areas varies proportionally with the perturbation of the lines, as can be seen in the example by replacing the red lines with the blue ones, where the areas are indicated with blue patterns (see SM\ref{sm_metric}). Best viewed in color and zoomed in.}
\end{figure}

In the case of non-planar scenes, the proposed accuracy definition is derived from~\cite{noransac}. It basically consists of sampling non-occluded points in the image and considering the average of the areas enclosed by the GT and the tested epipolar lines corresponding to the given points, see Fig.~\ref{epi_err_}. This area value is divided by the image diagonal to get a linear error analogous to the average reprojection error employed for planar scenes.

Unlike the angular distance between the GT and the tested epipolar lines, this weighted error has the advantage to stay on the images, as virtual points outside the image canvas are not be taken into account. Furthermore, the proposed non-planar accuracy metric is robust to noisy fundamental matrices, and requires the same kind data for both the GT and the tested matching pipeline, i.e. a set of matches to compute the fundamental matrix. Notice also that the area enclosed by tow lines is much more stable than the point of intersection within lines. In this way the measured value will be consistent as long as the hand-taken matches are visually reasonable and cover uniformly the whole allowable scene area.

In contrast to the camera pose estimation, this metric can be computed when only one image pair is available, and for challenging images where SfM could fail, as long as a expert user provides hand-taken correspondences with a sufficient localization precision. Lastly, the proposed accuracy metric requires only to compute the fundamental matrix and not to decompose it (actually the essential matrix gets decomposed), decreasing the number of steps of the process, each of which incrementally accumulates noise from the previous one.

MegaDepth and ScanNet benchmarks consider instead the AUC of the pose error estimated by decomposing the essential matrix into the camera rotation matrix and the translation vector, needing an accurate estimation of the cameras by SfM, integrated with further sensor data for ScanNet. In the absence of loop-closure, as many scenes in MegaDepth or IMC-PT, SfM camera registration strongly depends on the camera model employed as well as the density and the distribution of the matches, which can lead to banana effect undetectable by qualitative visual inspections as in the Paestum  Wall scene reported in~\cite{kfc}, for which the camera position gets wrongly minimized by the non-optimal solution.

Furthermore, in the general case only the angular component of the translation vector can be taken into account and not its magnitude, i.e. the baseline length. As discussed in~\cite{heb}, the translation angular error has a minor impact in the pose estimation as the baseline length decreases. This aspect should be taken into account especially in case of a large depth-of-field (as for MegaDepth and IMC-PT), but it is not considered in any way in the usual definition of pose error in terms of the maximum angular error between the rotation and translation components. According to these observations, the mean Average Accuracy (mAA)~\cite{imw2020} as an aggregate measure distinguishing two distinct thresholds for each pose error component and counting when both are concurrently satisfied would be more appropriate, being understood that a different threshold value for each component should be required. 

That said, it is the author's opinion that as long as reasonable thresholds are used to be tolerant to noise, i.e. greater than the GT precision, all the discussed error metrics are valid alternatives. This is confirmed by the good correlation within the metrics for the results analyzed in Sec.~\ref{evaluation}.

\subsection{Ground-truth estimation}\label{sm_gt}
GT hand-taken correspondences have been taken using the \texttt{find\_corr\_gui} Matlab Graphical User Interface (GUI) application designed by the author and available within the released code. The application interface allows to easily navigate and zoom into the images, select the corresponding points and edit them. The tool also visualizes on the fly the estimated epipolar line or homography reprojected point in one image when a point in the other image is clicked (at least eight and four matches must be present to satisfy respectively the minimum model). Clearly, in order to avoid degenerate model configurations, one has to avoid to choose only aligned and coplanar correspondences for the planar and non-planar case, respectively.

In this way the user can visually check the current model, but also better localize further matches in ambiguous image areas when the current estimation is adequate. It is also possible to combine match sets, e.g. estimating more distinct planar homographies and then merging the match sets to get the fundamental matrix to guide the insertion of more out-of-the-planes matches through visual epipolar check. The tool can also mark occluded points.

The estimation of the GT planar homography or fundamental matrix inside this tool is also based on the Peter Kovesi's code and does not consider the radial distortion. The expected localization error introduced by the user when taking matches is within 3 px, and the one added by the radial distortion within 5 px. The error metrics defined in Sec.~\ref{err_metric} are based on aggregated reprojection errors where the threshold tolerance is $t_\perp=15$ px, robust enough to GT measurement errors.

\subsubsection{Planar dataset} 
\begin{figure}[t!]
	\center
	\includegraphics[width=0.45\textwidth]{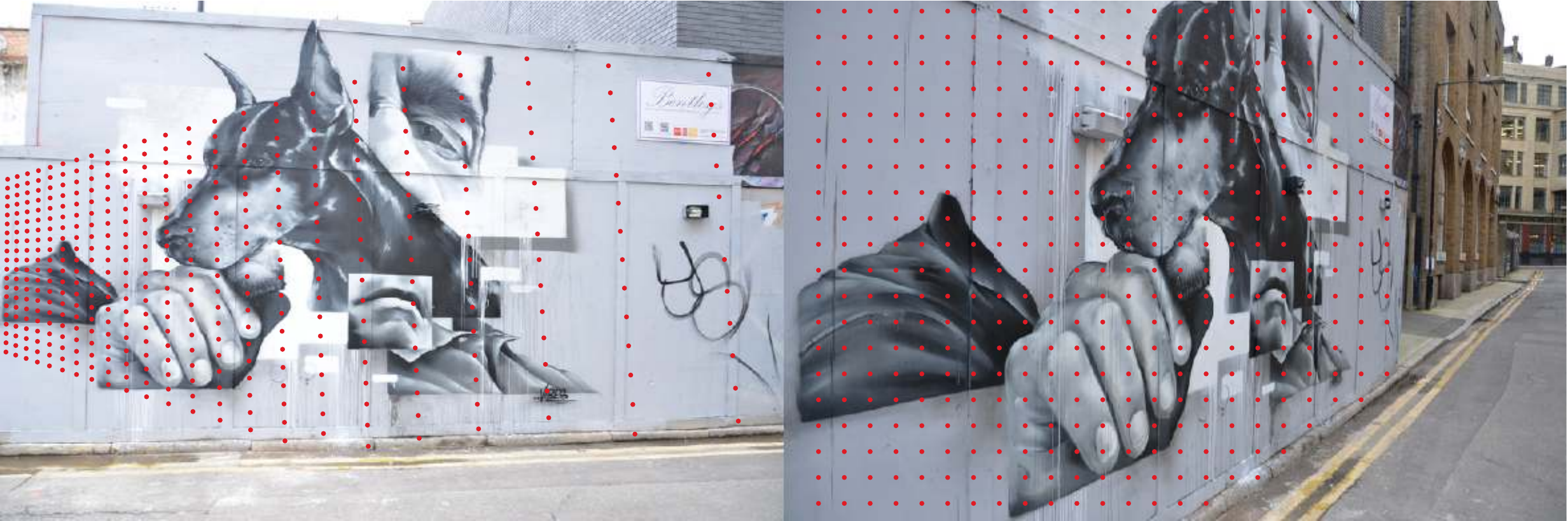} \\
	\vspace{0.3em}
	\includegraphics[width=0.45\textwidth]{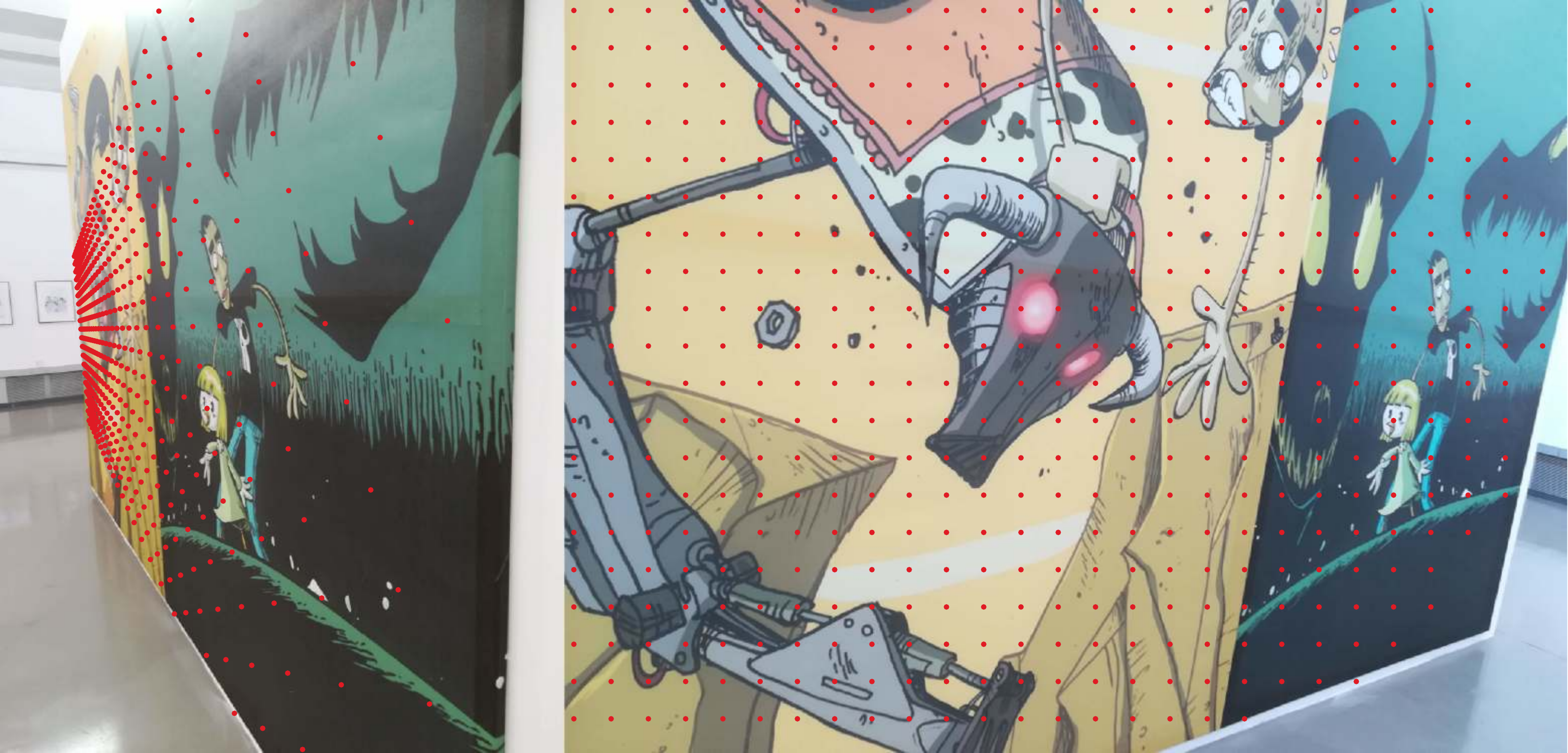}
	\caption{\label{gt_hom}
		\hspace{-0.5em}Visual qualitative example of estimated GT homographies as sampled grid point correspondences overlayed on the images (see SM\ref{sm_gt}). Best viewed in color and zoomed in.}
\end{figure}

Actually, almost all the image pairs come with GT homography matrices with the exception to two scenes (the new ZeroCalcare scene, see Fig.~\ref{planar_nonplanar_match}, and the historical aerial image pair from~\cite{historical_aerial}) and do not need to compute GT matches. Nevertheless, some image pairs from the scenes in~\cite{affine_eval,learning_ori} presented visible inaccurate reprojections for points far from the image centers by inspection, which required to re-estimate the GT homography matrices using the above protocol. Furthermore, for five image pairs that it was found to contain out-of-the-plane areas, binary masks were computed by hand using an image manipulation software. In this way it was possible to bound the true matches on the planar surfaces of the scene to better deal with semi-dense matching methods. Examples of estimated GT homography quality are shown in Fig.~\ref{gt_hom} as sampled grid point correspondences; visual results for all the image pairs in the dataset can be found \href{https://unipa-my.sharepoint.com/:f:/g/personal/fabio_bellavia_unipa_it/EjIO3vhNcRNFslDbJIiW5_oBmyrur9kdqGpKJ3WymWQ-RQ?e=UXOTr9}{here}.

\subsubsection{Non-planar dataset} 
\begin{figure}[t!]
	\center
	\includegraphics[width=0.43\textwidth]{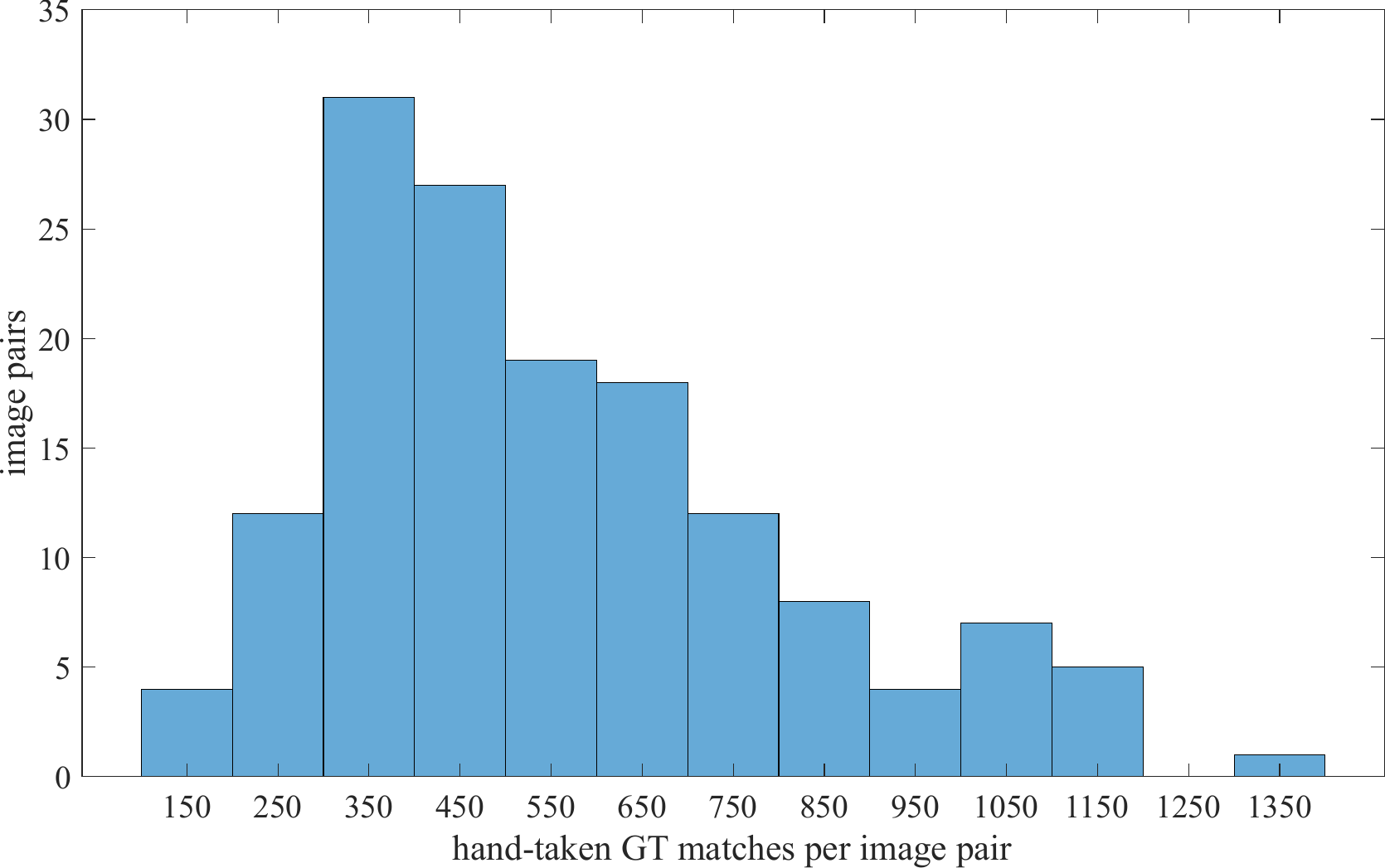}
	\caption{\label{gt_stats}
		Histogram of the number of GT hand-taken correspondences per image pair in the non-planar dataset (see SM\ref{sm_gt}). Best viewed zoomed in.}
\end{figure}

\begin{figure}[t]
	\center
	\includegraphics[width=0.41\textwidth]{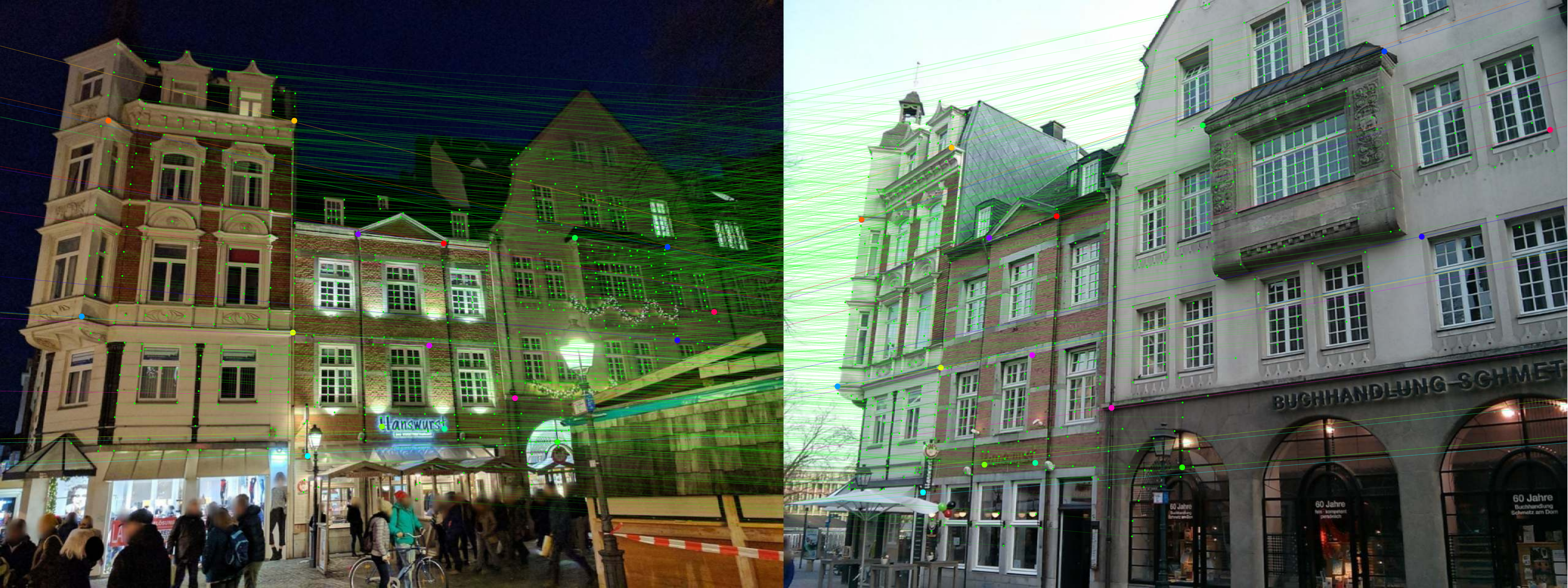} \\
	\vspace{0.3em}
	\includegraphics[width=0.41\textwidth]{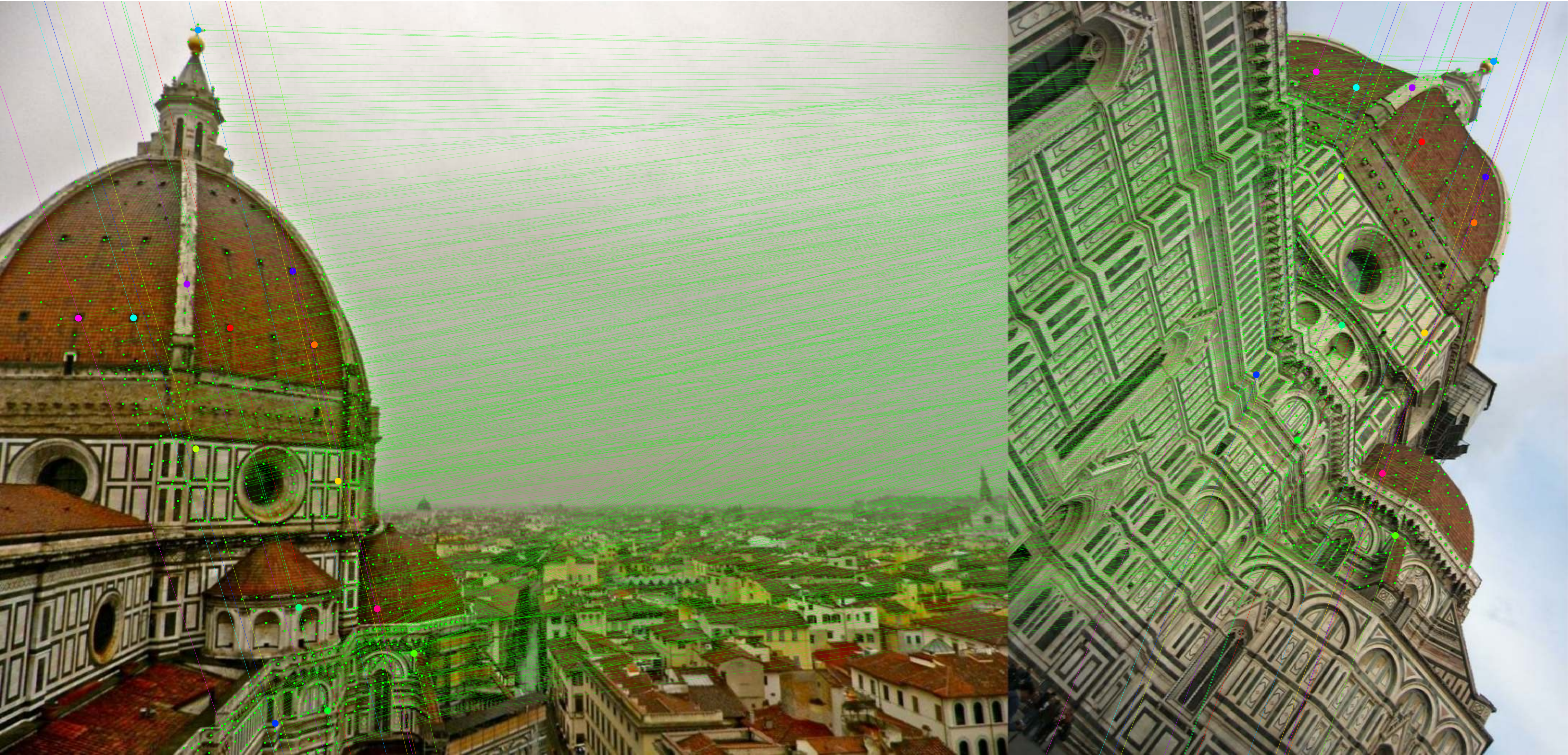} \\
	\vspace{0.3em}
	\includegraphics[width=0.41\textwidth]{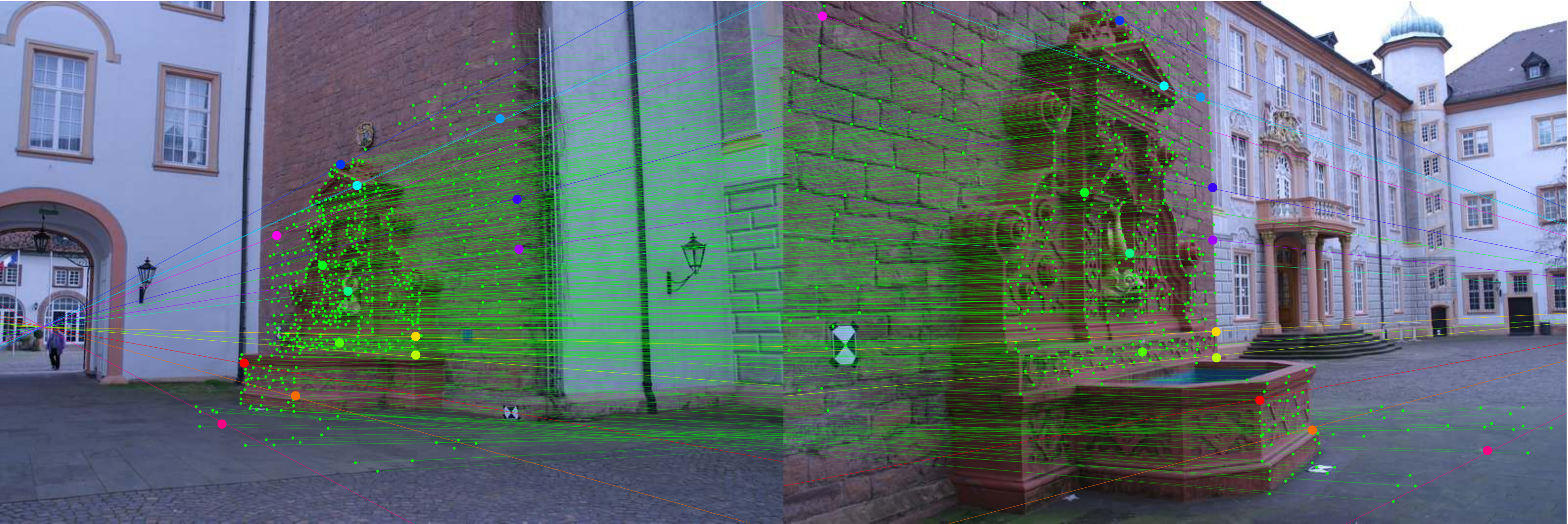}
	\caption{\label{gt_fun}
		Visual qualitative example of acquired GT correspondences, colored in green, together with corresponding epipolar lines for a sample of them, indicated by pairs of lines and big dots of other colors (see SM\ref{sm_gt}). Best viewed in color and zoomed in.}
\end{figure}

Figure~\ref{gt_stats} shows the histogram of the number of hand-taken correspondences per image pair in the case of the non-planar dataset. Corresponding minimum, mean, maximum and standard deviation values are 96, 561, 1405 and 264, respectively. Qualitative examples of the hand-taken matches together with epipolar correspondences for some of these matches are illustrated in Fig.\ref{gt_fun}; the results for all the image pairs of the dataset can be found \href{https://unipa-my.sharepoint.com/:f:/g/personal/fabio_bellavia_unipa_it/EkXt46YIrpVKm2cG87GZPVMBEsRsxRq94kn3M5WgUuM_cQ?e=pDAURN}{here}. By inspection, one can note that GT matches are uniformly distributed over the whole non-occluded image areas and the epipolar estimation is valid.

\subsection{Standard pose-based benchmarks}\label{sm_standard}
In order to estimate the pose to be decomposed into the rotation matrix and the normalized translation vector (the camera pose can only be recovered up to scale~\cite{multiview}), the standard protocol of~\cite{loftr} runs the OpenCV function \texttt{findEssentialMat}, setting an inner RANSAC with a threshold of $0.5$ px.

As reported in~\cite{imw2020}, the RANSAC optimal setup varies with the matching method, so different thresholds have been tested in the comparison. Furthermore, the post-processing RANSAC implemented as discussed above is employed to filter matches before passing them to the function \texttt{findEssentialMat}, so that overall a two-step RANSAC is executed, within the spirit of LO-RANSAC.

ScanNet images are resize to $640\times480$ px, while MegaDepth images are resized proportionally only when the maximum side exceeds 1200 px to fit this constraint.

\subsection{Detailed results on the planar and non-planar datasets}\label{sm_res1}
All the compared methods have been evaluated with or without a final RANSAC post-processing. The tested RANSAC thresholds are $t_R=5,3,2,1$ px. Table~\ref{all} reports the obtained error values without RANSAC and for $t_R=3,2$ px. These latter threshold values are the best choices in order to maximize the accuracy, considered as the most relevant error metric. The visual results for all the tested methods are available \href{https://unipa-my.sharepoint.com/:f:/g/personal/fabio_bellavia_unipa_it/Ej_efwdihjRCo8w1PeU6CbUBc6qSF0W0gEZlPvlkw6V1TQ?e=vViTF7}{here} for a qualitative inspection. 

For the planar dataset RANSAC decreases the coverage of about 5\% with respect to the raw matches. This is reasonable since RANSAC can remove poorly-localized matches which are still acceptable for the definition of coverage, but also good matches when the homography estimation is contaminated by noise. The coverage is less affected by RANSAC in the non-planar case, suggesting an increased complexity of the scenes which reduces the good matches which are fed to RANSAC.

Concerning the precision, RootSIFT and DISK seem to rely more on RANSAC than other methods to filter matches on both datasets, with a variation of about 20\% of coverage with or without post-processing.

Moreover, SuperGlue, DKM, ECO-TR, Key.Net, slimed or not, and Slimed Hz$^+$ are in order the methods that obtain the best accuracy without RANSAC. 

Finally, one can observe that failures increase with RANSAC post-processing for upright methods, which include end-to-end architectures. This mainly happens due to the presence of non-upright image pairs in the datasets for which RANSAC removes good matches found accidentally, as the whole solution is not sufficiently robust and stable.
\subsection{Detailed results on MegaDepth and ScanNet}\label{sm_res2}
Tested threshold values are $t_R=5,3,2,1$ px for the RANSAC post-processing which computes a fundamental matrix $F$, and $t_{R'}=3,2,1,0.5$ px for the successive pose estimation internal RANSAC which computes an essential matrix $E$. Also the case where there is no initial RANSAC post-processing has been tested. Table~\ref{megadepth_scannet_other_big} reports the best setup for each dataset in terms of the pose error AUC. AUC values for the individual components of the pose error are also included in the table.

The predominant RANSAC post-processing threshold values are $t_R=3,2$ px, respectively for MegaDepth and ScanNet. This indicates a better keypoint localization for MegaDepth than ScanNet, reasonably related to a higher blur in the latter case. The additional complexity of ScanNet with respect to MegaDepth is confirmed by the lower AUC values for the former dataset.

Furthermore, the pose error is dominated by the angular component related to the translation vector since the corresponding values are almost identical; AUC values are always higher for the rotation error than the translation error. Basically, the pose error is only the angular error on the translation vector. Nevertheless, all the three AUCs correlates well each other. 

\begin{table*}[t!]
	\renewcommand{\arraystretch}{0}
	\setlength{\tabcolsep}{4pt}
	\centering
	\vspace{1em}
	\caption{Detailed comparative results on the planar and non-planar datasets (see SM\ref{sm_res1}). Best viewed in color and zoomed in.}\label{all}
	\resizebox{\textwidth}{!}{
		\begin{tabular}{r<{}c<{}l<{}l<{}l<{}l<{}l<{}l<{}l<{}l<{}l<{}l}
			\toprule
			& & \multicolumn{5}{c}{Planar} & \multicolumn{5}{c}{Non-planar}\\
			\cmidrule(lr){3-7}\cmidrule(lr){8-12}
			& \shortstack[c]{RANSAC\\thr. (px)} & \shortstack[c]{Coverage\\(\%)} & \shortstack[c]{Precision\textcolor{white}{y}\\(\%)} & \shortstack[c]{Accuracy\\(\%)} & \shortstack[l]{No match\textcolor{white}{y}\\pairs} & \shortstack[l]{Wrong only\\pairs} & \shortstack[c]{Coverage\\(\%)} & \shortstack[c]{Precision\textcolor{white}{y}\\(\%)} & \shortstack[c]{Accuracy\\(\%)} & \shortstack[l]{No match\textcolor{white}{y}\\pairs} & \shortstack[l]{Wrong only\\pairs} \\[5pt]
			\midrule
			& -- & \Chart{9.60}{0.077}{blue}{25}{12}\hphantom{999999} & \Chart{33.53}{0.045}{red}{25}{12}\hphantom{99999} & \Chart{0.90}{0.010}{teal}{25}{12}\hphantom{999999} & \Chart{0}{0.000}{orange}{62}{12}\hphantom{99999999.} & \Chart{10}{0.212}{purple}{62}{12}\hphantom{9999999.} & \Chart{5.83}{0.064}{blue}{25}{12}\hphantom{999999} & \Chart{31.56}{0.045}{red}{25}{12}\hphantom{99999} & \Chart{0.00}{0.000}{teal}{25}{12}\hphantom{999999} & \Chart{0}{0.000}{orange}{62}{12}\hphantom{99999999.} & \Chart{18}{0.408}{purple}{62}{12}\hphantom{9999999.} \\
			& 3 & \Chart{8.61}{0.055}{blue}{25}{12} & \Chart{63.31}{0.454}{red}{34}{12} & \Chart{48.94}{0.553}{teal}{34}{12} & \Chart{0}{0.000}{orange}{62}{12} & \Chart{41}{0.868}{purple}{25}{12} & \Chart{5.45}{0.056}{blue}{25}{12} & \Chart{51.46}{0.339}{red}{25}{12} & \Chart{14.43}{0.201}{teal}{25}{12} & \Chart{0}{0.000}{orange}{62}{12} & \Chart{36}{0.816}{purple}{34}{12} \\
			\multirow[t]{-3}[4]{*}{RootSIFT\hphantom{$\Rsh$}} & 2 & \Chart{8.19}{0.045}{blue}{25}{12} & \Chart{64.30}{0.468}{red}{34}{12} & \Chart{50.40}{0.569}{teal}{34}{12} & \Chart{0}{0.000}{orange}{62}{12} & \Chart{41}{0.868}{purple}{25}{12} & \Chart{4.99}{0.045}{blue}{25}{12} & \Chart{53.02}{0.362}{red}{25}{12} & \Chart{17.12}{0.239}{teal}{25}{12} & \Chart{0}{0.000}{orange}{62}{12} & \Chart{42}{0.952}{purple}{25}{12} \\
			\rowcolor{gray!15}& -- & \Chart{9.96}{0.085}{blue}{25}{12} & \Chart{39.81}{0.132}{red}{25}{12} & \Chart{0.71}{0.008}{teal}{25}{12} & \Chart{0}{0.000}{orange}{62}{12} & \Chart{30}{0.635}{purple}{44}{12} & \Chart{6.09}{0.070}{blue}{25}{12} & \Chart{37.19}{0.128}{red}{25}{12} & \Chart{0.00}{0.000}{teal}{25}{12} & \Chart{0}{0.000}{orange}{62}{12} & \Chart{20}{0.454}{purple}{53}{12} \\
			\rowcolor{gray!15}& 3 & \Chart{9.03}{0.064}{blue}{25}{12} & \Chart{62.21}{0.439}{red}{34}{12} & \Chart{49.93}{0.564}{teal}{34}{12} & \Chart{3}{0.068}{orange}{62}{12} & \Chart{45}{0.952}{purple}{25}{12} & \Chart{5.81}{0.064}{blue}{25}{12} & \Chart{53.42}{0.368}{red}{25}{12} & \Chart{17.31}{0.241}{teal}{25}{12} & \Chart{1}{0.034}{orange}{62}{12} & \Chart{41}{0.930}{purple}{25}{12} \\
			\rowcolor{gray!15}\multirow[t]{-3}[4]{*}{RootSIFT$^\Rsh$} & 2 & \Chart{8.54}{0.053}{blue}{25}{12} & \Chart{60.87}{0.420}{red}{25}{12} & \Chart{47.80}{0.540}{teal}{34}{12} & \Chart{2}{0.045}{orange}{62}{12} & \Chart{45}{0.952}{purple}{25}{12} & \Chart{5.56}{0.058}{blue}{25}{12} & \Chart{55.21}{0.394}{red}{25}{12} & \Chart{24.11}{0.336}{teal}{25}{12} & \Chart{0}{0.000}{orange}{62}{12} & \Chart{41}{0.930}{purple}{25}{12} \\
			& -- & \Chart{22.91}{0.377}{blue}{25}{12} & \Chart{76.85}{0.640}{red}{44}{12} & \Chart{59.53}{0.672}{teal}{44}{12} & \Chart{13}{0.295}{orange}{62}{12} & \Chart{1}{0.021}{purple}{62}{12} & \Chart{13.84}{0.247}{blue}{25}{12} & \Chart{62.21}{0.498}{red}{34}{12} & \Chart{14.02}{0.196}{teal}{25}{12} & \Chart{22}{0.748}{orange}{34}{12} & \Chart{12}{0.272}{purple}{62}{12} \\
			& 3 & \Chart{20.17}{0.315}{blue}{25}{12} & \Chart{89.91}{0.819}{red}{53}{12} & \Chart{70.35}{0.795}{teal}{53}{12} & \Chart{14}{0.317}{orange}{62}{12} & \Chart{0}{0.000}{purple}{62}{12} & \Chart{13.45}{0.238}{blue}{25}{12} & \Chart{68.45}{0.590}{red}{34}{12} & \Chart{27.08}{0.378}{teal}{25}{12} & \Chart{27}{0.918}{orange}{25}{12} & \Chart{8}{0.181}{purple}{62}{12} \\
			\multirow[t]{-3}[4]{*}{Slimed RootSIFT\hphantom{$\Rsh$}} & 2 & \Chart{18.60}{0.280}{blue}{25}{12} & \Chart{89.94}{0.819}{red}{53}{12} & \Chart{69.36}{0.783}{teal}{53}{12} & \Chart{14}{0.317}{orange}{62}{12} & \Chart{0}{0.000}{purple}{62}{12} & \Chart{12.74}{0.221}{blue}{25}{12} & \Chart{70.60}{0.621}{red}{44}{12} & \Chart{33.47}{0.467}{teal}{34}{12} & \Chart{27}{0.918}{orange}{25}{12} & \Chart{7}{0.159}{purple}{62}{12} \\
			\rowcolor{gray!15}& -- & \Chart{17.23}{0.249}{blue}{25}{12} & \Chart{59.40}{0.400}{red}{25}{12} & \Chart{41.18}{0.465}{teal}{34}{12} & \Chart{38}{0.862}{orange}{25}{12} & \Chart{4}{0.085}{purple}{62}{12} & \Chart{11.71}{0.198}{blue}{25}{12} & \Chart{58.58}{0.444}{red}{34}{12} & \Chart{13.56}{0.189}{teal}{25}{12} & \Chart{22}{0.748}{orange}{34}{12} & \Chart{14}{0.317}{purple}{62}{12} \\
			\rowcolor{gray!15}& 3 & \Chart{15.19}{0.203}{blue}{25}{12} & \Chart{68.59}{0.526}{red}{34}{12} & \Chart{54.37}{0.614}{teal}{44}{12} & \Chart{42}{0.952}{orange}{25}{12} & \Chart{1}{0.021}{purple}{62}{12} & \Chart{11.38}{0.191}{blue}{25}{12} & \Chart{63.09}{0.510}{red}{34}{12} & \Chart{24.02}{0.335}{teal}{25}{12} & \Chart{28}{0.952}{orange}{25}{12} & \Chart{12}{0.272}{purple}{62}{12} \\
			\rowcolor{gray!15}\multirow[t]{-3}[4]{*}{Slimed RootSIFT$^\Rsh$} & 2 & \Chart{14.06}{0.178}{blue}{25}{12} & \Chart{68.84}{0.530}{red}{34}{12} & \Chart{52.96}{0.598}{teal}{34}{12} & \Chart{42}{0.952}{orange}{25}{12} & \Chart{1}{0.021}{purple}{62}{12} & \Chart{10.62}{0.173}{blue}{25}{12} & \Chart{65.63}{0.548}{red}{34}{12} & \Chart{27.31}{0.381}{teal}{25}{12} & \Chart{28}{0.952}{orange}{25}{12} & \Chart{12}{0.272}{purple}{62}{12} \\
			& -- & \Chart{35.22}{0.654}{blue}{44}{12} & \Chart{95.24}{0.892}{red}{62}{12} & \Chart{79.10}{0.893}{teal}{62}{12} & \Chart{1}{0.023}{orange}{62}{12} & \Chart{1}{0.021}{purple}{62}{12} & \Chart{21.58}{0.422}{blue}{25}{12} & \Chart{86.93}{0.862}{red}{62}{12} & \Chart{38.40}{0.536}{teal}{34}{12} & \Chart{7}{0.238}{orange}{62}{12} & \Chart{0}{0.000}{purple}{62}{12} \\
			& 3 & \Chart{31.14}{0.562}{blue}{34}{12} & \Chart{98.57}{0.938}{red}{62}{12} & \Chart{84.35}{0.953}{teal}{62}{12} & \Chart{2}{0.045}{orange}{62}{12} & \Chart{0}{0.000}{purple}{62}{12} & \Chart{21.43}{0.419}{blue}{25}{12} & \Chart{88.42}{0.884}{red}{62}{12} & \Chart{48.95}{0.683}{teal}{44}{12} & \Chart{7}{0.238}{orange}{62}{12} & \Chart{2}{0.045}{purple}{62}{12} \\
			\multirow[t]{-3}[4]{*}{Key.Net\hphantom{$\Rsh$}} & 2 & \Chart{28.04}{0.493}{blue}{34}{12} & \Chart{98.58}{0.938}{red}{62}{12} & \Chart{82.84}{0.936}{teal}{62}{12} & \Chart{2}{0.045}{orange}{62}{12} & \Chart{0}{0.000}{purple}{62}{12} & \Chart{20.68}{0.402}{blue}{25}{12} & \Chart{89.10}{0.894}{red}{62}{12} & \Chart{51.87}{0.724}{teal}{44}{12} & \Chart{8}{0.272}{orange}{62}{12} & \Chart{2}{0.045}{purple}{62}{12} \\
			\rowcolor{gray!15}& -- & \Chart{32.63}{0.596}{blue}{34}{12} & \Chart{86.49}{0.772}{red}{53}{12} & \Chart{71.58}{0.809}{teal}{53}{12} & \Chart{6}{0.136}{orange}{62}{12} & \Chart{2}{0.042}{purple}{62}{12} & \Chart{22.19}{0.436}{blue}{34}{12} & \Chart{87.05}{0.864}{red}{62}{12} & \Chart{38.45}{0.536}{teal}{34}{12} & \Chart{7}{0.238}{orange}{62}{12} & \Chart{0}{0.000}{purple}{62}{12} \\
			\rowcolor{gray!15}& 3 & \Chart{28.99}{0.514}{blue}{34}{12} & \Chart{90.03}{0.821}{red}{53}{12} & \Chart{77.07}{0.871}{teal}{62}{12} & \Chart{6}{0.136}{orange}{62}{12} & \Chart{8}{0.169}{purple}{62}{12} & \Chart{22.01}{0.432}{blue}{34}{12} & \Chart{87.84}{0.876}{red}{62}{12} & \Chart{46.67}{0.651}{teal}{44}{12} & \Chart{8}{0.272}{orange}{62}{12} & \Chart{3}{0.068}{purple}{62}{12} \\
			\rowcolor{gray!15}\multirow[t]{-3}[4]{*}{Key.Net$^\Rsh$} & 2 & \Chart{26.34}{0.454}{blue}{34}{12} & \Chart{90.22}{0.823}{red}{53}{12} & \Chart{77.07}{0.871}{teal}{62}{12} & \Chart{6}{0.136}{orange}{62}{12} & \Chart{7}{0.148}{purple}{62}{12} & \Chart{21.29}{0.416}{blue}{25}{12} & \Chart{88.48}{0.885}{red}{62}{12} & \Chart{50.46}{0.704}{teal}{44}{12} & \Chart{8}{0.272}{orange}{62}{12} & \Chart{2}{0.045}{purple}{62}{12} \\
			& -- & \Chart{35.48}{0.660}{blue}{44}{12} & \Chart{92.97}{0.861}{red}{62}{12} & \Chart{73.43}{0.829}{teal}{53}{12} & \Chart{2}{0.045}{orange}{62}{12} & \Chart{1}{0.021}{purple}{62}{12} & \Chart{23.70}{0.470}{blue}{34}{12} & \Chart{83.87}{0.817}{red}{53}{12} & \Chart{36.62}{0.511}{teal}{34}{12} & \Chart{0}{0.000}{orange}{62}{12} & \Chart{0}{0.000}{purple}{62}{12} \\
			& 3 & \Chart{31.32}{0.567}{blue}{34}{12} & \Chart{96.35}{0.907}{red}{62}{12} & \Chart{79.67}{0.900}{teal}{62}{12} & \Chart{3}{0.068}{orange}{62}{12} & \Chart{2}{0.042}{purple}{62}{12} & \Chart{23.19}{0.459}{blue}{34}{12} & \Chart{88.63}{0.887}{red}{62}{12} & \Chart{43.11}{0.601}{teal}{34}{12} & \Chart{0}{0.000}{orange}{62}{12} & \Chart{0}{0.000}{purple}{62}{12} \\
			\multirow[t]{-3}[4]{*}{Slimed Key.Net\hphantom{$\Rsh$}} & 2 & \Chart{29.13}{0.517}{blue}{34}{12} & \Chart{96.32}{0.907}{red}{62}{12} & \Chart{79.67}{0.900}{teal}{62}{12} & \Chart{2}{0.045}{orange}{62}{12} & \Chart{3}{0.063}{purple}{62}{12} & \Chart{21.90}{0.429}{blue}{34}{12} & \Chart{90.41}{0.914}{red}{62}{12} & \Chart{41.46}{0.578}{teal}{34}{12} & \Chart{0}{0.000}{orange}{62}{12} & \Chart{0}{0.000}{purple}{62}{12} \\
			\rowcolor{gray!15}& -- & \Chart{32.49}{0.593}{blue}{34}{12} & \Chart{87.07}{0.780}{red}{53}{12} & \Chart{63.83}{0.721}{teal}{44}{12} & \Chart{4}{0.091}{orange}{62}{12} & \Chart{5}{0.106}{purple}{62}{12} & \Chart{23.41}{0.464}{blue}{34}{12} & \Chart{79.77}{0.757}{red}{53}{12} & \Chart{31.19}{0.435}{teal}{34}{12} & \Chart{0}{0.000}{orange}{62}{12} & \Chart{6}{0.136}{purple}{62}{12} \\
			\rowcolor{gray!15}& 3 & \Chart{28.65}{0.506}{blue}{34}{12} & \Chart{91.41}{0.839}{red}{53}{12} & \Chart{74.04}{0.836}{teal}{53}{12} & \Chart{5}{0.113}{orange}{62}{12} & \Chart{6}{0.127}{purple}{62}{12} & \Chart{22.91}{0.452}{blue}{34}{12} & \Chart{84.61}{0.828}{red}{53}{12} & \Chart{43.74}{0.610}{teal}{44}{12} & \Chart{0}{0.000}{orange}{62}{12} & \Chart{8}{0.181}{purple}{62}{12} \\
			\rowcolor{gray!15}\multirow[t]{-3}[4]{*}{Slimed Key.Net$^\Rsh$} & 2 & \Chart{26.45}{0.457}{blue}{34}{12} & \Chart{90.63}{0.829}{red}{53}{12} & \Chart{73.19}{0.827}{teal}{53}{12} & \Chart{6}{0.136}{orange}{62}{12} & \Chart{6}{0.127}{purple}{62}{12} & \Chart{21.54}{0.421}{blue}{25}{12} & \Chart{85.87}{0.847}{red}{53}{12} & \Chart{40.41}{0.564}{teal}{34}{12} & \Chart{0}{0.000}{orange}{62}{12} & \Chart{8}{0.181}{purple}{62}{12} \\
			& -- & \Chart{30.00}{0.537}{blue}{34}{12} & \Chart{90.37}{0.825}{red}{53}{12} & \Chart{22.65}{0.256}{teal}{25}{12} & \Chart{0}{0.000}{orange}{62}{12} & \Chart{0}{0.000}{purple}{62}{12} & \Chart{20.60}{0.400}{blue}{25}{12} & \Chart{76.61}{0.710}{red}{44}{12} & \Chart{1.92}{0.027}{teal}{25}{12} & \Chart{0}{0.000}{orange}{62}{12} & \Chart{3}{0.068}{purple}{62}{12} \\
			& 3 & \Chart{28.10}{0.494}{blue}{34}{12} & \Chart{98.67}{0.939}{red}{62}{12} & \Chart{83.83}{0.947}{teal}{62}{12} & \Chart{0}{0.000}{orange}{62}{12} & \Chart{1}{0.021}{purple}{62}{12} & \Chart{20.34}{0.394}{blue}{25}{12} & \Chart{85.79}{0.845}{red}{53}{12} & \Chart{49.32}{0.688}{teal}{44}{12} & \Chart{0}{0.000}{orange}{62}{12} & \Chart{5}{0.113}{purple}{62}{12} \\
			\multirow[t]{-3}[4]{*}{Hz$^+$\hphantom{$\Rsh$}} & 2 & \Chart{26.33}{0.454}{blue}{34}{12} & \Chart{98.99}{0.943}{red}{62}{12} & \Chart{83.07}{0.938}{teal}{62}{12} & \Chart{0}{0.000}{orange}{62}{12} & \Chart{0}{0.000}{purple}{62}{12} & \Chart{19.74}{0.380}{blue}{25}{12} & \Chart{87.58}{0.872}{red}{62}{12} & \Chart{53.01}{0.740}{teal}{53}{12} & \Chart{0}{0.000}{orange}{62}{12} & \Chart{6}{0.136}{purple}{62}{12} \\
			\rowcolor{gray!15}& -- & \Chart{28.69}{0.507}{blue}{34}{12} & \Chart{85.72}{0.761}{red}{53}{12} & \Chart{18.96}{0.214}{teal}{25}{12} & \Chart{0}{0.000}{orange}{62}{12} & \Chart{4}{0.085}{purple}{62}{12} & \Chart{21.22}{0.414}{blue}{25}{12} & \Chart{77.13}{0.718}{red}{44}{12} & \Chart{2.60}{0.036}{teal}{25}{12} & \Chart{0}{0.000}{orange}{62}{12} & \Chart{3}{0.068}{purple}{62}{12} \\
			\rowcolor{gray!15}& 3 & \Chart{26.79}{0.464}{blue}{34}{12} & \Chart{95.34}{0.893}{red}{62}{12} & \Chart{81.42}{0.920}{teal}{62}{12} & \Chart{0}{0.000}{orange}{62}{12} & \Chart{5}{0.106}{purple}{62}{12} & \Chart{20.90}{0.407}{blue}{25}{12} & \Chart{87.37}{0.869}{red}{62}{12} & \Chart{48.45}{0.676}{teal}{44}{12} & \Chart{0}{0.000}{orange}{62}{12} & \Chart{5}{0.113}{purple}{62}{12} \\
			\rowcolor{gray!15}\multirow[t]{-3}[4]{*}{Hz$^{+\Rsh}$} & 2 & \Chart{25.00}{0.424}{blue}{25}{12} & \Chart{95.35}{0.893}{red}{62}{12} & \Chart{79.29}{0.896}{teal}{62}{12} & \Chart{0}{0.000}{orange}{62}{12} & \Chart{5}{0.106}{purple}{62}{12} & \Chart{20.24}{0.392}{blue}{25}{12} & \Chart{88.17}{0.880}{red}{62}{12} & \Chart{52.88}{0.738}{teal}{44}{12} & \Chart{0}{0.000}{orange}{62}{12} & \Chart{4}{0.091}{purple}{62}{12} \\
			& -- & \Chart{38.72}{0.733}{blue}{44}{12} & \Chart{94.10}{0.876}{red}{62}{12} & \Chart{79.39}{0.897}{teal}{62}{12} & \Chart{0}{0.000}{orange}{62}{12} & \Chart{0}{0.000}{purple}{62}{12} & \Chart{31.83}{0.655}{blue}{44}{12} & \Chart{85.22}{0.837}{red}{53}{12} & \Chart{35.66}{0.498}{teal}{34}{12} & \Chart{0}{0.000}{orange}{62}{12} & \Chart{1}{0.023}{purple}{62}{12} \\
			& 3 & \Chart{34.87}{0.647}{blue}{44}{12} & \Chart{99.49}{0.950}{red}{62}{12} & \Chart{82.74}{0.935}{teal}{62}{12} & \Chart{0}{0.000}{orange}{62}{12} & \Chart{0}{0.000}{purple}{62}{12} & \Chart{31.14}{0.639}{blue}{44}{12} & \Chart{91.69}{0.932}{red}{62}{12} & \Chart{53.20}{0.742}{teal}{53}{12} & \Chart{0}{0.000}{orange}{62}{12} & \Chart{1}{0.023}{purple}{62}{12} \\
			\multirow[t]{-3}[4]{*}{Slimed Hz$^+$\hphantom{$\Rsh$}} & 2 & \Chart{33.59}{0.618}{blue}{44}{12} & \Chart{99.80}{0.955}{red}{62}{12} & \Chart{83.50}{0.943}{teal}{62}{12} & \Chart{0}{0.000}{orange}{62}{12} & \Chart{0}{0.000}{purple}{62}{12} & \Chart{29.70}{0.607}{blue}{44}{12} & \Chart{93.19}{0.955}{red}{62}{12} & \Chart{49.27}{0.687}{teal}{44}{12} & \Chart{0}{0.000}{orange}{62}{12} & \Chart{1}{0.023}{purple}{62}{12} \\
			\rowcolor{gray!15}& -- & \Chart{35.52}{0.661}{blue}{44}{12} & \Chart{87.21}{0.782}{red}{53}{12} & \Chart{69.74}{0.788}{teal}{53}{12} & \Chart{5}{0.113}{orange}{62}{12} & \Chart{3}{0.063}{purple}{62}{12} & \Chart{30.43}{0.623}{blue}{44}{12} & \Chart{82.26}{0.793}{red}{53}{12} & \Chart{34.75}{0.485}{teal}{34}{12} & \Chart{0}{0.000}{orange}{62}{12} & \Chart{4}{0.091}{purple}{62}{12} \\
			\rowcolor{gray!15}& 3 & \Chart{32.23}{0.587}{blue}{34}{12} & \Chart{93.18}{0.864}{red}{62}{12} & \Chart{77.40}{0.874}{teal}{62}{12} & \Chart{7}{0.159}{orange}{62}{12} & \Chart{2}{0.042}{purple}{62}{12} & \Chart{29.79}{0.609}{blue}{44}{12} & \Chart{89.39}{0.898}{red}{62}{12} & \Chart{53.52}{0.747}{teal}{53}{12} & \Chart{0}{0.000}{orange}{62}{12} & \Chart{4}{0.091}{purple}{62}{12} \\
			\rowcolor{gray!15}\multirow[t]{-3}[4]{*}{Slimed Hz$^{+\Rsh}$} & 2 & \Chart{30.60}{0.550}{blue}{34}{12} & \Chart{93.32}{0.866}{red}{62}{12} & \Chart{77.16}{0.872}{teal}{62}{12} & \Chart{7}{0.159}{orange}{62}{12} & \Chart{2}{0.042}{purple}{62}{12} & \Chart{28.27}{0.574}{blue}{34}{12} & \Chart{90.86}{0.920}{red}{62}{12} & \Chart{47.26}{0.659}{teal}{44}{12} & \Chart{0}{0.000}{orange}{62}{12} & \Chart{6}{0.136}{purple}{62}{12} \\
			& -- & \Chart{26.74}{0.463}{blue}{34}{12} & \Chart{64.34}{0.468}{red}{34}{12} & \Chart{7.52}{0.085}{teal}{25}{12} & \Chart{0}{0.000}{orange}{62}{12} & \Chart{11}{0.233}{purple}{62}{12} & \Chart{24.35}{0.485}{blue}{34}{12} & \Chart{62.41}{0.500}{red}{34}{12} & \Chart{2.83}{0.039}{teal}{25}{12} & \Chart{0}{0.000}{orange}{62}{12} & \Chart{1}{0.023}{purple}{62}{12} \\
			& 3 & \Chart{24.21}{0.406}{blue}{25}{12} & \Chart{80.43}{0.689}{red}{44}{12} & \Chart{65.72}{0.742}{teal}{53}{12} & \Chart{0}{0.000}{orange}{62}{12} & \Chart{24}{0.508}{purple}{53}{12} & \Chart{23.51}{0.466}{blue}{34}{12} & \Chart{78.91}{0.744}{red}{53}{12} & \Chart{40.68}{0.568}{teal}{34}{12} & \Chart{0}{0.000}{orange}{62}{12} & \Chart{14}{0.317}{purple}{62}{12} \\
			\multirow[t]{-3}[4]{*}{DISK\hphantom{$\Rsh$}} & 2 & \Chart{22.57}{0.369}{blue}{25}{12} & \Chart{80.52}{0.690}{red}{44}{12} & \Chart{63.97}{0.723}{teal}{44}{12} & \Chart{0}{0.000}{orange}{62}{12} & \Chart{23}{0.487}{purple}{53}{12} & \Chart{22.47}{0.442}{blue}{34}{12} & \Chart{80.23}{0.763}{red}{53}{12} & \Chart{44.57}{0.622}{teal}{44}{12} & \Chart{0}{0.000}{orange}{62}{12} & \Chart{13}{0.295}{purple}{62}{12} \\
			\rowcolor{gray!15}& -- & \Chart{27.22}{0.474}{blue}{34}{12} & \Chart{89.17}{0.809}{red}{53}{12} & \Chart{78.58}{0.888}{teal}{62}{12} & \Chart{0}{0.000}{orange}{62}{12} & \Chart{11}{0.233}{purple}{62}{12} & \Chart{24.00}{0.477}{blue}{34}{12} & \Chart{85.98}{0.848}{red}{53}{12} & \Chart{50.68}{0.707}{teal}{44}{12} & \Chart{0}{0.000}{orange}{62}{12} & \Chart{4}{0.091}{purple}{62}{12} \\
			\rowcolor{gray!15}& 3 & \Chart{23.65}{0.394}{blue}{25}{12} & \Chart{90.86}{0.832}{red}{53}{12} & \Chart{79.86}{0.902}{teal}{62}{12} & \Chart{3}{0.068}{orange}{62}{12} & \Chart{9}{0.190}{purple}{62}{12} & \Chart{23.81}{0.473}{blue}{34}{12} & \Chart{88.80}{0.890}{red}{62}{12} & \Chart{59.59}{0.831}{teal}{53}{12} & \Chart{0}{0.000}{orange}{62}{12} & \Chart{6}{0.136}{purple}{62}{12} \\
			\rowcolor{gray!15}\multirow[t]{-3}[4]{*}{SuperGlue\hphantom{$\Rsh$}} & 2 & \Chart{20.18}{0.316}{blue}{25}{12} & \Chart{90.69}{0.830}{red}{53}{12} & \Chart{78.91}{0.891}{teal}{62}{12} & \Chart{3}{0.068}{orange}{62}{12} & \Chart{10}{0.212}{purple}{62}{12} & \Chart{23.14}{0.458}{blue}{34}{12} & \Chart{90.10}{0.909}{red}{62}{12} & \Chart{60.41}{0.843}{teal}{53}{12} & \Chart{0}{0.000}{orange}{62}{12} & \Chart{4}{0.091}{purple}{62}{12} \\
			& -- & \Chart{38.22}{0.722}{blue}{44}{12} & \Chart{77.75}{0.652}{red}{44}{12} & \Chart{39.10}{0.442}{teal}{34}{12} & \Chart{0}{0.000}{orange}{62}{12} & \Chart{8}{0.169}{purple}{62}{12} & \Chart{34.60}{0.718}{blue}{44}{12} & \Chart{78.61}{0.740}{red}{44}{12} & \Chart{17.72}{0.247}{teal}{25}{12} & \Chart{0}{0.000}{orange}{62}{12} & \Chart{3}{0.068}{purple}{62}{12} \\
			& 3 & \Chart{31.58}{0.572}{blue}{34}{12} & \Chart{85.63}{0.760}{red}{53}{12} & \Chart{65.63}{0.741}{teal}{53}{12} & \Chart{0}{0.000}{orange}{62}{12} & \Chart{16}{0.339}{purple}{62}{12} & \Chart{34.17}{0.708}{blue}{44}{12} & \Chart{86.32}{0.853}{red}{53}{12} & \Chart{53.24}{0.743}{teal}{53}{12} & \Chart{0}{0.000}{orange}{62}{12} & \Chart{6}{0.136}{purple}{62}{12} \\
			\multirow[t]{-3}[4]{*}{LoFTR\hphantom{$\Rsh$}} & 2 & \Chart{26.70}{0.462}{blue}{34}{12} & \Chart{85.06}{0.752}{red}{53}{12} & \Chart{65.39}{0.739}{teal}{53}{12} & \Chart{0}{0.000}{orange}{62}{12} & \Chart{17}{0.360}{purple}{62}{12} & \Chart{32.97}{0.681}{blue}{44}{12} & \Chart{88.22}{0.881}{red}{62}{12} & \Chart{55.84}{0.779}{teal}{53}{12} & \Chart{0}{0.000}{orange}{62}{12} & \Chart{6}{0.136}{purple}{62}{12} \\
			\rowcolor{gray!15}& -- & \Chart{38.52}{0.729}{blue}{44}{12} & \Chart{84.26}{0.741}{red}{53}{12} & \Chart{39.67}{0.448}{teal}{34}{12} & \Chart{0}{0.000}{orange}{62}{12} & \Chart{3}{0.063}{purple}{62}{12} & \Chart{32.89}{0.679}{blue}{44}{12} & \Chart{76.41}{0.707}{red}{44}{12} & \Chart{14.52}{0.203}{teal}{25}{12} & \Chart{0}{0.000}{orange}{62}{12} & \Chart{1}{0.023}{purple}{62}{12} \\
			\rowcolor{gray!15}& 3 & \Chart{34.97}{0.649}{blue}{44}{12} & \Chart{93.04}{0.862}{red}{62}{12} & \Chart{77.97}{0.881}{teal}{62}{12} & \Chart{0}{0.000}{orange}{62}{12} & \Chart{8}{0.169}{purple}{62}{12} & \Chart{32.45}{0.669}{blue}{44}{12} & \Chart{86.96}{0.863}{red}{62}{12} & \Chart{53.88}{0.752}{teal}{53}{12} & \Chart{0}{0.000}{orange}{62}{12} & \Chart{1}{0.023}{purple}{62}{12} \\
			\rowcolor{gray!15}\multirow[t]{-3}[4]{*}{SE2-LoFTR\hphantom{$\Rsh$}} & 2 & \Chart{32.69}{0.597}{blue}{34}{12} & \Chart{93.09}{0.862}{red}{62}{12} & \Chart{76.31}{0.862}{teal}{62}{12} & \Chart{0}{0.000}{orange}{62}{12} & \Chart{7}{0.148}{purple}{62}{12} & \Chart{31.37}{0.645}{blue}{44}{12} & \Chart{88.05}{0.879}{red}{62}{12} & \Chart{58.58}{0.817}{teal}{53}{12} & \Chart{0}{0.000}{orange}{62}{12} & \Chart{2}{0.045}{purple}{62}{12} \\
			& -- & \Chart{40.22}{0.767}{blue}{53}{12} & \Chart{80.26}{0.686}{red}{44}{12} & \Chart{38.25}{0.432}{teal}{34}{12} & \Chart{0}{0.000}{orange}{62}{12} & \Chart{8}{0.169}{purple}{62}{12} & \Chart{36.36}{0.758}{blue}{53}{12} & \Chart{83.51}{0.812}{red}{53}{12} & \Chart{14.25}{0.199}{teal}{25}{12} & \Chart{0}{0.000}{orange}{62}{12} & \Chart{0}{0.000}{purple}{62}{12} \\
			& 3 & \Chart{32.83}{0.601}{blue}{34}{12} & \Chart{86.77}{0.776}{red}{53}{12} & \Chart{66.19}{0.748}{teal}{53}{12} & \Chart{0}{0.000}{orange}{62}{12} & \Chart{16}{0.339}{purple}{62}{12} & \Chart{35.98}{0.749}{blue}{53}{12} & \Chart{90.24}{0.911}{red}{62}{12} & \Chart{55.39}{0.773}{teal}{53}{12} & \Chart{0}{0.000}{orange}{62}{12} & \Chart{3}{0.068}{purple}{62}{12} \\
			\multirow[t]{-3}[4]{*}{MatchFormer\hphantom{$\Rsh$}} & 2 & \Chart{27.56}{0.482}{blue}{34}{12} & \Chart{85.83}{0.763}{red}{53}{12} & \Chart{64.82}{0.732}{teal}{44}{12} & \Chart{0}{0.000}{orange}{62}{12} & \Chart{17}{0.360}{purple}{62}{12} & \Chart{34.68}{0.720}{blue}{44}{12} & \Chart{91.86}{0.935}{red}{62}{12} & \Chart{57.72}{0.805}{teal}{53}{12} & \Chart{0}{0.000}{orange}{62}{12} & \Chart{2}{0.045}{purple}{62}{12} \\
			\rowcolor{gray!15}& -- & \Chart{39.33}{0.747}{blue}{53}{12} & \Chart{83.08}{0.725}{red}{44}{12} & \Chart{53.29}{0.602}{teal}{34}{12} & \Chart{0}{0.000}{orange}{62}{12} & \Chart{7}{0.148}{purple}{62}{12} & \Chart{37.04}{0.773}{blue}{53}{12} & \Chart{85.85}{0.846}{red}{53}{12} & \Chart{26.67}{0.372}{teal}{25}{12} & \Chart{0}{0.000}{orange}{62}{12} & \Chart{2}{0.045}{purple}{62}{12} \\
			\rowcolor{gray!15}& 3 & \Chart{36.31}{0.679}{blue}{44}{12} & \Chart{87.41}{0.785}{red}{53}{12} & \Chart{76.41}{0.863}{teal}{62}{12} & \Chart{0}{0.000}{orange}{62}{12} & \Chart{16}{0.339}{purple}{62}{12} & \Chart{36.79}{0.768}{blue}{53}{12} & \Chart{90.30}{0.912}{red}{62}{12} & \Chart{61.28}{0.855}{teal}{62}{12} & \Chart{0}{0.000}{orange}{62}{12} & \Chart{4}{0.091}{purple}{62}{12} \\
			\rowcolor{gray!15}\multirow[t]{-3}[4]{*}{QuadTree Att.\hphantom{$\Rsh$}} & 2 & \Chart{34.18}{0.631}{blue}{44}{12} & \Chart{87.69}{0.788}{red}{53}{12} & \Chart{74.99}{0.847}{teal}{53}{12} & \Chart{0}{0.000}{orange}{62}{12} & \Chart{13}{0.275}{purple}{62}{12} & \Chart{36.05}{0.751}{blue}{53}{12} & \Chart{91.87}{0.935}{red}{62}{12} & \Chart{68.26}{0.952}{teal}{62}{12} & \Chart{0}{0.000}{orange}{62}{12} & \Chart{4}{0.091}{purple}{62}{12} \\
			& -- & \Chart{39.95}{0.761}{blue}{53}{12} & \Chart{82.04}{0.711}{red}{44}{12} & \Chart{69.36}{0.783}{teal}{53}{12} & \Chart{17}{0.385}{orange}{62}{12} & \Chart{5}{0.106}{purple}{62}{12} & \Chart{32.69}{0.675}{blue}{44}{12} & \Chart{85.03}{0.834}{red}{53}{12} & \Chart{47.95}{0.669}{teal}{44}{12} & \Chart{15}{0.510}{orange}{53}{12} & \Chart{1}{0.023}{purple}{62}{12} \\
			& 3 & \Chart{36.70}{0.688}{blue}{44}{12} & \Chart{82.65}{0.719}{red}{44}{12} & \Chart{70.50}{0.796}{teal}{53}{12} & \Chart{17}{0.385}{orange}{62}{12} & \Chart{6}{0.127}{purple}{62}{12} & \Chart{32.52}{0.671}{blue}{44}{12} & \Chart{85.75}{0.845}{red}{53}{12} & \Chart{53.70}{0.749}{teal}{53}{12} & \Chart{15}{0.510}{orange}{53}{12} & \Chart{1}{0.023}{purple}{62}{12} \\
			\multirow[t]{-3}[4]{*}{ECO-TR\hphantom{$\Rsh$}} & 2 & \Chart{34.48}{0.638}{blue}{44}{12} & \Chart{82.87}{0.722}{red}{44}{12} & \Chart{68.79}{0.777}{teal}{53}{12} & \Chart{17}{0.385}{orange}{62}{12} & \Chart{5}{0.106}{purple}{62}{12} & \Chart{31.76}{0.654}{blue}{44}{12} & \Chart{85.76}{0.845}{red}{53}{12} & \Chart{54.79}{0.764}{teal}{53}{12} & \Chart{15}{0.510}{orange}{53}{12} & \Chart{1}{0.023}{purple}{62}{12} \\
			\rowcolor{gray!15}& -- & \Chart{48.54}{0.955}{blue}{62}{12} & \Chart{87.38}{0.784}{red}{53}{12} & \Chart{70.35}{0.795}{teal}{53}{12} & \Chart{0}{0.000}{orange}{62}{12} & \Chart{6}{0.127}{purple}{62}{12} & \Chart{45.01}{0.955}{blue}{62}{12} & \Chart{83.65}{0.814}{red}{53}{12} & \Chart{48.63}{0.678}{teal}{44}{12} & \Chart{0}{0.000}{orange}{62}{12} & \Chart{0}{0.000}{purple}{62}{12} \\
			\rowcolor{gray!15}& 3 & \Chart{44.66}{0.867}{blue}{62}{12} & \Chart{90.74}{0.830}{red}{53}{12} & \Chart{81.28}{0.918}{teal}{62}{12} & \Chart{0}{0.000}{orange}{62}{12} & \Chart{11}{0.233}{purple}{62}{12} & \Chart{44.71}{0.948}{blue}{62}{12} & \Chart{86.60}{0.857}{red}{62}{12} & \Chart{62.37}{0.870}{teal}{62}{12} & \Chart{0}{0.000}{orange}{62}{12} & \Chart{3}{0.068}{purple}{62}{12} \\
			\rowcolor{gray!15}\multirow[t]{-3}[4]{*}{DKM\hphantom{$\Rsh$}} & 2 & \Chart{42.13}{0.810}{blue}{53}{12} & \Chart{90.78}{0.831}{red}{53}{12} & \Chart{81.09}{0.916}{teal}{62}{12} & \Chart{0}{0.000}{orange}{62}{12} & \Chart{11}{0.233}{purple}{62}{12} & \Chart{43.81}{0.927}{blue}{62}{12} & \Chart{88.27}{0.882}{red}{62}{12} & \Chart{66.85}{0.933}{teal}{62}{12} & \Chart{0}{0.000}{orange}{62}{12} & \Chart{3}{0.068}{purple}{62}{12} \\
			\bottomrule
		\end{tabular}
	}
	\vspace{1em}
\end{table*}

\begin{table*}[t!]
	\vspace{4em}
	\renewcommand{\arraystretch}{0}
	\setlength{\tabcolsep}{4pt}
	\centering
	\caption{Detailed results on MegaDepth and ScanNet (see SM\ref{sm_res2}). Best viewed in color and zoomed in.}\label{megadepth_scannet_other_big}	
	\resizebox{0.85\textwidth}{!}{
		\begin{tabular}{c<{}r<{}c<{}c<{}c<{}c<{}c<{}c<{}c<{}c<{}c<{}l}
			\toprule
			& & \multicolumn{5}{c}{MegaDepth} & \multicolumn{5}{c}{ScanNet}\\
			\cmidrule(r{0.2em}){3-7}\cmidrule(l{0.2em}){8-12}
			& & \multicolumn{2}{c}{thr. (px)} & \multicolumn{3}{c}{AUC (\%)} & \multicolumn{2}{c}{thr. (px)} & \multicolumn{3}{c}{AUC (\%)} \\
			\cmidrule(l{0.1em}r{0.1em}){3-4}\cmidrule(lr){5-7}\cmidrule(l{0.1em}r{0.1em}){8-9}\cmidrule(lr){10-12}
			& & $F$ & $E$ & $R$ & $\mathbf{t}$ & $\max{(R,\mathbf{t})}$ & $F$ & $E$ & $R$ & $\mathbf{t}$ & $\max{(R,\mathbf{t})}$ \\[5pt]
			\midrule
			\cellcolor{white} & RootSIFT\hphantom{$\Rsh$}& 5 & 0.5 & \Chart{51.80}{0.056}{blue}{25}{12}\hphantom{9999.} & \Chart{28.31}{0.049}{red}{25}{12}\hphantom{9999.} & \Chart{27.63}{0.050}{teal}{25}{12}\hphantom{9999.}& 2 & 1.0 & \Chart{12.97}{0.045}{blue}{25}{12}\hphantom{9999.} & \Chart{3.18}{0.045}{red}{25}{12}\hphantom{99999.} & \Chart{2.75}{0.043}{teal}{25}{12}\hphantom{99999.} \\
			\rowcolor{gray!15}\cellcolor{white} & RootSIFT$^\Rsh$& 5 & 0.5 & \Chart{51.33}{0.045}{blue}{25}{12}\hphantom{9999.} & \Chart{28.08}{0.045}{red}{25}{12}\hphantom{9999.} & \Chart{27.37}{0.045}{teal}{25}{12}\hphantom{9999.}& 5 & 1.0 & \Chart{15.59}{0.081}{blue}{25}{12}\hphantom{9999.} & \Chart{3.63}{0.052}{red}{25}{12}\hphantom{99999.} & \Chart{3.18}{0.049}{teal}{25}{12}\hphantom{99999.} \\
			\cellcolor{white} & Key.Net\hphantom{$\Rsh$}& 2 & 0.5 & \Chart{69.69}{0.442}{blue}{34}{12}\hphantom{9999.} & \Chart{42.84}{0.288}{red}{25}{12}\hphantom{9999.} & \Chart{40.94}{0.267}{teal}{25}{12}\hphantom{9999.}& -- & 0.5 & \Chart{32.21}{0.303}{blue}{25}{12}\hphantom{9999.} & \Chart{10.93}{0.163}{red}{25}{12}\hphantom{9999.} & \Chart{9.69}{0.150}{teal}{25}{12}\hphantom{99999.} \\
			\rowcolor{gray!15}\cellcolor{white} & Key.Net$^\Rsh$& 2 & 0.5 & \Chart{71.72}{0.486}{blue}{34}{12}\hphantom{9999.} & \Chart{45.85}{0.338}{red}{25}{12}\hphantom{9999.} & \Chart{44.01}{0.318}{teal}{25}{12}\hphantom{9999.}& -- & 0.5 & \Chart{32.61}{0.309}{blue}{25}{12}\hphantom{9999.} & \Chart{11.07}{0.166}{red}{25}{12}\hphantom{9999.} & \Chart{9.91}{0.154}{teal}{25}{12}\hphantom{99999.} \\
			\cellcolor{white} & Hz$^+$\hphantom{$\Rsh$}& 5 & 0.5 & \Chart{73.17}{0.517}{blue}{34}{12}\hphantom{9999.} & \Chart{47.15}{0.359}{red}{25}{12}\hphantom{9999.} & \Chart{45.52}{0.342}{teal}{25}{12}\hphantom{9999.}& 2 & 0.5 & \Chart{33.01}{0.314}{blue}{25}{12}\hphantom{9999.} & \Chart{12.11}{0.181}{red}{25}{12}\hphantom{9999.} & \Chart{10.77}{0.167}{teal}{25}{12}\hphantom{9999.} \\
			\rowcolor{gray!15}\cellcolor{white} & Hz$^{+\Rsh}$& 2 & 0.5 & \Chart{74.76}{0.552}{blue}{34}{12}\hphantom{9999.} & \Chart{48.61}{0.383}{red}{25}{12}\hphantom{9999.} & \Chart{46.88}{0.365}{teal}{25}{12}\hphantom{9999.}& -- & 0.5 & \Chart{33.10}{0.315}{blue}{25}{12}\hphantom{9999.} & \Chart{11.77}{0.176}{red}{25}{12}\hphantom{9999.} & \Chart{10.58}{0.164}{teal}{25}{12}\hphantom{9999.} \\
			\cellcolor{white} & Slimed Hz$^+$\hphantom{$\Rsh$}& 2 & 0.5 & \Chart{71.11}{0.473}{blue}{34}{12}\hphantom{9999.} & \Chart{44.23}{0.311}{red}{25}{12}\hphantom{9999.} & \Chart{42.49}{0.293}{teal}{25}{12}\hphantom{9999.}& 2 & 0.5 & \Chart{40.67}{0.417}{blue}{25}{12}\hphantom{9999.} & \Chart{16.20}{0.244}{red}{25}{12}\hphantom{9999.} & \Chart{14.69}{0.228}{teal}{25}{12}\hphantom{9999.} \\
			\rowcolor{gray!15}\cellcolor{white} & Slimed Hz$^{+\Rsh}$& 2 & 0.5 & \Chart{71.69}{0.485}{blue}{34}{12}\hphantom{9999.} & \Chart{43.45}{0.298}{red}{25}{12}\hphantom{9999.} & \Chart{41.80}{0.281}{teal}{25}{12}\hphantom{9999.}& -- & 0.5 & \Chart{38.93}{0.393}{blue}{25}{12}\hphantom{9999.} & \Chart{15.53}{0.233}{red}{25}{12}\hphantom{9999.} & \Chart{14.10}{0.219}{teal}{25}{12}\hphantom{9999.} \\
			\cellcolor{white} & DISK\hphantom{$\Rsh$}& 5 & 0.5 & \Chart{69.61}{0.440}{blue}{34}{12}\hphantom{9999.} & \Chart{42.90}{0.289}{red}{25}{12}\hphantom{9999.} & \Chart{41.57}{0.278}{teal}{25}{12}\hphantom{9999.}& 3 & 0.5 & \Chart{31.51}{0.294}{blue}{25}{12}\hphantom{9999.} & \Chart{10.79}{0.161}{red}{25}{12}\hphantom{9999.} & \Chart{9.59}{0.149}{teal}{25}{12}\hphantom{99999.} \\
			\rowcolor{gray!15}\cellcolor{white} & SuperGlue\hphantom{$\Rsh$}& 5 & 0.5 & \Chart{73.80}{0.531}{blue}{34}{12}\hphantom{9999.} & \Chart{47.44}{0.364}{red}{25}{12}\hphantom{9999.} & \Chart{45.67}{0.345}{teal}{25}{12}\hphantom{9999.}& 2 & 1.0 & \Chart{41.56}{0.429}{blue}{25}{12}\hphantom{9999.} & \Chart{16.80}{0.253}{red}{25}{12}\hphantom{9999.} & \Chart{14.93}{0.232}{teal}{25}{12}\hphantom{9999.} \\
			\cellcolor{white} & LoFTR\hphantom{$\Rsh$}& -- & 1.0 & \Chart{68.65}{0.420}{blue}{25}{12}\hphantom{9999.} & \Chart{41.07}{0.259}{red}{25}{12}\hphantom{9999.} & \Chart{39.30}{0.241}{teal}{25}{12}\hphantom{9999.}& 5 & 1.0 & \Chart{48.86}{0.527}{blue}{34}{12}\hphantom{9999.} & \Chart{20.45}{0.308}{red}{25}{12}\hphantom{9999.} & \Chart{18.37}{0.285}{teal}{25}{12}\hphantom{9999.} \\
			\rowcolor{gray!15}\cellcolor{white} & SE2-LoFTR\hphantom{$\Rsh$}& 5 & 0.5 & \Chart{77.54}{0.612}{blue}{44}{12}\hphantom{9999.} & \Chart{51.94}{0.438}{red}{34}{12}\hphantom{9999.} & \Chart{50.44}{0.423}{teal}{25}{12}\hphantom{9999.}& 2 & 0.5 & \Chart{49.41}{0.534}{blue}{34}{12}\hphantom{9999.} & \Chart{21.04}{0.317}{red}{25}{12}\hphantom{9999.} & \Chart{18.83}{0.292}{teal}{25}{12}\hphantom{9999.} \\
			\cellcolor{white} & MatchFormer\hphantom{$\Rsh$}& 2 & 0.5 & \Chart{71.15}{0.474}{blue}{34}{12}\hphantom{9999.} & \Chart{45.60}{0.333}{red}{25}{12}\hphantom{9999.} & \Chart{43.86}{0.315}{teal}{25}{12}\hphantom{9999.}& 3 & 0.5 & \Chart{45.32}{0.479}{blue}{34}{12}\hphantom{9999.} & \Chart{18.58}{0.280}{red}{25}{12}\hphantom{9999.} & \Chart{16.44}{0.255}{teal}{25}{12}\hphantom{9999.} \\
			\rowcolor{gray!15}\cellcolor{white} & QuadTree Att.\hphantom{$\Rsh$}& 5 & 0.5 & \Chart{78.49}{0.632}{blue}{44}{12}\hphantom{9999.} & \Chart{52.54}{0.447}{red}{34}{12}\hphantom{9999.} & \Chart{51.12}{0.434}{teal}{34}{12}\hphantom{9999.}& 3 & 0.5 & \Chart{51.91}{0.567}{blue}{34}{12}\hphantom{9999.} & \Chart{23.53}{0.355}{red}{25}{12}\hphantom{9999.} & \Chart{21.21}{0.329}{teal}{25}{12}\hphantom{9999.} \\
			\cellcolor{white} \multirow[c]{-59}[59]{*}{\rotatebox{90}{@5$^\circ$}}  & DKM\hphantom{$\Rsh$}& 5 & 0.5 & \Chart{80.91}{0.685}{blue}{44}{12}\hphantom{9999.} & \Chart{57.63}{0.531}{red}{34}{12}\hphantom{9999.} & \Chart{56.43}{0.521}{teal}{34}{12}\hphantom{9999.}& 3 & 0.5 & \Chart{56.14}{0.624}{blue}{44}{12}\hphantom{9999.} & \Chart{26.74}{0.404}{red}{25}{12}\hphantom{9999.} & \Chart{24.37}{0.378}{teal}{25}{12}\hphantom{9999.} \\
			\midrule
			\rowcolor{gray!15}\cellcolor{white} & RootSIFT\hphantom{$\Rsh$}& 5 & 0.5 & \Chart{61.57}{0.267}{blue}{25}{12}\hphantom{9999.} & \Chart{41.45}{0.265}{red}{25}{12}\hphantom{9999.} & \Chart{40.58}{0.261}{teal}{25}{12}\hphantom{9999.}& 5 & 1.0 & \Chart{21.02}{0.153}{blue}{25}{12}\hphantom{9999.} & \Chart{6.83}{0.101}{red}{25}{12}\hphantom{99999.} & \Chart{6.26}{0.097}{teal}{25}{12}\hphantom{99999.} \\
			\cellcolor{white} & RootSIFT$^\Rsh$& 3 & 0.5 & \Chart{61.17}{0.258}{blue}{25}{12}\hphantom{9999.} & \Chart{40.88}{0.256}{red}{25}{12}\hphantom{9999.} & \Chart{39.91}{0.251}{teal}{25}{12}\hphantom{9999.}& 5 & 1.0 & \Chart{23.47}{0.186}{blue}{25}{12}\hphantom{9999.} & \Chart{8.16}{0.121}{red}{25}{12}\hphantom{99999.} & \Chart{7.38}{0.114}{teal}{25}{12}\hphantom{99999.} \\
			\rowcolor{gray!15}\cellcolor{white} & Key.Net\hphantom{$\Rsh$}& 3 & 0.5 & \Chart{81.81}{0.704}{blue}{44}{12}\hphantom{9999.} & \Chart{60.15}{0.573}{red}{34}{12}\hphantom{9999.} & \Chart{59.02}{0.563}{teal}{34}{12}\hphantom{9999.}& 2 & 0.5 & \Chart{46.05}{0.489}{blue}{34}{12}\hphantom{9999.} & \Chart{21.81}{0.329}{red}{25}{12}\hphantom{9999.} & \Chart{20.95}{0.325}{teal}{25}{12}\hphantom{9999.} \\
			\cellcolor{white} & Key.Net$^\Rsh$& 2 & 0.5 & \Chart{83.02}{0.730}{blue}{44}{12}\hphantom{9999.} & \Chart{62.97}{0.619}{red}{44}{12}\hphantom{9999.} & \Chart{61.71}{0.607}{teal}{44}{12}\hphantom{9999.}& 5 & 0.5 & \Chart{44.92}{0.474}{blue}{34}{12}\hphantom{9999.} & \Chart{21.93}{0.331}{red}{25}{12}\hphantom{9999.} & \Chart{20.94}{0.325}{teal}{25}{12}\hphantom{9999.} \\
			\rowcolor{gray!15}\cellcolor{white} & Hz$^+$\hphantom{$\Rsh$}& 5 & 0.5 & \Chart{83.72}{0.745}{blue}{53}{12}\hphantom{9999.} & \Chart{63.56}{0.629}{red}{44}{12}\hphantom{9999.} & \Chart{62.50}{0.620}{teal}{44}{12}\hphantom{9999.}& 2 & 0.5 & \Chart{45.42}{0.480}{blue}{34}{12}\hphantom{9999.} & \Chart{23.53}{0.355}{red}{25}{12}\hphantom{9999.} & \Chart{22.38}{0.347}{teal}{25}{12}\hphantom{9999.} \\
			\cellcolor{white} & Hz$^{+\Rsh}$& 2 & 0.5 & \Chart{84.50}{0.762}{blue}{53}{12}\hphantom{9999.} & \Chart{65.04}{0.653}{red}{44}{12}\hphantom{9999.} & \Chart{63.96}{0.644}{teal}{44}{12}\hphantom{9999.}& 3 & 0.5 & \Chart{45.95}{0.488}{blue}{34}{12}\hphantom{9999.} & \Chart{23.29}{0.352}{red}{25}{12}\hphantom{9999.} & \Chart{22.01}{0.341}{teal}{25}{12}\hphantom{9999.} \\
			\rowcolor{gray!15}\cellcolor{white} & Slimed Hz$^+$\hphantom{$\Rsh$}& 2 & 0.5 & \Chart{82.37}{0.716}{blue}{44}{12}\hphantom{9999.} & \Chart{61.34}{0.592}{red}{34}{12}\hphantom{9999.} & \Chart{60.14}{0.581}{teal}{34}{12}\hphantom{9999.}& 2 & 0.5 & \Chart{54.65}{0.604}{blue}{34}{12}\hphantom{9999.} & \Chart{30.45}{0.461}{red}{34}{12}\hphantom{9999.} & \Chart{29.28}{0.454}{teal}{34}{12}\hphantom{9999.} \\
			\cellcolor{white} & Slimed Hz$^{+\Rsh}$& 2 & 0.5 & \Chart{83.11}{0.732}{blue}{44}{12}\hphantom{9999.} & \Chart{60.50}{0.578}{red}{34}{12}\hphantom{9999.} & \Chart{59.43}{0.570}{teal}{34}{12}\hphantom{9999.}& 3 & 0.5 & \Chart{53.19}{0.585}{blue}{34}{12}\hphantom{9999.} & \Chart{29.02}{0.439}{red}{34}{12}\hphantom{9999.} & \Chart{27.82}{0.432}{teal}{34}{12}\hphantom{9999.} \\
			\rowcolor{gray!15}\cellcolor{white} & DISK\hphantom{$\Rsh$}& 5 & 0.5 & \Chart{79.32}{0.650}{blue}{44}{12}\hphantom{9999.} & \Chart{58.65}{0.548}{red}{34}{12}\hphantom{9999.} & \Chart{57.79}{0.543}{teal}{34}{12}\hphantom{9999.}& 3 & 1.0 & \Chart{46.28}{0.492}{blue}{34}{12}\hphantom{9999.} & \Chart{21.63}{0.326}{red}{25}{12}\hphantom{9999.} & \Chart{20.30}{0.315}{teal}{25}{12}\hphantom{9999.} \\
			\cellcolor{white} & SuperGlue\hphantom{$\Rsh$}& 5 & 0.5 & \Chart{84.68}{0.766}{blue}{53}{12}\hphantom{9999.} & \Chart{64.02}{0.636}{red}{44}{12}\hphantom{9999.} & \Chart{62.79}{0.625}{teal}{44}{12}\hphantom{9999.}& 5 & 1.0 & \Chart{58.49}{0.656}{blue}{44}{12}\hphantom{9999.} & \Chart{32.75}{0.496}{red}{34}{12}\hphantom{9999.} & \Chart{30.86}{0.479}{teal}{34}{12}\hphantom{9999.} \\
			\rowcolor{gray!15}\cellcolor{white} & LoFTR\hphantom{$\Rsh$}& -- & 1.0 & \Chart{81.92}{0.706}{blue}{44}{12}\hphantom{9999.} & \Chart{58.49}{0.545}{red}{34}{12}\hphantom{9999.} & \Chart{57.32}{0.535}{teal}{34}{12}\hphantom{9999.}& 5 & 1.0 & \Chart{63.18}{0.718}{blue}{44}{12}\hphantom{9999.} & \Chart{37.11}{0.562}{red}{34}{12}\hphantom{9999.} & \Chart{35.45}{0.550}{teal}{34}{12}\hphantom{9999.} \\
			\cellcolor{white} & SE2-LoFTR\hphantom{$\Rsh$}& 5 & 0.5 & \Chart{87.28}{0.822}{blue}{53}{12}\hphantom{9999.} & \Chart{67.92}{0.700}{red}{44}{12}\hphantom{9999.} & \Chart{67.11}{0.695}{teal}{44}{12}\hphantom{9999.}& 5 & 0.5 & \Chart{63.98}{0.729}{blue}{44}{12}\hphantom{9999.} & \Chart{37.89}{0.574}{red}{34}{12}\hphantom{9999.} & \Chart{36.45}{0.565}{teal}{34}{12}\hphantom{9999.} \\
			\rowcolor{gray!15}\cellcolor{white} & MatchFormer\hphantom{$\Rsh$}& 2 & 0.5 & \Chart{83.22}{0.735}{blue}{44}{12}\hphantom{9999.} & \Chart{62.09}{0.604}{red}{34}{12}\hphantom{9999.} & \Chart{60.94}{0.594}{teal}{34}{12}\hphantom{9999.}& 3 & 0.5 & \Chart{60.25}{0.679}{blue}{44}{12}\hphantom{9999.} & \Chart{34.38}{0.520}{red}{34}{12}\hphantom{9999.} & \Chart{32.81}{0.509}{teal}{34}{12}\hphantom{9999.} \\
			\cellcolor{white} & QuadTree Att.\hphantom{$\Rsh$}& 5 & 0.5 & \Chart{87.88}{0.835}{blue}{53}{12}\hphantom{9999.} & \Chart{68.55}{0.711}{red}{44}{12}\hphantom{9999.} & \Chart{67.76}{0.706}{teal}{44}{12}\hphantom{9999.}& 3 & 0.5 & \Chart{66.31}{0.760}{blue}{53}{12}\hphantom{9999.} & \Chart{40.96}{0.621}{red}{44}{12}\hphantom{9999.} & \Chart{39.05}{0.606}{teal}{44}{12}\hphantom{9999.} \\
			\rowcolor{gray!15}\cellcolor{white} \multirow[c]{-59}[59]{*}{\rotatebox{90}{@10$^\circ$}}  & DKM\hphantom{$\Rsh$}& 5 & 0.5 & \Chart{88.85}{0.856}{blue}{62}{12}\hphantom{9999.} & \Chart{72.57}{0.777}{red}{53}{12}\hphantom{9999.} & \Chart{71.87}{0.773}{teal}{53}{12}\hphantom{9999.}& 3 & 0.5 & \Chart{71.16}{0.825}{blue}{53}{12}\hphantom{9999.} & \Chart{46.11}{0.699}{red}{44}{12}\hphantom{9999.} & \Chart{44.27}{0.687}{teal}{44}{12}\hphantom{9999.} \\
			\midrule
			\cellcolor{white} & RootSIFT\hphantom{$\Rsh$}& 5 & 1.0 & \Chart{70.01}{0.449}{blue}{34}{12}\hphantom{9999.} & \Chart{54.01}{0.472}{red}{34}{12}\hphantom{9999.} & \Chart{52.18}{0.451}{teal}{34}{12}\hphantom{9999.}& 5 & 1.0 & \Chart{29.09}{0.262}{blue}{25}{12}\hphantom{9999.} & \Chart{13.13}{0.197}{red}{25}{12}\hphantom{9999.} & \Chart{11.57}{0.179}{teal}{25}{12}\hphantom{9999.} \\
			\rowcolor{gray!15}\cellcolor{white} & RootSIFT$^\Rsh$& 3 & 0.5 & \Chart{69.17}{0.431}{blue}{34}{12}\hphantom{9999.} & \Chart{53.85}{0.469}{red}{34}{12}\hphantom{9999.} & \Chart{52.20}{0.452}{teal}{34}{12}\hphantom{9999.}& 5 & 1.0 & \Chart{32.22}{0.303}{blue}{25}{12}\hphantom{9999.} & \Chart{15.04}{0.226}{red}{25}{12}\hphantom{9999.} & \Chart{13.56}{0.210}{teal}{25}{12}\hphantom{9999.} \\
			\cellcolor{white} & Key.Net\hphantom{$\Rsh$}& -- & 0.5 & \Chart{89.18}{0.863}{blue}{62}{12}\hphantom{9999.} & \Chart{74.41}{0.807}{red}{53}{12}\hphantom{9999.} & \Chart{73.59}{0.801}{teal}{53}{12}\hphantom{9999.}& 5 & 1.0 & \Chart{57.05}{0.636}{blue}{44}{12}\hphantom{9999.} & \Chart{34.97}{0.529}{red}{34}{12}\hphantom{9999.} & \Chart{33.65}{0.522}{teal}{34}{12}\hphantom{9999.} \\
			\rowcolor{gray!15}\cellcolor{white} & Key.Net$^\Rsh$& 2 & 0.5 & \Chart{89.86}{0.878}{blue}{62}{12}\hphantom{9999.} & \Chart{76.46}{0.841}{red}{53}{12}\hphantom{9999.} & \Chart{75.57}{0.834}{teal}{53}{12}\hphantom{9999.}& -- & 1.0 & \Chart{57.29}{0.640}{blue}{44}{12}\hphantom{9999.} & \Chart{35.28}{0.534}{red}{34}{12}\hphantom{9999.} & \Chart{34.19}{0.530}{teal}{34}{12}\hphantom{9999.} \\
			\cellcolor{white} & Hz$^+$\hphantom{$\Rsh$}& 5 & 0.5 & \Chart{90.06}{0.882}{blue}{62}{12}\hphantom{9999.} & \Chart{76.33}{0.839}{red}{53}{12}\hphantom{9999.} & \Chart{75.56}{0.834}{teal}{53}{12}\hphantom{9999.}& 3 & 0.5 & \Chart{55.94}{0.621}{blue}{44}{12}\hphantom{9999.} & \Chart{35.95}{0.544}{red}{34}{12}\hphantom{9999.} & \Chart{34.69}{0.538}{teal}{34}{12}\hphantom{9999.} \\
			\rowcolor{gray!15}\cellcolor{white} & Hz$^{+\Rsh}$& 2 & 0.5 & \Chart{90.25}{0.886}{blue}{62}{12}\hphantom{9999.} & \Chart{77.80}{0.863}{red}{62}{12}\hphantom{9999.} & \Chart{77.02}{0.857}{teal}{62}{12}\hphantom{9999.}& 5 & 1.0 & \Chart{57.57}{0.643}{blue}{44}{12}\hphantom{9999.} & \Chart{36.07}{0.546}{red}{34}{12}\hphantom{9999.} & \Chart{34.97}{0.543}{teal}{34}{12}\hphantom{9999.} \\
			\cellcolor{white} & Slimed Hz$^+$\hphantom{$\Rsh$}& 2 & 0.5 & \Chart{89.31}{0.866}{blue}{62}{12}\hphantom{9999.} & \Chart{75.01}{0.817}{red}{53}{12}\hphantom{9999.} & \Chart{74.09}{0.809}{teal}{53}{12}\hphantom{9999.}& 3 & 0.5 & \Chart{65.39}{0.748}{blue}{53}{12}\hphantom{9999.} & \Chart{45.54}{0.690}{red}{44}{12}\hphantom{9999.} & \Chart{44.35}{0.688}{teal}{44}{12}\hphantom{9999.} \\
			\rowcolor{gray!15}\cellcolor{white} & Slimed Hz$^{+\Rsh}$& 2 & 0.5 & \Chart{89.89}{0.879}{blue}{62}{12}\hphantom{9999.} & \Chart{74.71}{0.812}{red}{53}{12}\hphantom{9999.} & \Chart{73.98}{0.808}{teal}{53}{12}\hphantom{9999.}& -- & 1.0 & \Chart{63.37}{0.721}{blue}{44}{12}\hphantom{9999.} & \Chart{42.93}{0.651}{red}{44}{12}\hphantom{9999.} & \Chart{42.02}{0.652}{teal}{44}{12}\hphantom{9999.} \\
			\cellcolor{white} & DISK\hphantom{$\Rsh$}& 5 & 0.5 & \Chart{85.25}{0.778}{blue}{53}{12}\hphantom{9999.} & \Chart{71.48}{0.759}{red}{53}{12}\hphantom{9999.} & \Chart{70.56}{0.752}{teal}{53}{12}\hphantom{9999.}& 3 & 1.0 & \Chart{57.51}{0.642}{blue}{44}{12}\hphantom{9999.} & \Chart{35.75}{0.541}{red}{34}{12}\hphantom{9999.} & \Chart{34.02}{0.528}{teal}{34}{12}\hphantom{9999.} \\
			\rowcolor{gray!15}\cellcolor{white} & SuperGlue\hphantom{$\Rsh$}& 5 & 0.5 & \Chart{91.23}{0.908}{blue}{62}{12}\hphantom{9999.} & \Chart{77.86}{0.864}{red}{62}{12}\hphantom{9999.} & \Chart{76.99}{0.857}{teal}{62}{12}\hphantom{9999.}& 5 & 1.0 & \Chart{70.88}{0.822}{blue}{53}{12}\hphantom{9999.} & \Chart{49.40}{0.749}{red}{53}{12}\hphantom{9999.} & \Chart{47.94}{0.744}{teal}{53}{12}\hphantom{9999.} \\
			\cellcolor{white} & LoFTR\hphantom{$\Rsh$}& 2 & 1.0 & \Chart{89.82}{0.877}{blue}{62}{12}\hphantom{9999.} & \Chart{73.42}{0.791}{red}{53}{12}\hphantom{9999.} & \Chart{72.40}{0.782}{teal}{53}{12}\hphantom{9999.}& 5 & 1.0 & \Chart{72.68}{0.846}{blue}{53}{12}\hphantom{9999.} & \Chart{53.67}{0.814}{red}{53}{12}\hphantom{9999.} & \Chart{51.97}{0.806}{teal}{53}{12}\hphantom{9999.} \\
			\rowcolor{gray!15}\cellcolor{white} & SE2-LoFTR\hphantom{$\Rsh$}& 5 & 0.5 & \Chart{92.70}{0.939}{blue}{62}{12}\hphantom{9999.} & \Chart{80.17}{0.902}{red}{62}{12}\hphantom{9999.} & \Chart{79.60}{0.900}{teal}{62}{12}\hphantom{9999.}& 5 & 1.0 & \Chart{75.49}{0.884}{blue}{62}{12}\hphantom{9999.} & \Chart{55.16}{0.837}{red}{53}{12}\hphantom{9999.} & \Chart{53.76}{0.834}{teal}{53}{12}\hphantom{9999.} \\
			\cellcolor{white} & MatchFormer\hphantom{$\Rsh$}& 2 & 0.5 & \Chart{90.52}{0.892}{blue}{62}{12}\hphantom{9999.} & \Chart{75.46}{0.824}{red}{53}{12}\hphantom{9999.} & \Chart{74.80}{0.821}{teal}{53}{12}\hphantom{9999.}& -- & 1.0 & \Chart{71.44}{0.829}{blue}{53}{12}\hphantom{9999.} & \Chart{50.88}{0.772}{red}{53}{12}\hphantom{9999.} & \Chart{49.63}{0.770}{teal}{53}{12}\hphantom{9999.} \\
			\rowcolor{gray!15}\cellcolor{white} & QuadTree Att.\hphantom{$\Rsh$}& -- & 0.5 & \Chart{92.57}{0.937}{blue}{62}{12}\hphantom{9999.} & \Chart{81.10}{0.917}{red}{62}{12}\hphantom{9999.} & \Chart{80.52}{0.915}{teal}{62}{12}\hphantom{9999.}& 2 & 0.5 & \Chart{75.26}{0.880}{blue}{62}{12}\hphantom{9999.} & \Chart{57.23}{0.868}{red}{62}{12}\hphantom{9999.} & \Chart{55.70}{0.864}{teal}{62}{12}\hphantom{9999.} \\
			\cellcolor{white} \multirow[c]{-59}[59]{*}{\rotatebox{90}{@20$^\circ$}}  & DKM\hphantom{$\Rsh$}& -- & 0.5 & \Chart{93.40}{0.955}{blue}{62}{12}\hphantom{9999.} & \Chart{83.39}{0.955}{red}{62}{12}\hphantom{9999.} & \Chart{82.96}{0.955}{teal}{62}{12}\hphantom{9999.}& 5 & 0.5 & \Chart{80.79}{0.955}{blue}{62}{12}\hphantom{9999.} & \Chart{62.90}{0.955}{red}{62}{12}\hphantom{9999.} & \Chart{61.52}{0.954}{teal}{62}{12}\hphantom{9999.} \\
			\bottomrule
		\end{tabular}
	}
	\vspace{4em}
\end{table*}

\begin{table}[t!]
	\renewcommand{\arraystretch}{0}
	\setlength{\tabcolsep}{4pt}
	\centering
	\caption{Result comparison between the IMC-PT validation dataset and MegaDepth (see SM\ref{sm_res3}). Best viewed in color and zoomed in.}\label{imc_scannet}	
	\resizebox{0.3\textwidth}{!}{
		\begin{tabular}{r<{}c<{}c}
			& {IMC-PT val.} & {MegaDepth}\\[5pt]
			& {\footnotesize mAA@10$^\circ$ (\%)} & {\footnotesize AUC@10$^\circ$ (\%)} \\
			\midrule
			RootSIFT\hphantom{$\Rsh$} & \Chart{10.39}{0.045}{orange}{25}{12}\hphantom{9999.} & \Chart{40.58}{0.065}{blue}{25}{12}\hphantom{9999.} \\
			\rowcolor{gray!15}RootSIFT$^\Rsh$ & \Chart{12.47}{0.072}{orange}{25}{12}\hphantom{9999.} & \Chart{39.91}{0.045}{blue}{25}{12}\hphantom{9999.} \\
			Key.Net\hphantom{$\Rsh$} & \Chart{49.17}{0.541}{orange}{34}{12}\hphantom{9999.} & \Chart{59.02}{0.589}{blue}{34}{12}\hphantom{9999.} \\
			\rowcolor{gray!15}Key.Net$^\Rsh$ & \Chart{52.46}{0.583}{orange}{34}{12}\hphantom{9999.} & \Chart{61.71}{0.666}{blue}{44}{12}\hphantom{9999.} \\
			Hz$^+$\hphantom{$\Rsh$} & \Chart{52.00}{0.577}{orange}{34}{12}\hphantom{9999.} & \Chart{62.50}{0.688}{blue}{44}{12}\hphantom{9999.} \\
			\rowcolor{gray!15}Hz$^{+\Rsh}$ & \Chart{58.39}{0.658}{orange}{44}{12}\hphantom{9999.} & \Chart{63.96}{0.730}{blue}{44}{12}\hphantom{9999.} \\
			Slimed Hz$^+$\hphantom{$\Rsh$} & \Chart{54.60}{0.610}{orange}{44}{12}\hphantom{9999.} & \Chart{60.14}{0.621}{blue}{44}{12}\hphantom{9999.} \\
			\rowcolor{gray!15}DISK\hphantom{$\Rsh$} & \Chart{57.46}{0.646}{orange}{44}{12}\hphantom{9999.} & \Chart{57.79}{0.554}{blue}{34}{12}\hphantom{9999.} \\
			SuperGlue\hphantom{$\Rsh$} & \Chart{62.45}{0.710}{orange}{44}{12}\hphantom{9999.} & \Chart{62.79}{0.696}{blue}{44}{12}\hphantom{9999.} \\
			\rowcolor{gray!15}LoFTR\hphantom{$\Rsh$} & \Chart{59.92}{0.678}{orange}{44}{12}\hphantom{9999.} & \Chart{57.32}{0.541}{blue}{34}{12}\hphantom{9999.} \\
			SE2-LoFTR\hphantom{$\Rsh$} & \Chart{75.99}{0.883}{orange}{62}{12}\hphantom{9999.} & \Chart{67.11}{0.819}{blue}{53}{12}\hphantom{9999.} \\
			\rowcolor{gray!15}MatchFormer\hphantom{$\Rsh$} & \Chart{60.09}{0.680}{orange}{44}{12}\hphantom{9999.} & \Chart{60.94}{0.644}{blue}{44}{12}\hphantom{9999.} \\
			QuadTree Att.\hphantom{$\Rsh$} & \Chart{81.60}{0.955}{orange}{62}{12}\hphantom{9999.} & \Chart{67.76}{0.838}{blue}{53}{12}\hphantom{9999.} \\
			\rowcolor{gray!15}DKM\hphantom{$\Rsh$} & \Chart{76.71}{0.892}{orange}{62}{12}\hphantom{9999.} & \Chart{71.87}{0.955}{blue}{62}{12}\hphantom{9999.} \\
		\end{tabular}
	}
\end{table}

\subsection{Additional comparisons on non-planar scenes}\label{sm_res3}
Table~\ref{imc_scannet} compares the pose error in terms of mAA on the IMC-PT validation dataset using the standard protocol of~\cite{imw2020} with the pose error in terms of AUC on MegaDepth for the same 10$^\circ$ angular threshold. The upright version of Slimed Hz$^+$ is not included since on all the above analyses it has given worse results than its base version. It can be observed that results are quite correlated, so that mainly the same considerations drawn out in Sec.~\ref{mdepth_scannet} for MegaDepth holds. Notice that the kind of scenes of the two datasets is very similar and the both have three scenes of which one in common. Moreover, the current computation of the mAA is quite similar to that of the AUC.     

Actually, an additional comparison on the SUN3D indoor dataset, which include camera pose GT data, has been tempted using the protocol described in~\cite{dtm}. Unlike the original comparison which considered coverage and precision as error metric, the error was evaluated on the accuracy following the current trend of focusing on the camera orientation, see SM\ref{sm_metric}. While in terms of rank the result was similar to the accuracy on the non-planar dataset discussed in Sec.~\ref{npsec}, in terms of absolute values the accuracy did not exceed 5\%.

\begin{figure}[t]
	\center
	\includegraphics[width=0.15\textwidth]{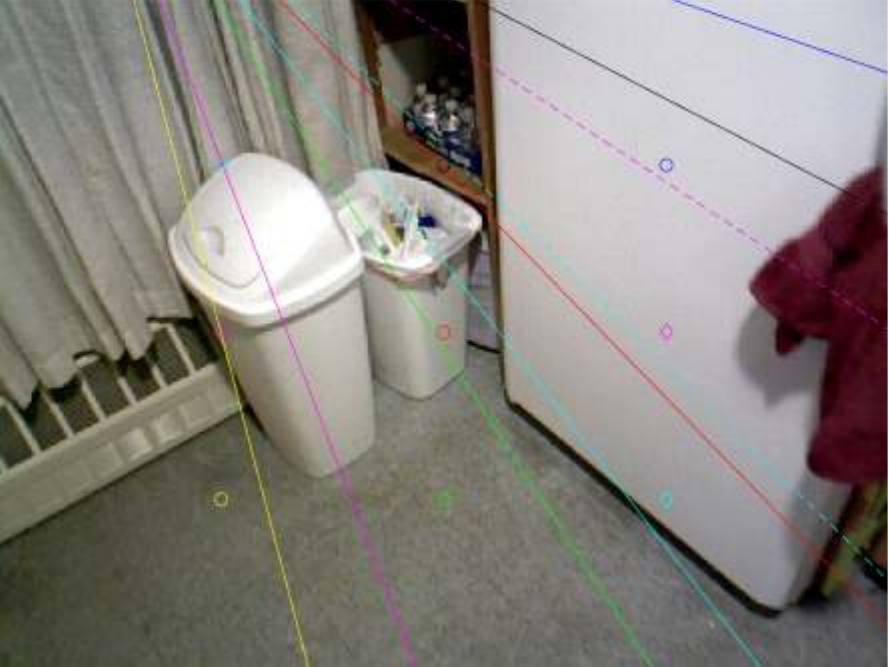}
	\includegraphics[width=0.15\textwidth]{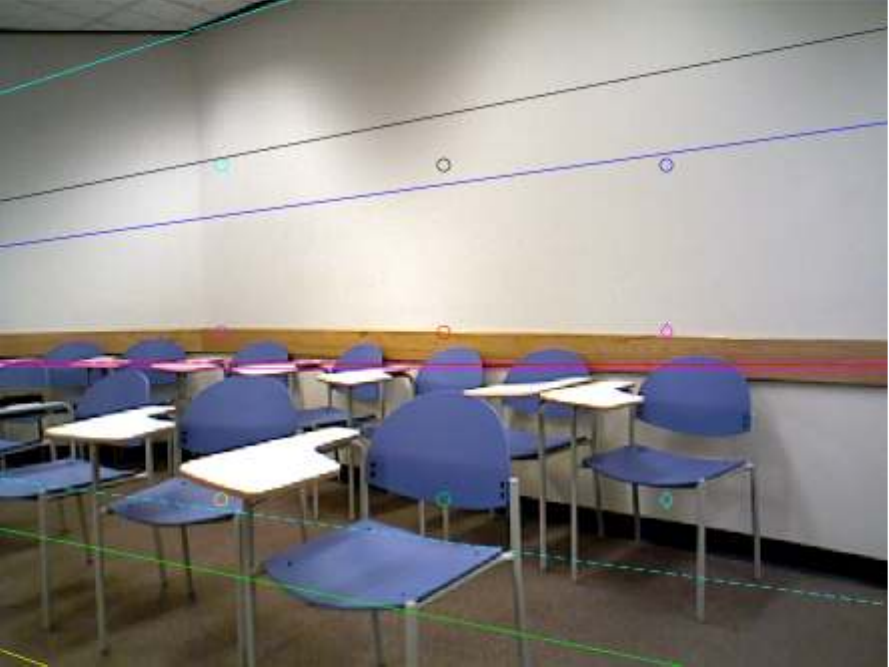}
	\includegraphics[width=0.15\textwidth]{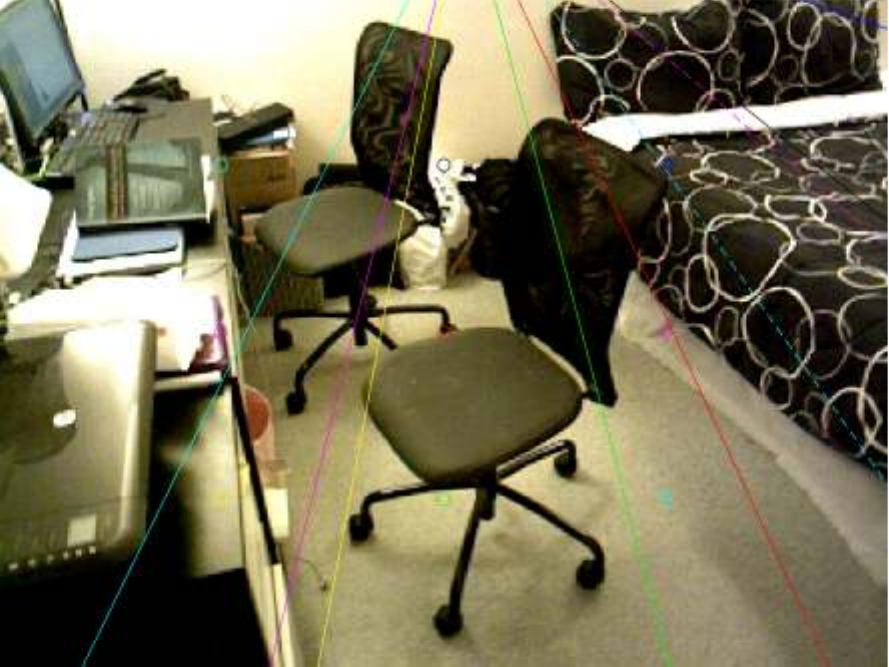} \\
	\vspace{0.3em}
	\includegraphics[width=0.15\textwidth]{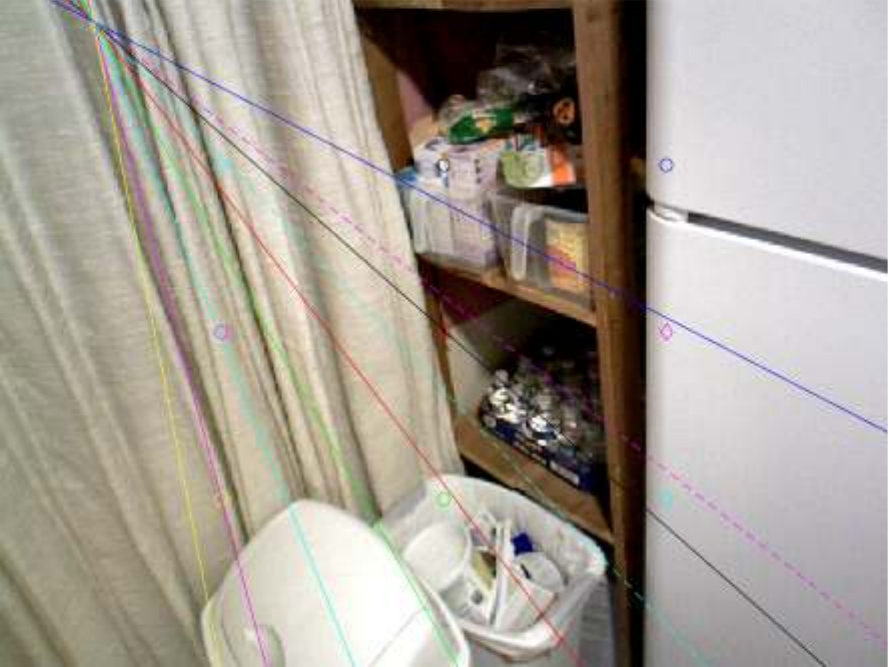}
	\includegraphics[width=0.15\textwidth]{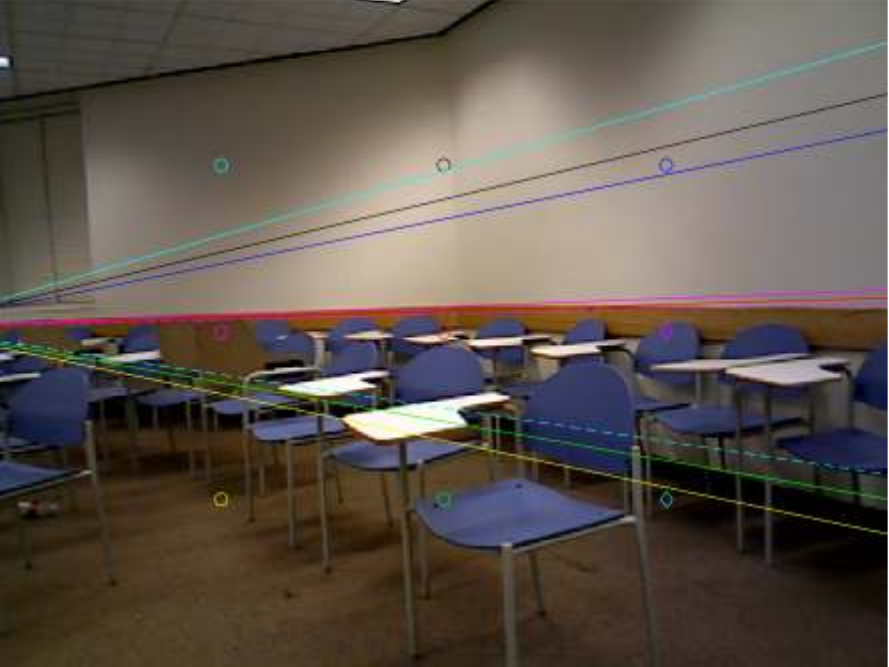}
	\includegraphics[width=0.15\textwidth]{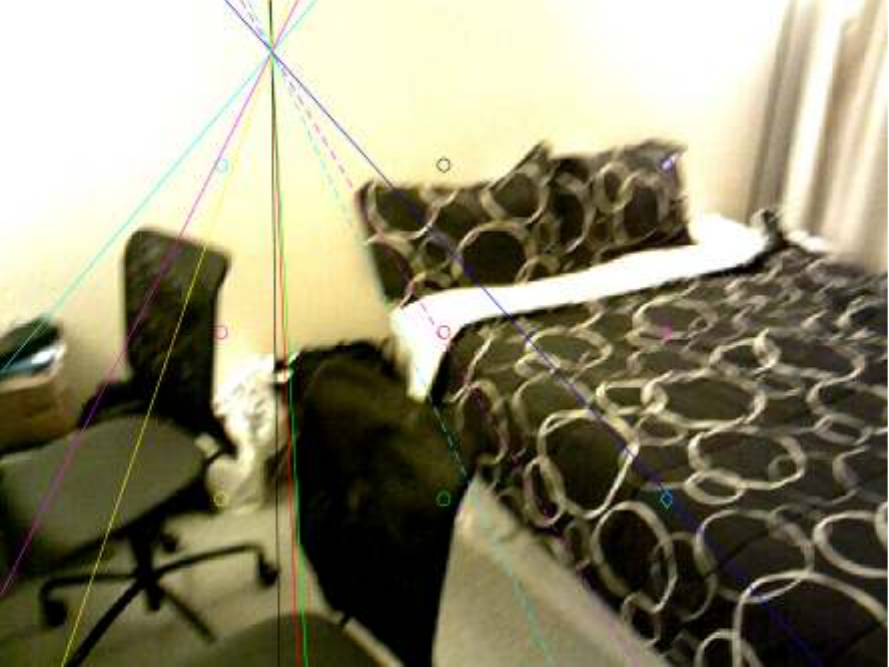}
	\caption{\label{sun}
		Visual qualitative check for the SUN3D GT camera poses. Each column represents a different image pair. In the images, for each of the six point marked the corresponding epipolar line is visualized, if inside the image, on the other image. Within the same color, circle marks are paired with the solid lines and diamond marks with dashed lines. It can be observed, for instance, that on the left image pair, the solid line in red on the top image that should correspond to the red circle mark in the bottom image is clearly unaligned. The same holds for the middle image pair considering the solid yellow line in the bottom image, and for the right image pair in the case of the solid yellow line in the top image (see SM\ref{sm_res3}). Best viewed in color and zoomed in.}
\end{figure}

This weird behavior is due to the inaccurate GT pose estimation provided within the dataset, with camera poses that quickly accumulate errors so that the relative pose becomes inconsistent in the interval of about 80 frames that divides the tested paired images. Figure~\ref{sun} shows visual examples of the incorrect epipolar geometry provided by the GT to validate this statement. For this reason the quantitative comparative results are omitted. Notice however that the precision of the camera pose GT estimation is sufficient to discriminate correct and wrong matches with a loose threshold, as one can visually check in~\cite{dtm}.

\begin{figure}[t]
	\center
	\subfloat[\small EVD]{\label{evd_plot}
		\includegraphics[width=0.43\textwidth]{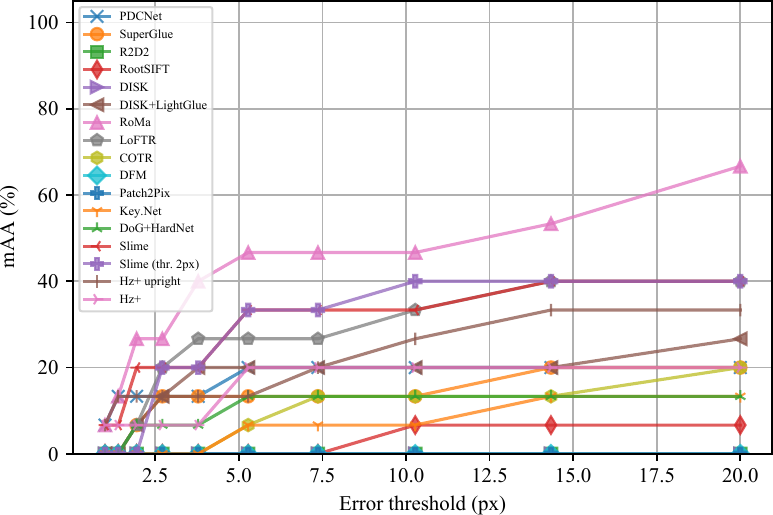}
	}
	\\	
	\subfloat[\small WxBS dataset]{\label{wxbs_plot}
		\includegraphics[width=0.43\textwidth]{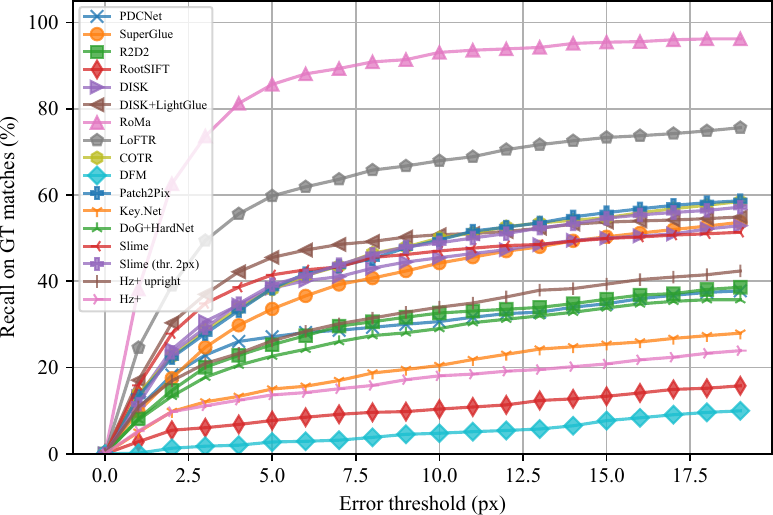}
	}
	\vspace{0.5em}	
	\caption{Comparative evaluation on \protect\subref{evd_plot} EVD and \protect\subref{wxbs_plot} the WxBS dataset (see SM\ref{sm_res4}). Best viewed in color and zoomed in.}\label{evd_wxbs}
\end{figure}

\subsection{Additional evaluations on matching robustness}\label{sm_res4}
Figure~\ref{evd_wxbs} shows Slime compared against several SOTA matching pipelines on the Extreme Viewpoint Dataset (EVD)~\cite{mods} and the Wide Baseline Stereo (WxBS) dataset~\cite{wxbs} following to the default protocol described \href{https://ducha-aiki.github.io/wxbs-benchmark/}{here}. Specifically, EVD contains planar images subject to strong viewpoint changes and error is given in terms of mAA as the average mean reprojection error on the common image part with respect to the GT on different error thresholds. The WxBS dataset presents instead challenging non-planar scene transformations where the error is measured in terms of recall on GT matches, defined as the percentage of GT correspondences consistent with estimated fundamental matrix. In both cases, the matches are processed using MAGSAC++~[\hyperref[magsacpp]{66}] with default parameters; Slime is also reported by applying the implemented RANSAC with a threshold of 2 px before MAGSAC++, analogously to MegaDepth and ScanNet evaluations. These datasets are both designed to highlight more the match robustness to extreme scene changes than measuring the localization precision of the matches and, although small, are quite hard.

Slime is remarkably better than both the base and the upright Hz$^+$ on both EVD and the WxBS dataset, being the latter matching method generally better than its non-upright counterpart. On one hand, on EVD Slime is only surpassed by the recent Robust losses for dense feature Matching (RoMa) deep architecture~[\hyperref[roma]{67}], which extends DKM, and it is challenged by LoFTR, which achieves slightly lower results tham Slime. On the other hand, on the WxBS dataset the best results are provided again by RoMa, followed by LoFTR. Nevertheless, Slime comes next on par with the other last-generation deep architectures. These matching pipelines are quite better than those belonging to the previous generation, which in turn surpass handcrafted and hybrid matching pipelines. These results corroborate Slime robustness and its boosted performances with respect to other hybrid and handcrafted pipelines in term of ability to discriminate matches.

\subsection{Slime running time}\label{sm_time}
Slime main code is written in Matlab exploiting only a trivial \texttt{parfor} for parallel computation and no further optimizations. The base Hz$^+$ pipeline mixes together Matlab code and PyTorch functions executed through system calls. All the different modules communicates through files with compatible formats, e.g. blocks and tiles are saved as different \texttt{.png} images and keypoint and match data as \texttt{.mat} files that can be read and write in Python by the SciPy library. This guarantees an easier interface management of the different modules and of the data structures, but obviously introduces an overhead in the data access. Furthermore, heavy code optimization are missing in order provide a simpler program prototyping and debugging. 

\begin{figure}[t]
	\center
	\includegraphics[width=0.48\textwidth]{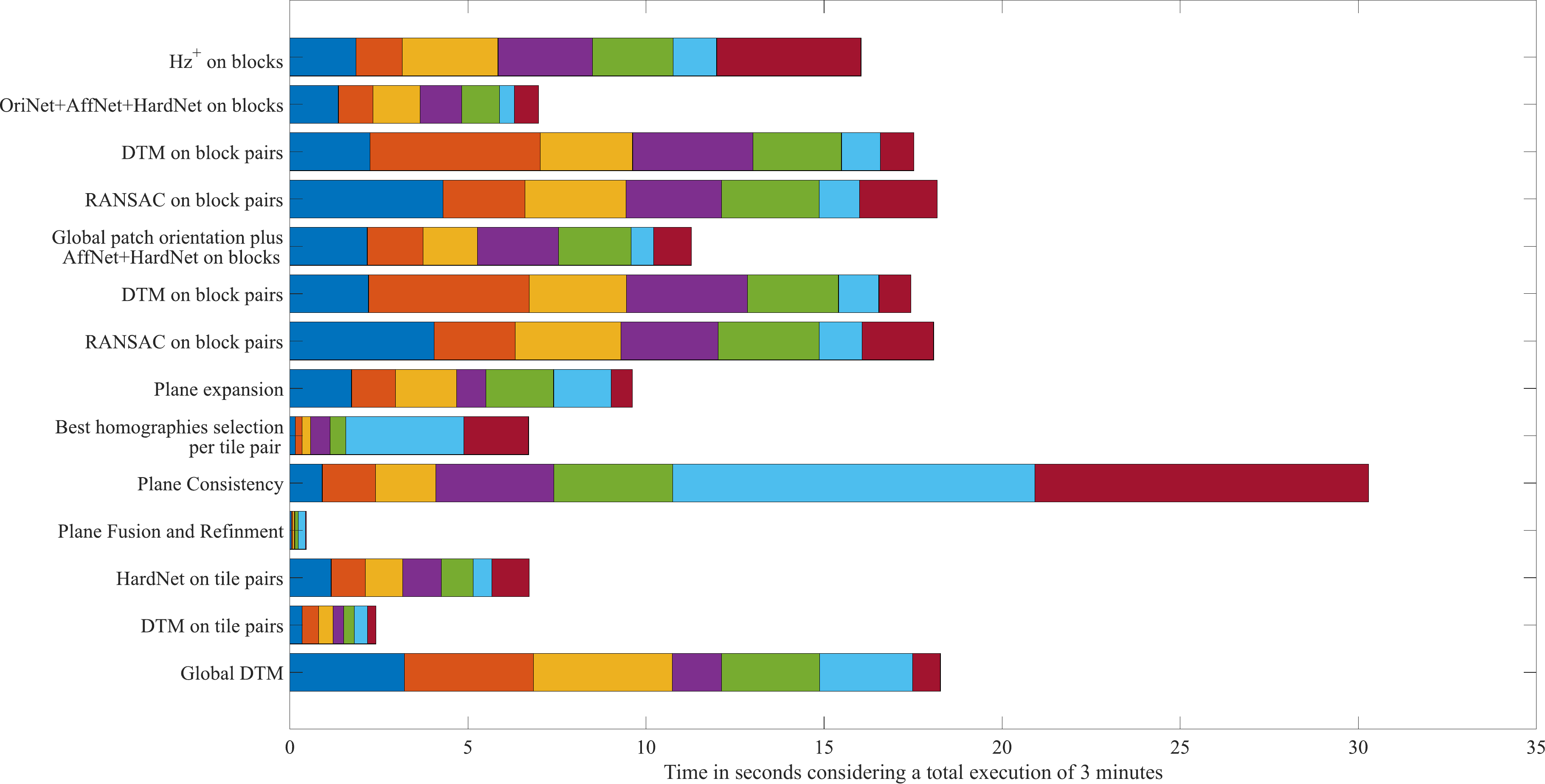} \\
	\caption{\label{time_err}
		Slime running time split over the different steps. Cumulative times for seven different runs are reported together using distinct colors and the total time is normalized by the average execution time (see SM\ref{sm_time}). Best viewed in color and zoomed in.}
\end{figure}

Figure~\ref{time_err} shows the histogram of the execution time of Slime, partitioned by logical execution phases named by their principal tasks. Each bar accumulates the execution time of seven runs on distinct image pairs. The overall time is normalized by the expected average running times of 3 min. The most time-consuming phases are the plane consistency filtering, the keypoint extraction, RANSAC and DTM, the last two executed more than once. The next computationally expensive phases are the patch and description computation, followed by the plane expansion and the best homography selection. Notice that the plane expansion and the best homography selection are relevant only for the last two runs, colored in azure and red. These correspond to easier image pairs where the number of surviving good raw matches is high. Otherwise, their impact on the whole run is reduced. The computational time analysis in the case of other slimed pipelines has not been analyzed but a similar trend is expected.  

Some reasonable strategies for improving the running time of Slime are discussed hereafter:
\begin{itemize}
	\item \textit{Compiled code without system calls and data sharing by files.} This kind of implementation will clearly provide a faster code, also reducing the overhead of interfacing within the different modules.
	\item \textit{Shared data structures to avoid repeated computation.} Each image region is present in nine blocks due to the overlap within them, and likewise for tiles. This means the same computation is repeated roughly nine times during the keypoint extraction, the patch normalization and the descriptor computation. This is a raw estimation since the keypoint extraction is adaptive. Nevertheless, a better integration and shared data structure should avoid this redundant computation.
	\item \textit{Optimized DTM implementation.} The current implementation of DTM is written in Matlab and heavily relies on cells, which are notorious to slow the code execution. Furthermore, DTM spends most of its computation in set operations, specifically in the union and intersection of match sets. DTM was originally designed  to work with about 2K matches, for which the implementation of sets by simple bit arrays provides a quite efficient data structure. This design is not optimal for a input size greater than 8K, which is the expected number of matches by the base Hz$^+$ for a block pair. In this case the code employed in LPM~\cite{lpm} should provide a better implementation. Rewriting DTM code according to these observations could improve the Slime running time.
	\item \textit{RANSAC and plane consistency filtering optimizations.} These two phases spent most of their time in the computation of reprojection and epipolar errors, defined respectively by Eq.~\ref{max_reproj} and Eq.~\ref{max_epi_error}. The current code is written to be readable than efficient. For instance, the fixed 1 as third component of the points in homogeneous coordinates is added at every execution to implement the operation as a matrix multiplication.
	
	Nevertheless, a more critical aspect to consider is that Slime matches are generally arranged as clusters with keypoint positions differentiated by few pixels. This is also due to the overlap within blocks and tiles. The visual results reported along the paper, where the links between the correspondences appear in bold although transparency is used to show them, testify this phenomenon. A cluster of matches can then be replaced by its centroid, sided by a weight indicating the cluster size. This raw representation can decrease the total number of matches during the computation, hence the running time.
	\item \textit{Multi-threaded and parallel execution.} The current Matlab implementation of Slime uses \texttt{parfor} loops to parallelize some parts of the code dealing with the base pipeline matching. Nevertheless, according to the author' experience compiled multi-threaded code is better than the Matlab parallel system which has more overhead. A better code implementation should help.
	\item \textit{Easier image pairs.} In the case of non-complex scenes the amount of matches elaborated across the Slime pipeline is higher, as the method tends to discover correspondences over all the common region areas. In this case, eliminating some block levels or the tiles from the computation can be a possible solution to improve the speed, as an early stopping approach following the strategy of LightGlue. 
\end{itemize} 

In the author' opinion the Slime framework is quite promising, yet there are still additional strategies to improve the match selection to be investigated and possibly included in Slime, that will be more difficult to implement when working to heavy optimized code. This is the main reason for which the code has not been optimized until now.

\section*{Additional References}
\vspace{0.5em}
{\small
\begin{itemize}
	\item[{\label{magsacpp}[66]}] D. Barath, J. Noskova, M. Ivashechkin and J. Matas, ``MAGSAC++, a fast, reliable and accurate robust estimator,'' in \textit{Proc. IEEE/CVF Conf. Comput. Vis. Pattern Recognit. (CVPR)}, 2020.
	\item[{\label{roma}[67]}] J. Edstedt, Q. Sun, G. B\"{o}kman, M. Wadenb\"{a}ck and M. Felsberg, ``RoMa: Revisiting Robust losses for dense feature Matching,'' \textit{arXiv:2305.15404}, 2023.
\end{itemize}
}

\end{document}